%% file: main.tex
\documentclass[10pt,twocolumn,letterpaper]{article}

\usepackage{cvpr}
\usepackage{times}
\usepackage{epsfig}
\usepackage{graphicx}
\usepackage{amsmath}
\usepackage{amssymb}
\usepackage[utf8]{inputenc}
\usepackage[T1]{fontenc}
\usepackage{url}
\usepackage{booktabs}
\usepackage{amsfonts}
\usepackage{nicefrac}
\usepackage{microtype}
\usepackage{amsmath}
\usepackage{bm}
\usepackage{graphicx}
\usepackage{tabularx}
\usepackage{pifont}
\usepackage{multirow}

\newcommand{\cmark}{\ding{51}}
\newcommand{\xmark}{}
\newcolumntype{Y}{>{\centering\arraybackslash}X}
\newcommand{\FAKEREF}[1]{\textcolor{red}{{#1}}}

\newcommand{\modelA}{$\mathbb{A}$}
\newcommand{\modelB}{$\mathbb{B}$}
\newcommand{\modelCI}{$\mathbb{C}1$}
\newcommand{\modelCII}{$\mathbb{C}2$}
\newcommand{\modelDI}{$\mathbb{D}1$}
\newcommand{\modelDII}{$\mathbb{D}2$}
\DeclareMathOperator*{\argmax}{argmax}

\usepackage[pagebackref=true,breaklinks=true,letterpaper=true,colorlinks,bookmarks=false]{hyperref}

\cvprfinalcopy

\ifcvprfinal\fi
\begin{document}

\title{Unifying Training and Inference for Panoptic Segmentation}

\author{Qizhu Li\hspace{1em}Xiaojuan Qi\thanks{Xiaojuan Qi is now with the University of Hong Kong.}\hspace{1em}Philip H.S. Torr\\
University of Oxford\\
{\tt\small \{qizhu.li,~xiaojuan.qi,~philip.torr\}@eng.ox.ac.uk}
}

\maketitle
\ifcvprfinal\thispagestyle{empty}\fi

\maketitle

\input{contents/abstract.tex}
\input{contents/introduction.tex}
\input{contents/background.tex}
\input{contents/method.tex}
\input{contents/experiments.tex}
\input{contents/conclusion.tex}
\input{contents/acknowledgement.tex}

\input{main.bbl}
\input{supplemental/supp_wrapper.tex}

\end{document}

%% file: contents/abstract.tex
\begin{abstract}
	We present an end-to-end network to bridge the gap between training and inference pipeline for panoptic segmentation, a task that seeks to partition an image into semantic regions for ``stuff'' and object instances for ``things''. 
	In contrast to recent works, our network exploits a parametrised, yet lightweight panoptic segmentation submodule, powered by an end-to-end learnt dense instance affinity, to capture the probability that any pair of pixels belong to the same instance.
	This panoptic submodule gives rise to a novel propagation mechanism for panoptic logits and enables the network to output a coherent panoptic segmentation map for both ``stuff'' and ``thing'' classes, without any post-processing.
	Reaping the benefits of end-to-end training, our full system sets new records on the popular street scene dataset, Cityscapes, achieving $61.4$ PQ with a ResNet-50 backbone using only the \texttt{fine} annotations. On the challenging COCO dataset, our ResNet-50-based network also delivers state-of-the-art accuracy of $43.4$ PQ.
	Moreover, our network flexibly works with and without object mask cues, performing competitively under both settings, which is of interest for applications with computation budgets.
\end{abstract}

%% file: contents/introduction.tex
\section{Introduction}
As a pixel-wise classification task, panoptic segmentation aims to achieve a seamless semantic understanding of all countable and uncountable objects in a scene - \textit{a.k.a.}\ ``things'' and ``stuff'' respectively, and delineate the instance boundaries of objects where semantically possible.

While early attempts at tackling panoptic segmentation often resort to two separate networks for instance and semantic segmentation, recent works~\cite{li2018attention,li2018learning,kirillov2019panoptic,xiong2019upsnet,yang2019deeperlab} are able to improve the overall efficiency by constructing the two branches on a single, shared feature extractor, and training the multi-head, multi-task network jointly.
However, these works have stopped short of devising an end-to-end pipeline for panoptic segmentation, as they all adopt a post-processing stage with heuristics to combine the different outputs of their multi-task networks, following \cite{kirillov2018panoptic,kirillov2019panoptic}.
Such pipelines suffer from several shortcomings. 
Firstly, post-processing often requires a time-consuming trial-and-error procedure to mine a good set of hyperparameters, which may need to be repeated for each image domain.
As the performance of an algorithm can be quite sensitive to the choice of hyperparameters, how well a method performs can quickly degenerate to a function of the amount of computation resources at its disposal~\cite{krahenbuhl2011efficient,kirillov2019panoptic}.
Secondly, methods without an explicit loss function for panoptic segmentation~\cite{li2018attention,li2018learning,kirillov2019panoptic,yang2019deeperlab} cannot directly optimise for the ultimate goal.
Even with expert knowledge, it is difficult to design an exhaustive set of rules and remedies for all failure modes. An example is shown in Fig.~\ref{fig:teaser} (c): after the heuristic post-processing, the missing part of the car cannot be recovered.

\input{figures/teaser.tex}

To achieve an end-to-end system, we reckon three challenging steps need to be taken: (1) unify the training and inference, enabling the network to \textit{differentiably} produce panoptic segmentation during training; (2) embed a data-driven mechanism in the multi-task network whereby imperfect and coarse cues can be cleaned and corrected; (3) design an appropriate loss function to directly optimise the global objective for panoptic segmentation.

To achieve (1) and (2), we propose a novel pipeline using segmentation and localisation cues to predict a coherent panoptic segmentation in an end-to-end manner. At the heart of this pipeline lie a \textit{dynamic potential head} -- a parameter-free stage that represents a dynamic number of panoptic instances, and a \textit{dense instance affinity head} -- a parametrised, efficient, and data-driven module that predicts and utilises the likelihood for any pair of pixels to belong to the same ``thing'' instance or ``stuff'' class.
These two differentiable heads produces full panoptic segmentation during training and inference, eradicating the train-test logic discrepancy.

Furthermore, to fulfil (3), we propose a \textit{panoptic matching loss} which computes loss directly on panoptic segmentations. This objective function, together with the differentiable nature of our proposed panoptic head, enables the network to learn in an end-to-end manner.
To our best knowledge, our loss is the first to perform online segment matching before computing a cross entropy loss in an end-to-end panoptic segmentation system. The matching step allows training the network with \textit{predicted} detections, thereby incentivising it to handle imperfect localisation cues. While the idea is not convoluted, our ablation studies (Table \FAKEREF{C}, Supplementary) show that doing so -- as opposed to training with ground truth detections -- yields performance gains.

By closing the gap between training and inference, the network enjoys improved accuracy in challenging scenarios.
As illustrated in Fig.~\ref{fig:teaser}, by aggregating panoptic logits across the whole image according to the predicted affinity strengths (Fig.~\ref{fig:teaser}e), our parametrised panoptic head is able to fix inaccurate predictions from a previous stage - truncated objects due to imperfect bounding box localisations (Fig.~\ref{fig:teaser}c).

Last but not least, thanks to its power of improving coarse panoptic logits, our network achieves competitive performance even without using object mask cues, which are required in most recent approaches~\cite{li2018attention,li2018learning,kirillov2019panoptic,xiong2019upsnet}. This means our method can offer an additional degree of flexibility in terms of network design, a trait desirable for applications with a limited computation and time budget. On the challenging Cityscapes and COCO datasets, our models set new records for ResNet-50-based networks, achieving panoptic qualities (PQ) of 61.4 and 43.4 respectively.

%% file: figures/teaser.tex
\begin{figure*}
	\centering
	\setlength{\tabcolsep}{1pt}
	\begin{tabularx}{\textwidth}{YYYYY}
		\global \def \VisScale{1}
		\includegraphics[width=\VisScale\linewidth]{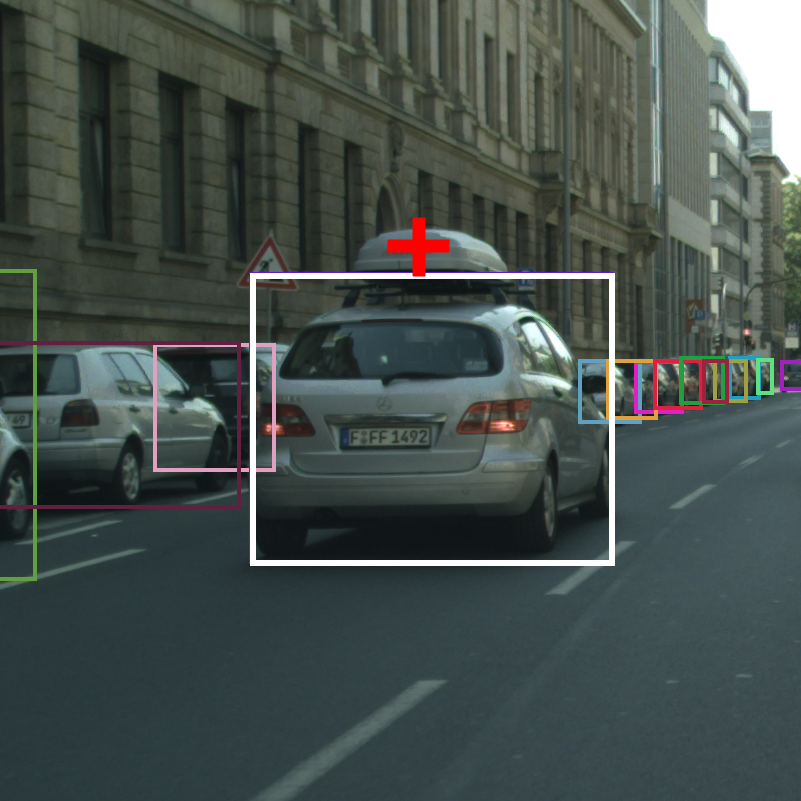}&
		\includegraphics[width=\VisScale\linewidth]{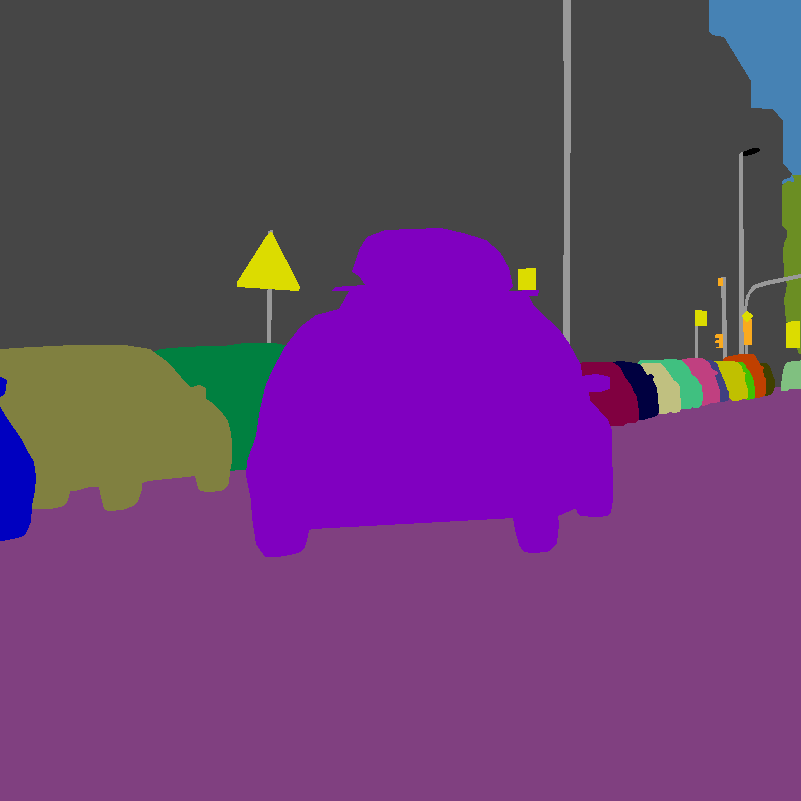}&
		\includegraphics[width=\VisScale\linewidth]{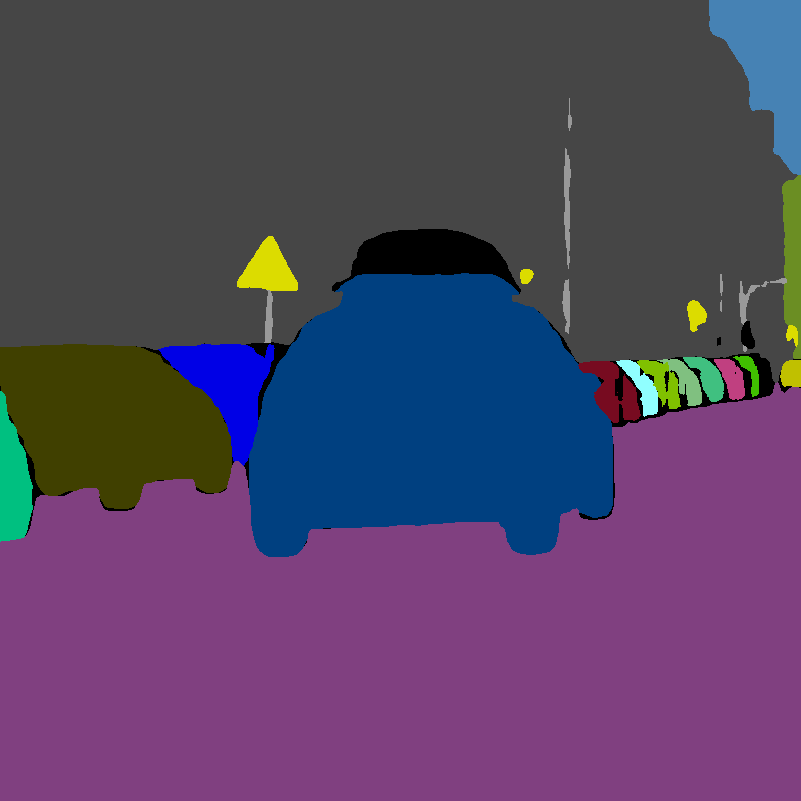}&
		\includegraphics[width=\VisScale\linewidth]{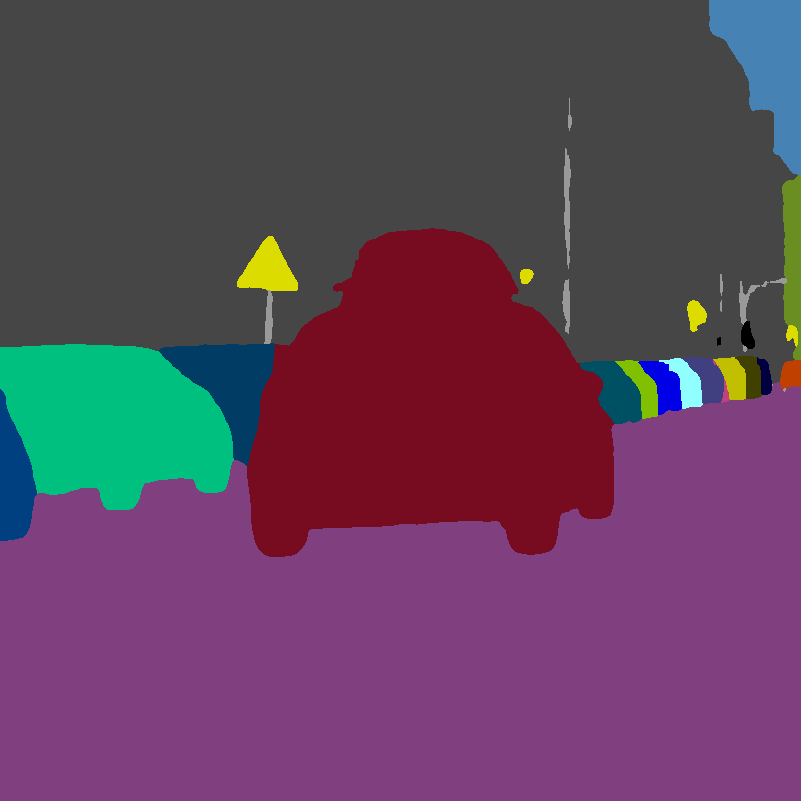}&
		\includegraphics[width=\VisScale\linewidth]{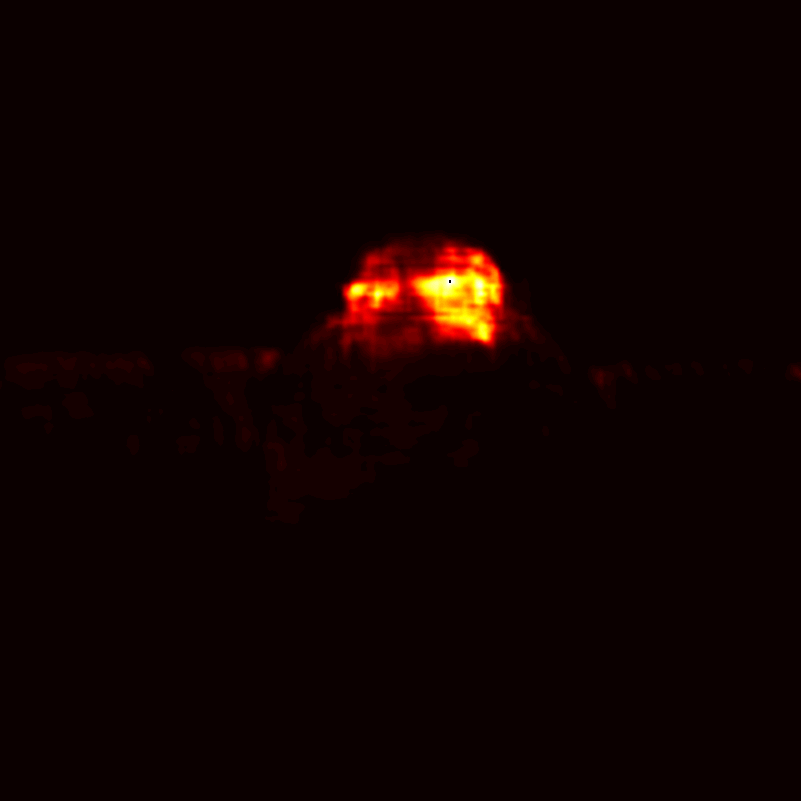}\\
		{\footnotesize{(a) Image with boxes}}&{\footnotesize{(b) Ground truth}}&{\footnotesize{(c) Heuristic fusion}}&{\footnotesize{(d) Our approach}}&{\footnotesize{(e) Instance affinities}} \\
	\end{tabularx}
	\vspace{2pt}
	\caption{Comparison of our approach \textit{vs}. the heuristic rule-based method of ~\cite{kirillov2019panoptic}. We overlay the predicted bounding boxes on the input images for visualisation. For the cross-marked pixel in (a) which falls outside its bounding box, we show its instance affinities in (e). Heuristics-based fusion~\cite{kirillov2019panoptic} produces truncated objects when localisation is not accurate, while our instance affinity enables the network to recover the full object, by propagating information between pixels with strong instance affinities. Best viewed in colour.}
	\label{fig:teaser}
\end{figure*}

%% file: contents/background.tex
\section{Related work}
\label{sec:related_work}
Arguably, the problem of panoptic segmentation can be viewed as a combination of instance and semantic segmentation. Indeed, this interpretation has guided many recent works on panoptic segmentation~\cite{kirillov2018panoptic,kirillov2019panoptic}, where it is largely approached as a bi-task problem, and the focus is placed on solving both sub-problems simultaneously and efficiently. Shared features of these works include the use of networks with multiple specialised subnets for each sub-task, and the lack of an explicit objective on panoptic segmentation.

In addition to the inclusion of ``stuff'' classes, another major difference between panoptic and instance segmentation is that the former requires all pixels to be given a unique label, whereas the latter does not.
As a result, ``thing'' predictions from an off-the-shelf detection-driven instance segmentation network -- \textit{e.g.}, Mask-RCNN~\cite{he2017mask} -- cannot be readily inserted into the panoptic prediction, as pixels need to have their conflicting instance labels resolved. Moreover, contradictions between the semantic and instance branch must also be carefully resolved. This prompted recent works to adopt an offline postprocessing step first described in~\cite{kirillov2018panoptic} to perform conflict resolution and merger of instance and semantic predictions, based on a set of carefully tuned heuristics. A number of works have also attempted to encourage consistency between semantic and instance predictions by adding a communication mechanism
between the two subnets~\cite{li2018learning,li2018attention}. However, as these proposed changes do not modify the output format of the network, they still rely on postprocessing to produce panoptic predictions.
In addition, Liu \etal proposes to directly learn the ordering of ``thing'' instances for conflict resolution~\cite{liu2019end}. However, this approach does not handle overlapping instances pixel-by-pixel -- as it predicts a single ranking score for each instance -- and does not reconcile conflicts between ``stuff'' and ``thing''.

A small number of works have attempted to advance towards an end-to-end network with a unified train-test logic. We observe that \cite{li2018weakly} extends a dynamically instantiated instance segmentation network described in ~\cite{arnab2017pixelwise} to solve the panoptic segmentation problem.
It produces non-overlapping segments by design, and is trained end-to-end, given detections.
However, it is prone to failures when objects of the same class are nearby and similarly coloured. Moreover, its Instance CRF suffers from the very small number of trainable parameters (since the compatibility transforms are frozen as the Potts model), and is made less attractive by the need to grid search good kernel variances for the bilateral filters in the message passing step.

Recently, Xiong~\etal~\cite{xiong2019upsnet} modifies the unary terms of~\cite{arnab2017pixelwise,li2018weakly} and proposes a parameter-free, differentiable panoptic head to fuse semantic and instance segmentation predictions during training. Similar to~\cite{li2018weakly}, it allows a panoptic loss to be directly applied on the fused probabilities. However, in the inference phase, it still resorts to several heuristic strategies (\textit{e.g.}, overlap-based instance mask pruning) and relies on a complex voting mechanism to determine the semantic categories of predicted segments, deviating from a unified training and inference pipeline.
Furthermore, the effectiveness of their parameter-free panoptic head heavily depends on the quality of semantic and instance predictions it receives, since it arguably functions as an online heuristic merger due to the absence of learnable weights.

Also pertinent to this work is the extensive research carried out around the techniques of long-range contextual aggregation. Aside from CRF-driven methods~\cite{krahenbuhl2011efficient,zheng2015conditional,arnab2017pixelwise}, Bertasius~\etal~proposes a semantic segmentation method based on random walks to learn and predict inter-pixel affinity graphs, and iteratively multiply the learnt affinity with an initial segmentation to achieve convergence~\cite{bertasius2017convolutional}. 
Lately,
another technique, self-attention, has been successful in several vision tasks~\cite{wang2018non, yuan2018ocnet, fu2018dual}.
However, its quadratic memory and computation complexity has cast doubt over its practicality.
To mitigate this problem, Shen~\etal~\cite{shen2018decomposed} suggests to invoke the associativity of matrix multiplication and avoid the explicit production of expensive attention maps.
This approach effectively reduces the complexity to a linear one, $O(HW)$, making it suitable for pixel-level labelling tasks.

Albeit sharing certain operational similarities with self-attention and non-local methods~\cite{yuan2018ocnet,huang2018ccnet,wang2018non}, our proposed dense instance affinity head serves a different purpose, and cannot be substituted by directly inserting these operations in the backbone. The aforementioned methods work by enhancing the expressiveness of extracted features, as reflected in the fact that these actions are performed in the feature space, and can generally lead to performance gains for many tasks. In contrast, our proposed instance affinity is not a generic feature enhancer. It is specifically designed and tasked to model the pairwise probability for any two pixels to belong in the same ``thing'' instance or ``stuff'' category. This relationship in turn enables our network to revise and resolve. With this purpose in mind, we incorporate insights from~\cite{shen2018decomposed} to construct a module that is lightweight, learnable, and agnostic to the number of channels, allowing us to model a dynamic number of instances across different images.

%% file: contents/method.tex
\section{Proposed approach}
\label{sec:proposed_approach}
\input{figures/model/overview.tex}
Our proposed network (Fig.~\ref{fig:model}) consists of four blocks. A shared \textit{fully convolutional backbone} extracts a set of features. Operating on these features, a \textit{semantic segmentation submodule} and an \textit{object detection submodule} produce segmentation and localisation cues, which are fused and revised by the proposed \textit{panoptic segmentation submodule}. All components are differentiable and trained jointly, end-to-end.

\subsection{Backbone}
The pipeline starts with a shared fully convolutional backbone, which takes an input image of spatial dimension $H\times W$, and generates a set of features $\bm{F}$. In our experiments, we adopt a simple ResNet-FPN backbone that outputs four multi-scale feature maps~\cite{lin2017feature}, following a common practice in prior works~\cite{kirillov2019panoptic,xiong2019upsnet}. To encourage global consistency, we carry out a squeeze-and-excitation operation~\cite{hu2018squeeze} on the top-level ResNet feature before producing the first FPN feature. A similar strategy is used in~\cite{xiong2019upsnet}.

\subsection{Semantic segmentation submodule}
The backbone features $\bm{F}$ are fed into the semantic segmentation submodule to produce a $\frac{H}{d} \times \frac{W}{d} \times (N_{st} + N_{th})$ tensor $\bm{V}$, where $N_{st}$ and $N_{th}$ are the number of ``stuff'' and ``thing'' classes respectively.
$V_i(l)$ denotes the probability that pixel $p_i$ belongs to semantic class $l$.
The spatial dimension is downsampled $d$ times to strike a balance between resolution and complexity. We choose $d$ as 4 in the experiments.

Multiple implementations for this submodule have been proposed in the literature, all showing decent performance~\cite{kirillov2019panoptic,xiong2019upsnet}. In this work, we modify the design in~\cite{xiong2019upsnet} by inserting a Group Normalisation operation~\cite{wu2018group} after each convolution, which has been observed to help stabilise training. Please refer to the supplementary for further details.

\subsection{Object detection submodule}
In parallel, the features $\bm{F}$ are also passed to an object detection submodule, which generates $D$ object detections, consisting of bounding boxes $\bm{B} = \{B_1, B_2, B_3, ..., B_D\}$, confidence scores $\bm{s} = \{s_1, s_2, s_3, ..., s_D\}$, and predicted classes $\bm{c} = \{c_1, c_2, c_3, ..., c_D\}$. Additionally, we add a whole image bounding box for each ``stuff'' class to the object detection predictions, raising the total number of detections to $D+N_{st}$. Doing so allows the panoptic submodule to process ``things'' and ``stuff'' with a unified architecture.

Notably, the versatility of the panoptic submodule allows our network to work with or without object masks. When the object detection submodule has the capability to predict instance masks for ``things'' $\bm{M} = \{M_1, M_2, M_3, ..., M_D\}$, they are easily incorporated into the dynamic potential $\bm{\Psi}$. Details will be given in Sec.~\ref{subsec:dynamic_head}.

\subsection{Panoptic segmentation submodule}
\input{figures/model/panoptic_seg_submod.tex}
This submodule serves as the mastermind of the pipeline. Receiving cues from the two prior submodules, the panoptic segmentation submodule combines them into a dynamic potential $\bm{\Psi}$ (Sec.~\ref{subsec:dynamic_head}) and revises it according to predicted pairwise instance affinities (Sec.~\ref{subsec:dense_instance_affinity}), producing the final panoptic segmentations with the same logic in training and inference. This pipeline is illustrated in Fig.~\ref{fig:pan_head}.

\subsubsection{Dynamic potential head}
\label{subsec:dynamic_head}
\input{figures/model/dynamic_potential_head.tex}

The dynamic potential head functions as an assembly node for segmentation and localisation cues from prior submodules.
This head is capable of representing varying numbers of instances as it outputs a \textit{dynamic} number of channels, one for each object instance or ``stuff'' class.
We present three variants of dynamic head design, as illustrated in Fig.~\ref{fig:dyn_head}.
Variant A is proposed in~\cite{xiong2019upsnet}, whereas the mask-free parent of B and C is first described in~\cite{arnab2017pixelwise} as the box consistency term. A main difference between variant A and the rest is the absence of detection score in A. We argue that leveraging detection scores can suppress false positives in the final output, as unconfident detections will be attenuated by its score. 
Thus, we will describe variant B and C in more details.

Given $(D + N_{st})$ bounding boxes $\bm{B}$ and box classes $\bm{c}$ (including the dummy full-image ``stuff'' boxes), it populates each box region with a combination of semantic segmentation probabilities $\bm{V}$ and box confidence scores $\bm{s}$ to produce a \textit{dynamic potential} $\bm{\Psi}$ with $(D + N_{st})$ channels:

\begin{equation}
\Psi_i (k) =
\begin{cases}
s_k V_i (c_k) & \mathrm{for}~i \in B_i \\
0 & \mathrm{otherwise} \\
\end{cases}
\end{equation}

Optionally, if provided with object masks $\bm{M}$, the dynamic potential head can also incorporate them into $\bm{\Psi}$. Defining $\bm{M}$ to be image-resolution instance masks where the raw masks have been resized to their actual dimensions and pasted to appropriate spatial locations in image, the dynamic potential with object masks can be summarised as:

\begin{equation}
\Psi_i (k) = 
\begin{cases}
s_k \big[V_i (c_k) \odot M_{i}(k)\big] & \mathrm{for}~i \in B_i \\
0 & \mathrm{otherwise} \\
\end{cases}
\end{equation}

In variant B and C, operator $\odot$ is multiplication and summation respectively. More analysis of the variant B and C are included in the supplementary.

\subsubsection{Dense instance affinity head}
\label{subsec:dense_instance_affinity}
\input{figures/model/dense_affinity_head.tex}
We observe that the dynamic potential $\bm{\Psi}$ often carries conflicts and errors due to imperfect cues from semantic segmentation and object localisation.
This motivates the design of this parametrised head, with the aim to enable a data-driven mechanism that resolves and revises the output of the dynamic potential head.
The main difficulty with injecting parameters into an instance-level head is the varying number of instances across images, which practically translates to a dynamic number of channels in the input tensor. On the other hand, the fundamental building block of a convolutional neural network -- convolution -- is designed to handle a fixed number of input channels. This apparent incompatibility has led prior works on panoptic segmentation to use either no parameter at all~\cite{xiong2019upsnet}, or only single scaling factors for entire tensors~\cite{li2018weakly} providing limited modelling capacity.

This conundrum can be tackled by driving this head with a pairwise dense instance affinity, which is predicted from data, fully differentiable, and compatible with a dynamic number of input channels. By integrating global information according to the pairwise affinities, it produces the final panoptic segmentation probabilities, from which inference can be trivially made with an \texttt{argmax} operation along the channel dimension. Thus, it is amenable to a direct panoptic loss, an ingredient of an end-to-end network.

To construct the dense instance affinity, this head first extracts from the backbone features $\bm{F}$ a single feature tensor $\bm{Q}$ of dimension $\frac{H}{d} \times \frac{W}{d} \times C$, where $C$ is the number of feature channels, and $d$ is a downsampling factor. This corresponds to the affinity feature extractor in Fig.~\ref{fig:affinity_head}. The spatial dimensions of $\bm{Q}$ can be easily collapsed to produce a $\frac{HW}{d^2} \times C$ feature matrix.

Normally, the pairwise instance affinities $\bm{A}$ -- a large $\frac{HW}{d^2} \times \frac{HW}{d^2}$ matrix -- would then be produced by performing a matrix multiplication $\bm{A} = \bm{Q}\bm{Q}^{T}$.
This would be followed by multiplying $\bm{A}$ with a $\frac{HW}{d^2} \times C'$ input tensor to complete the process. It is, however, prohibitively expensive due to the quadratic complexity with respect to $HW$.
In a typical training step, where $(H,W)=(800,1300)$ and $d=4$, a single precision matrix with the size of $\bm{A}$ would occupy 15.7GB of GPU memory, making this approach unpractical.

Drawing from insight of~\cite{shen2018decomposed}, we design a lightweight pipeline for computing and applying the dense instance affinities (Fig.~\ref{fig:affinity_head}). Instead of sequentially computing $\bm{Q}\bm{Q}^{T}\bm{\Psi}$ which explicitly produces $\bm{A}$, we compute $\bm{Q}\big(\bm{Q}^{T}\bm{\Psi}\big)$, since:
\begin{equation}
	\big(\bm{Q}\bm{Q}^{T}\big)\bm{\Psi} = \bm{Q}\big(\bm{Q}^{T}\bm{\Psi}\big)
\end{equation}
The result of $\bm{Q}^T\bm{\Psi}$ is a very small $C\times(D + N_{st})$ tensor, taking only tens of kilobytes.
In terms of computation, using the same $H$, $W$, $d$ as the example above and $(C,D,N_{st}=128,100,53)$ as typically used in experiments, the efficient implementation reduces the total number of multiply-adds by 99.8\% to 5 billion FLOPS. For reference, a ResNet-50-FPN backbone at the same input resolution requires 140 billion FLOPS.

Finally, we add the product back to the input, forming a residual connection to ease the learning task.
As such, the full action of our dense instance affinity applier can be summarised with the following expression:
\begin{equation}
    \bm{P} = \bm{\Psi} + \phi_0(\bm{Q}) \big(\phi_1(\bm{Q}^{T}) \bm{\Psi}\big)
\end{equation}
where $\phi_0$ and $\phi_1$ are each a $1\times1$ convolution followed by an activation. From this formulation, inference is straight forward and does not require any post-processing, as an \texttt{argmax} operation on $\bm{P}$ along the channel direction readily produces the panoptic segmentation prediction.

Note that we do not compute a loss directly over $\bm{Q}$; instead, the instance affinities are implicitly trained by supervision from the panoptic matching loss described in the next section. In the preliminary experiments, we tried directly supervising $\bm{Q}$ with a contrastive loss, but did not observe performance gains. This shows that our end-to-end training scheme with the panoptic matching loss is already able to guide the model to learn effectively. Detailed discussion of the dense instance affinity operation, with ablation studies and visualisations, is provided in Sec.~\ref{subsec:ablation_studies}.

For simplicity, the affinity feature extractor adopts the same architecture as our semantic segmentation submodule.
We use $C=128$ in all experiments.

\subsection{Panoptic matching loss}
For instance-level segmentation, different permutations of the indices in the segmentation map are qualitatively equivalent, since the indices merely act to distinguish between each other, and do not carry actual semantic meanings.

During training, we feed predicted object detections into the panoptic segmentation submodule. As a result, the indices of the instances are not fixed or known before hand. To compute loss, we first match the ground truth segmentation to the predicted detections by maximising the intersection over union between their bounding boxes (box IoU). Given a set of $\alpha$ ground truth segments $\bm{\mathcal{T}} = \{\mathcal{T}_1, \mathcal{T}_2, \mathcal{T}_3,...,\mathcal{T}_{\alpha}\}$, and a set of $\beta$ predicted bounding boxes $\bm{B} = \{B_1, B_2, B_3, ..., B_\beta\}$, we find the ``matched'' ground truth $\bm{\mathcal{T}}^{\star}$ which satisfies:
\begin{equation}
    \bm{\mathcal{T}}^{\star} = \argmax_{\bm{\mathcal{Z}}\in\pi(\bm{\mathcal{T})}} \mathrm{IoU}_{t}(\mathrm{box}(\bm{\mathcal{Z}}), \bm{B})
\end{equation}
where $\mathrm{box}(.)$ extracts tight bounding boxes from segments, $\pi(\bm{\mathcal{T}})$ refers to all permutations of $\bm{\mathcal{T}}$, and $t$ sets the minimum match threshold for a match to qualify as valid. Note that the box IoU between different semantic classes are taken to be $0$, and $\alpha$ and $\beta$ need not be the same. Ground truth segments without matched predictions are set to the ``ignore'' label, and detections matching to the same ground truth segment are all removed except the top match, before being fed into the panoptic submodule. Both cases do not contribute any gradients. With the ``matched'' ground truth segmentation $\bm{\mathcal{T}}^{\star}$, we can compute the loss on the predicted panoptic segmentation probabilities $\bm{P}$ as per normal with a cross-entropy loss. Our experiments use $0.5$ for $t$.

Unlike ours, the panoptic loss used by~\cite{xiong2019upsnet} does not have the matching stage and its panoptic head is trained with ground truth detections instead. As a result, the models of \cite{xiong2019upsnet} are not trained to handle imperfect localisations.
In addition, our loss differs from \cite{liu2019end} as the loss used by their \textit{spatial ranking module} does not directly supervise panoptic segmentation, does not take ``stuff'' into account, and thus does not globally optimise in an end-to-end way.

%% file: figures/model/overview.tex
\begin{figure*}[t]
\vspace{-\baselineskip}
\centering
\includegraphics[width=0.98\linewidth]{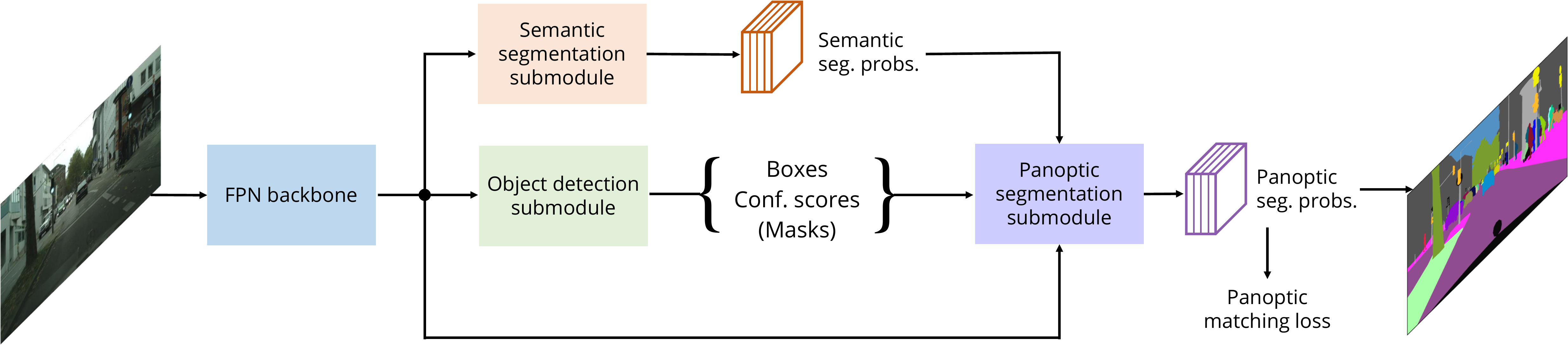}
\vspace{5pt}
\caption{Overview of the network architecture. Semantic segmentation and object detections are fed into the proposed panoptic segmentation submodule -- including a dynamic potential head and a dense instance affinity head -- to produce panoptic segmentation predictions without requiring post-processing. All components are differentiable, and the network is trained end-to-end.
}
\label{fig:model}
\end{figure*}

%% file: figures/model/panoptic_seg_submod.tex
\begin{figure}[t]
\centering
\includegraphics[width=0.98\linewidth]{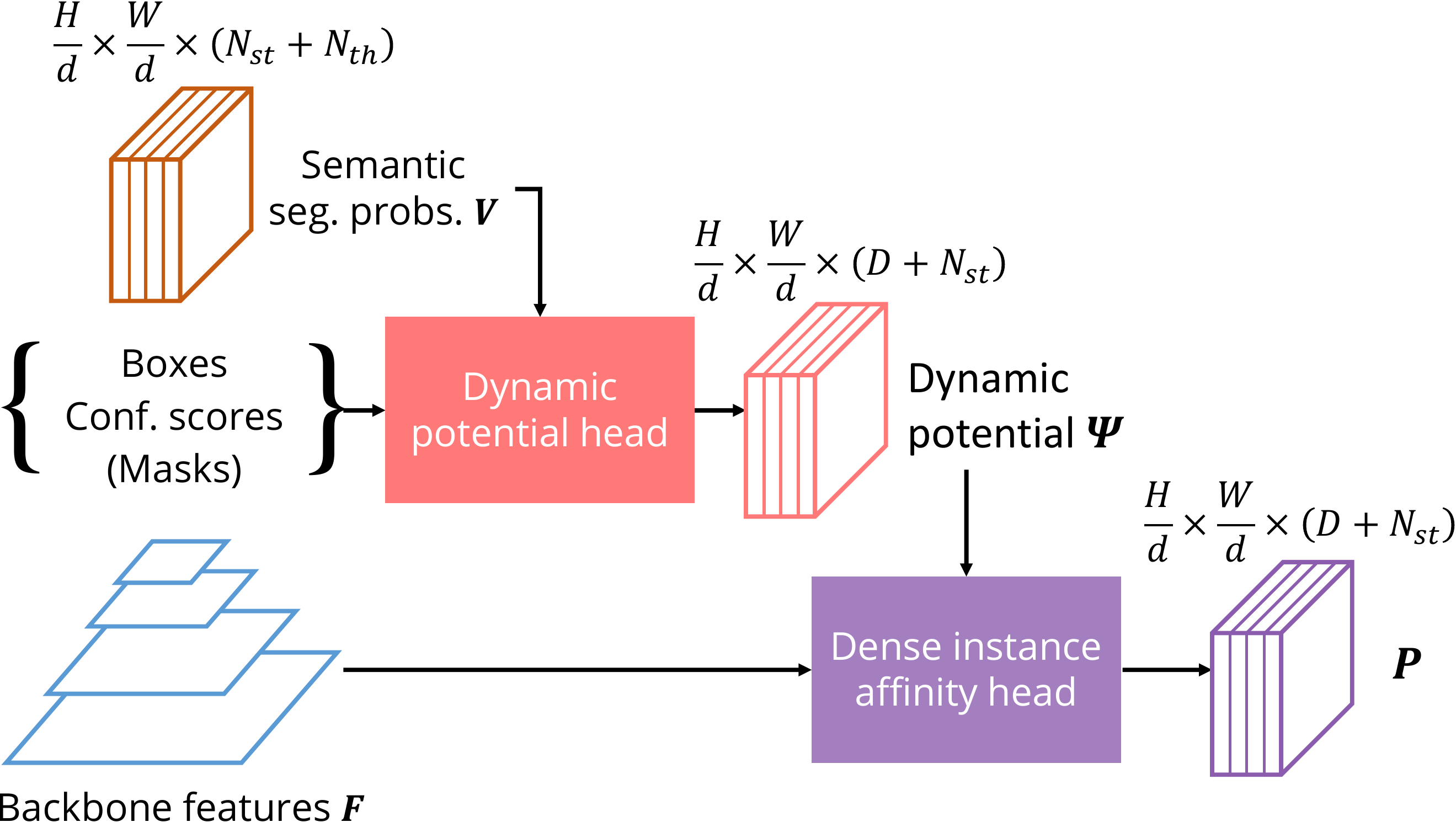}
\vspace{2pt}
\caption{The panoptic segmentation submodule. Details on the dynamic potential head and dense instance affinity head are further clarified in Fig.~\ref{fig:dyn_head} and~\ref{fig:affinity_head} respectively.}
\label{fig:pan_head}
\end{figure}

%% file: figures/model/dynamic_potential_head.tex
\begin{figure}[t]
\centering
\includegraphics[width=0.98\linewidth]{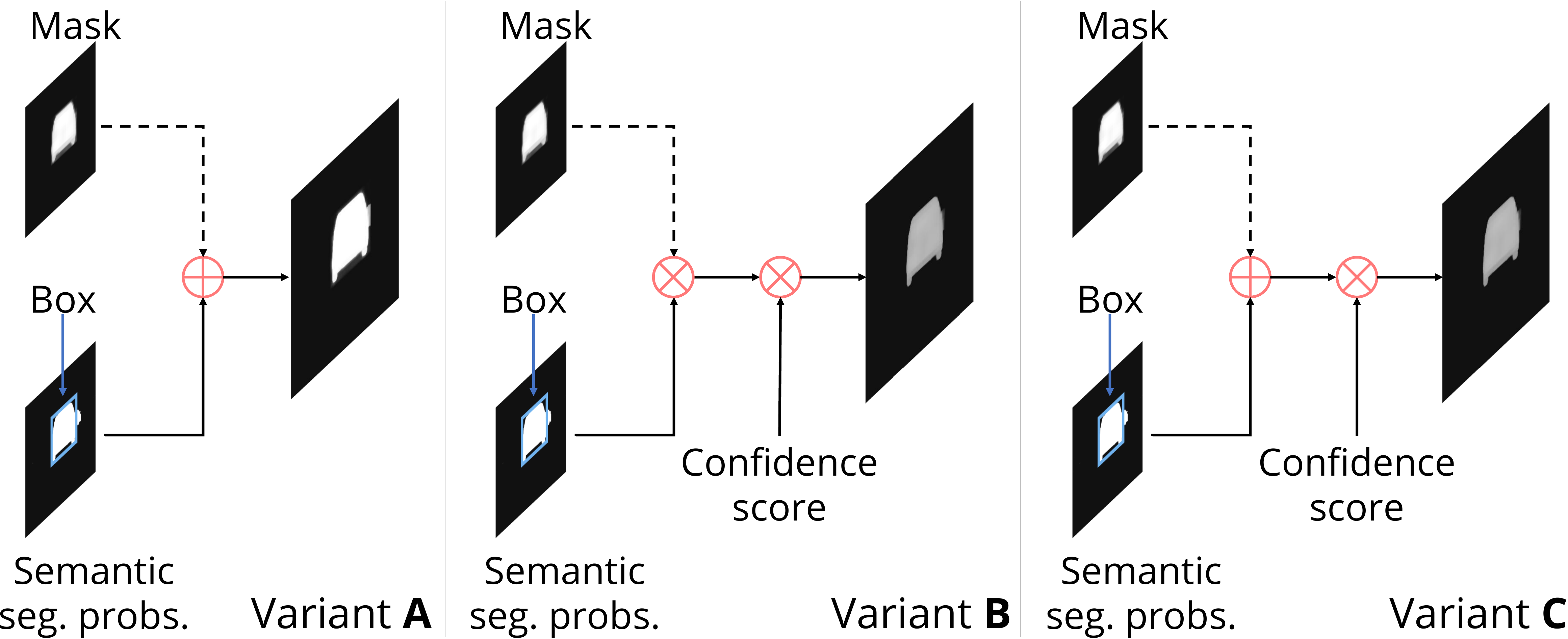}
\vspace{2pt}
\caption{Three variants of the dynamic potential head. For clarity, we only show one instance in each diagram. In practice, the same operation is extended to all detections and ``stuff''. Note that the dotted path is only activated when masks are provided to the head. When no masks are given, variant B and C are equivalent.}
\label{fig:dyn_head}
\end{figure}

%% file: figures/model/dense_affinity_head.tex
\begin{figure*}[ht]
\vspace{-\baselineskip}	
\centering
\includegraphics[width=0.9\linewidth]{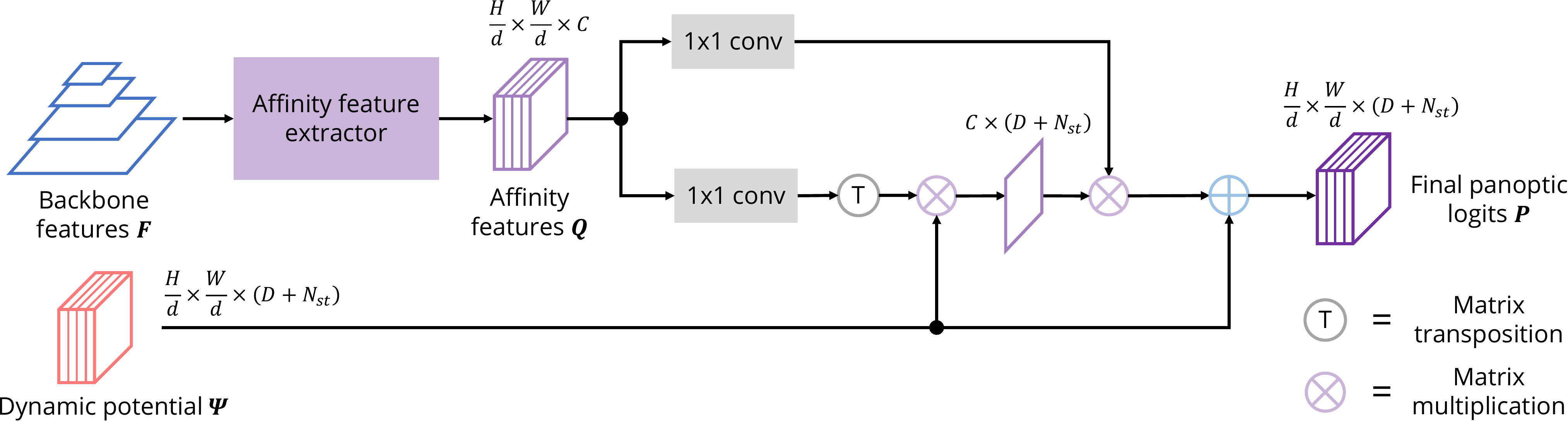}
\vspace{2pt}
\caption{The dense instance affinity head. It is parametrised, expressive, lightweight, and fully differentiable. }
\label{fig:affinity_head}
\end{figure*}

%% file: contents/experiments.tex
\section{Experimental evaluation}
\label{sec:experiments}
\paragraph{Cityscapes.}
The Cityscapes dataset features high resolution road scenes
with 11 ``stuff'' and 8 ``thing'' classes. There are 2,975 training images, 500 validation images, and 1,525 test images. We report on its validation set and test set.

\paragraph{COCO.}
The COCO panoptic dataset has a greater number of images and categories. It features 118k training images, 5k validation images, and 20k \textit{test-dev} images. There are 133 semantic classes, including 53 ``stuff'' and 80 ``thing'' categories. We report on its validation set and \textit{test-dev} set.

\paragraph{Evaluation metric.}
\label{para:evaluation_metric}
Our main evaluation metric is the panoptic quality (PQ), which is the product of segmentation quality (SQ) and recognition quality (RQ)~\cite{kirillov2018panoptic}.
SQ captures the average segmentation quality of matched segments, whereas RQ measures the ability of an algorithm to correctly detect objects.

We also report the mean Intersection over Union (IoU) score of our initial category-level segmentation $\bm{V}$, and the box Average Precision ($AP_{box}$) of our predicted bounding boxes $\bm{B}$. Additionally, for models which predict object instance masks $\bm{M}$ in the object detection submodule, we report its mask Average Preicision ($AP_{mask}$) as well. Both $AP_{box}$ and $AP_{mask}$ are averaged across IoU thresholds between $0.5$ and $0.95$, at increments of $0.05$.

\paragraph{Cityscapes training.} We follow most of the learning settings described in~\cite{kirillov2019panoptic}. We distribute the 32 crops in a minibatch over 4 GPUs instead.
The weights for the detection, semantic segmentation, and panoptic segmentation losses are set to 0.25, 1.0, and 1.0 respectively.

\paragraph{COCO training.} We follow most of the learning settings for COCO experiments in~\cite{kirillov2019panoptic}. For the learning schedule, we train for 200k iterations with a base learning rate of 0.02, and reduce it by a factor of 10 at 150k and 190k iterations.
While this learning schedule differs from that used in~\cite{kirillov2019panoptic}, we found that our panoptic submodule with its additional parameters benefits from the new schedule.
In terms of loss weights, we use 1.0, 0.2, and 0.1 for the object detection, semantic segmentation, and panoptic segmentation losses.

\subsection{Ablation studies}
\label{subsec:ablation_studies}
\input{tables/ablation_studies.tex}
We conduct detailed ablation studies for five different settings, including two architecture choices (msk. and aff.), one training strategy (e2e.), and two inference options (heu. and amx.). We report the results in Table~\ref{tab:ablation_studies}. Explanations for the abbreviations can be found in the table caption. For clarity, we provide a brief description of the ablation models:
\begin{itemize}
	\item Model \modelA{} uses a Faster-RCNN head as its object detection submodule, and has neither the dense instance affinity head nor the panoptic matching loss. The dynamic potential $\bm{\Psi}$ is used as the final output $\bm{P}$.
	\item Model \modelB{} differs from \modelA{} by employing the dense instance affinity head and the panoptic matching loss.
	\item In \modelCI{} and \modelCII{}, the model uses a Mask-RCNN head as its object detection submodule, and has neither the dense instance affinity head nor the panoptic matching loss. During inference, \modelCI{} merges the semantic and instance segmentation predictions using heuristics~\cite{kirillov2018panoptic}, whereas \modelCII{} outputs the dynamic potential $\bm{\Psi}$ as $\bm{P}$.
	\item The pair (\modelDI{}, \modelDII{}) differs from (\modelCI{}, \modelCII{})  by employing the instance affinity and the panoptic matching loss.
\end{itemize}
Note that model \modelA{} and \modelB{} do not produce nor use object mask predictions, and are therefore not possible to test with the heuristic merger strategy~\cite{kirillov2019panoptic}. In addition, the pair \modelCI{} and \modelCII{}, as well as \modelDI{} and \modelDII{}, are identical models using different inference methods.

\paragraph{Dense instance affinity.}
Comparing across model \modelA{} and \modelB{}, it is evident that training and testing with the proposed dense instance affinity leads to significant performance boosts. Increased performances are seen across all metrics, with the largest rises in PQ (+4.4 for all, +4.2 for ``things'' and +4.4 for ``stuff'') and RQ (+4.0). This testifies to the effectiveness of the dense instance affinity, even with only box predictions.
A similar trend is also evident with object masks enabled, between model \modelCII{} with \modelDII{}, recording a 1.8 rise in overall PQ. Fig.~\ref{fig:vis_affinity} visualises some examples of instance affinities, with more in the supplementary materials.

\paragraph{End-to-end training with panoptic matching loss.}
While \modelCI{} and \modelDI{} are trained differently -- with the former being trained jointly, and the latter being trained end-to-end with the panoptic matching loss -- they are tested using the same heuristic strategy~\cite{kirillov2019panoptic}. Therefore, the 1.3 increase in PQ of \modelDI{} over \modelCI{} solely stems from the fact that \modelDI{} undergoes end-to-end training, and shows that our end-to-end training strategy with the panoptic matching loss is effective.

\paragraph{Unified training and inference pipeline.}
For \modelDI{}, we test a model trained end-to-end with the panoptic matching loss using the heuristic merger strategies. In contrast, for \modelDII{}, we take the same model and take \texttt{argmax} from the final panoptic logits. We can see that the \modelDII{} still outperforms \modelDI{} by 0.8 PQ, giving proof for the benefit of having a unified training and testing pipeline.

\subsection{Comparison with state-of-the-art}
\paragraph{Cityscapes.} We compare our results with other methods on Cityscapes validation set in Table~\ref{tab:cityscapes_panoptic_comparison}. All entries are ResNet-50~\cite{he2016deep} based except~\cite{li2018weakly,yang2019deeperlab}.
We sort prior works into two tracks, depending on whether the network performs instance segmentation internally.
For both tracks, our method achieves the state-of-art.
The most telling comparison is between our model and UPSNet, as these methods have a similar network architecture other than our proposed panoptic segmentation submodule. Our network is able to outperform UPSNet by $2.1$ PQ.
On the other hand, among methods that do not rely on instance segmentation~\cite{li2018weakly,yang2019deeperlab}, our system outperforms the previous state-of-art by $3.5$ PQ, even though they utilise stronger backbones (Xception-71~\cite{chollet2017xception} and ResNet-101~\cite{he2016deep}) than ours (ResNet-50).

Speed-wise, our design compares favourably with other state-of-the-art models. On Cityscapes, inference takes $386$ms\footnote{Obtained by running our re-implementation.} and $201$ms\footnote{Obtained by running its publicly released code.} per image for~\cite{kirillov2019panoptic} and~\cite{xiong2019upsnet}, whereas our full model runs at $197$ms per image. All models are ResNet-50 based and timed on a single RTX 2080Ti card.

\input{tables/cityscapes_panoptic_results_halfcol.tex}
\input{tables/coco_panoptic_results.tex}

\paragraph{COCO.} Results on the COCO panoptic validation set are reported in Table~\ref{tab:coco_panoptic_comparison}. Due to the disentangling power of our proposed pipeline and unified train-test logic, we are able to outperform the previous state-of-art method by $0.9$ in terms of overall PQ, and 2.1 in terms of PQ for ``stuff''.

Results on the Cityscapes test set and COCO \textit{test-dev} set are reported in Table~\ref{tab:cityscapes_test_set_results} and~\ref{tab:coco_test_set_results}. We perform \textit{single-scale} inference, without any test-time augmentation. For fair comparison, only methods that are ResNe(X)t-based are reported. Our method achieves the state-of-art performance on both datasets with a PQ of $63.3$ and $47.2$ respectively.

\input{tables/cityscapes_test_set_results.tex}
\input{tables/coco_test_set_results.tex}

\input{figures/vis_affinity.tex}
\input{figures/cityscapes_results.tex}

Qualitative results are shown in Fig~\ref{fig:qual} where we compare with our re-implementation of Panoptic FPN.
As the instance affinity operation integrates information from pixels locally and globally, our method can resolve errors in the detection stage by propagating meaningful information from other pixels. The ``void'' region (displayed in black) shown in Fig~\ref{fig:qual}c are typically present in results produced by the heuristic merging process popularised by~\cite{kirillov2018panoptic}. They are due to the method's inability to resolve inconsistencies between semantic and instance predictions. In contrast, our method successfully handles such cases, as evident in Fig.~\ref{fig:qual}d.

%% file: tables/ablation_studies.tex
\begin{table*}[h!]
	\centering
	\scalebox{0.83}{
		\begin{tabularx}{1.2\linewidth}{lYYYYYYYYYYYYY}
			\toprule
			& \multicolumn{5}{c}{Settings} & \multicolumn{3}{c}{PQ} & \multicolumn{1}{c}{SQ} & \multicolumn{1}{c}{RQ} & \multicolumn{1}{c}{IoU} & $AP$ & $AP$\\

			Model & msk. & aff. & e2e. & heu. & amx. & all & th. & st. & all & all & all & mask & box \\
			\cmidrule(r){1-1} \cmidrule(r){2-6} \cmidrule(r){7-9} \cmidrule(l){10-10} \cmidrule(l){11-11} \cmidrule(l){12-12} \cmidrule(l){13-14}
			
			\modelA{} & \xmark & \xmark & \xmark & \xmark & \cmark & 54.6 & 46.0 & 60.9 & 77.9 & 68.4 & 75.0 & -- & 36.9\\
			
			\modelB{} & \xmark & \cmark & \cmark & \xmark & \cmark & 59.0 & 50.2 & 65.3 & 80.1 & 72.4 & 77.8 & -- & 38.1 \\
			
			\cmidrule(r){1-1} \cmidrule(r){2-6} \cmidrule(r){7-9} \cmidrule(l){10-10} \cmidrule(l){11-11} \cmidrule(l){12-12} \cmidrule(l){13-14}
			
			\modelCI{} & \cmark & \xmark & \xmark & \cmark & \xmark & 59.3 & 51.4 & 65.0 & 79.8 & 73.2 & 78.1 & 33.8 & 38.1 \\ 
			
			\modelCII{} & \cmark & \xmark & \xmark & \xmark & \cmark & 59.6 & 52.4 & 64.8 & 80.4 & 72.9 & 78.1 & 33.8 & 38.1 \\
			
			\modelDI{} & \cmark & \cmark & \cmark & \cmark & \xmark & 60.6 & 52.4 & 66.5 & 80.4 & 74.2 & 79.5 & 33.7 & 38.8 \\
			
			\modelDII{} & \cmark & \cmark & \cmark & \xmark & \cmark & 61.4 & 54.7 & 66.3 & 81.1 & 74.7 & 79.5 & 33.7 & 38.8 \\
			
			\bottomrule
		\end{tabularx}
	}
	\vspace{1pt}
	\caption{Ablation studies on Cityscapes validation set. Settings include two architecture variations: whether to utilise object masks (msk.), and whether to utilise the proposed instance affinity (aff.); one training option: whether to train end-to-end with the panoptic matching loss (e2e.); and two inference strategies: whether to directly take \texttt{argmax} (amx.) of the panoptic logits (which is either $\bm{\Psi}$ for \modelA{} and \modelCII{}, or $\bm{P}$ for \modelB{} and \modelDII{}) or use the heuristic merging strategy~\cite{kirillov2019panoptic} (heu.).
	}
	\label{tab:ablation_studies}
\end{table*}

%% file: tables/cityscapes_panoptic_results_halfcol.tex
\begin{table}[t]
	\centering
	\scalebox{0.83}{
		\begin{tabularx}{1.20\linewidth}{lYYYYYYYY}
			\toprule
			& \multicolumn{3}{c}{PQ} & SQ & RQ & IoU & $AP$ & $AP$ \\
			Method & all & th. & st. & all & all & all & mask & bbox \\
			\cmidrule(r){1-1} \cmidrule(r){2-4} \cmidrule(r){5-6} \cmidrule(l){7-7} \cmidrule(l){8-9}
			
			Li \etal~\cite{li2018weakly} & 53.8 & 42.5 & 62.1 & -- & -- & \textbf{79.8} & -- & -- \\
			
			DeeperLab~\cite{yang2019deeperlab} & 56.5 & -- & -- & -- & -- & -- & -- & -- \\
			
			SSAP~\cite{gao2019ssap} & 58.4 & \textbf{50.6} & -- & -- & -- & -- & -- & -- \\
			
			Ours (w/o mask) & \textbf{59.0} & 50.2 & \textbf{65.3} & \textbf{80.1} & \textbf{72.4} & 77.8 & -- & \textbf{38.1} \\
			
			\cmidrule(r){1-1} \cmidrule(r){2-4} \cmidrule(r){5-6} \cmidrule(l){7-7} \cmidrule(l){8-9}
			
			TASCNet~\cite{li2018learning}$\dagger$ & 55.9 & 50.5 & 59.8 & -- & -- & -- & -- & -- \\
			
			Attention~\cite{li2018attention}$\dagger$ & 56.4 & 52.7 & 59.0 & -- & -- & 73.6 & 33.6 & -- \\
			
			Pan. FPN~\cite{kirillov2019panoptic}$\dagger$ & 57.7 & 51.6 & 62.2 & -- & -- & 75.0 & 32.0 & -- \\
			
			UPSNet~\cite{xiong2019upsnet}$\dagger$ & 59.3 & 54.6 & 62.7 & 79.7 & 73.0 & 75.2 & 33.3 & \textbf{39.1} \\
			
			Pan. Deeplab~\cite{cheng2019panopticdeeplab}$\dagger$ & 59.7 & -- & -- & -- & -- & -- & -- & -- \\
			
			Seamless~\cite{Porzi2019seamless}$\dagger$ & 60.3 & \textbf{56.1} & 63.3 & -- & -- & 77.5 & 33.6 & -- \\
			
			Ours (w/ mask)$\dagger$ & \textbf{61.4} & 54.7 & \textbf{66.3} & \textbf{81.1} & \textbf{74.7} & \textbf{79.5} & \textbf{33.7} & 38.8 \\
			
			\bottomrule
		\end{tabularx}
	}
	\vspace{1pt}
	\caption{Panoptic segmentation results on Cityscapes \textit{val.} set.
	Models that run instance segmentation internally are marked with $\dagger$.
	Other than \cite{li2018weakly,yang2019deeperlab}, all works are ResNet-50~\cite{he2016deep} based.
	For fairness, we only include numbers obtained via \textit{single-scale} inference.
	}
	\label{tab:cityscapes_panoptic_comparison}
\end{table}

%% file: tables/coco_panoptic_results.tex
\begin{table}[t]
	\centering
	\scalebox{0.83}{
		\begin{tabularx}{1.20\linewidth}{lYYYYYYYY}
			\toprule
			& \multicolumn{3}{c}{PQ} & \multicolumn{1}{c}{SQ} & \multicolumn{1}{c}{RQ} & \multicolumn{1}{c}{IoU} & $AP$ & $AP$ \\
			Method & all & th. & st. & all & all & all & mask & bbox \\
			\cmidrule(r){1-1} \cmidrule(r){2-4} \cmidrule(r){5-6} \cmidrule(r){7-7} \cmidrule(l){8-9}
			JSIS-Net~\cite{de2018panoptic} & 26.9 & 29.3 & 23.3 & 72.4 & 35.7 & -- & -- & -- \\
			Pan.\ Deeplab~\cite{cheng2019panopticdeeplab} & 35.1 & -- & -- & -- & -- & -- & -- & -- \\
			Pan.\ FPN~\cite{kirillov2019panoptic} & 39.0 & 45.9 & 28.7 & -- & -- & -- & 33.3 & -- \\
			UPSNet~\cite{xiong2019upsnet} & 42.5 & \textbf{48.6} & 33.4 & 78.0 & 52.5 & \textbf{54.3} & 34.3 & 37.8 \\
			Ours (w/ mask) & \textbf{43.4} & \textbf{48.6} & \textbf{35.5} & \textbf{79.6} & \textbf{53.0} & 53.7 & \textbf{36.4} & \textbf{40.5} \\
			\bottomrule
		\end{tabularx}
	}
	\vspace{1pt}
	\caption{Panoptic segmentation results on COCO 2017 validation set. All methods are based on a ResNet-50 backbone.
	}
	\label{tab:coco_panoptic_comparison}
\end{table}

%% file: tables/cityscapes_test_set_results.tex
\begin{table}[t]
\centering
	\scalebox{0.71}{
	\begin{tabularx}{1.4\linewidth}{llYYYYYYYYY}
		\toprule
		& & \multicolumn{3}{c}{PQ} & \multicolumn{3}{c}{SQ} & \multicolumn{3}{c}{RQ} \\
		Method & Bb. & all & th. & st. & all & th. & st. & all & th. & st. \\
		
		\cmidrule(r){1-2} \cmidrule(r){3-5} \cmidrule(l){6-8} \cmidrule(l){9-11}
		
		P. Deeplab~\cite{cheng2019panopticdeeplab} & R-50 & 58.0 & -- & -- & -- & -- & -- & -- & -- & -- \\
		
		Ours (w/ mask) & R-50 & \textbf{61.0} & \textbf{52.7} & \textbf{67.1} & \textbf{81.4} & \textbf{79.6} & \textbf{82.8} & \textbf{73.9} & \textbf{66.2} & \textbf{79.6} \\
		
		\cmidrule(r){1-2} \cmidrule(r){3-5} \cmidrule(l){6-8} \cmidrule(l){9-11}
		
		Li \etal~\cite{li2018weakly,arnab2017pixelwise} & R-101 & 55.4 & 44.0 & 63.6 & 79.7 & 77.3 & 81.5 & 68.1 & 57.0 & 76.1 \\
		
		SSAP~\cite{gao2019ssap} & R-101 & 58.9 & 48.4 & 66.5 & \textbf{82.4} & \textbf{82.9} & 82.0 & 70.6 & 58.3 & 79.6 \\
		
		TASCNet~\cite{li2018learning}$\dagger$ & X-101 & 60.7 & 53.4 & 66.0 & 81.0 & 79.7 & 82.0 & 73.8 & 67.0 & 78.8 \\
		
		Ours (w/ mask)$\dagger$ & R-101 & \textbf{63.3} & \textbf{56.0} & \textbf{68.5} & \textbf{82.4} & 81.0 & \textbf{83.4} & \textbf{75.9} & \textbf{69.1} & \textbf{80.9} \\
		
		\bottomrule
	\end{tabularx}
	}
	\vspace{2pt}
	\caption{Performance on the Cityscapes test set. Models pretrained on the COCO dataset are marked with $\dagger$. Bb.: backbone, R: ResNet, X: ResNeXt.}
	\label{tab:cityscapes_test_set_results}
\end{table}

%% file: tables/coco_test_set_results.tex
\begin{table}[t]
\centering
	\scalebox{0.71}{
	\begin{tabularx}{1.4\linewidth}{llYYYYYYYYY}
		\toprule
		& & \multicolumn{3}{c}{PQ} & \multicolumn{3}{c}{SQ} & \multicolumn{3}{c}{RQ} \\
		Method & Bb. & all & th. & st. & all & th. & st. & all & th. & st. \\
		
		\cmidrule(r){1-2} \cmidrule(r){3-5} \cmidrule(l){6-8} \cmidrule(l){9-11}
		
		JSIS-Net~\cite{de2018panoptic} & R-50 & 27.2 & 29.6 & 23.4 & 71.9 & 71.6 & 72.3 & 35.9 & 39.4 & 30.6 \\
		
		P. Deeplab~\cite{cheng2019panopticdeeplab} & R-50 & 35.2 & -- & -- & -- & -- & -- & -- & -- & -- \\
		
		SSAP~\cite{gao2019ssap} & R-50 & 36.9 & 40.1 & 32.0 & \textbf{80.7} & \textbf{81.6} & \textbf{79.4} & 44.8 & 48.5 & 39.3 \\
		
		TASCNet~\cite{li2018learning} & R-50 & 40.7 & 47.0 & 31.0 & 78.5 & 80.6 & 75.3 & 50.1 & 57.1 & 39.6 \\
		
		Ours (w/ mask) & R-50 & \textbf{43.6} & \textbf{48.9} & \textbf{35.6} & 80.1 & 81.3 & 78.3 & \textbf{53.3} & \textbf{59.5} & \textbf{44.0} \\
		
		\cmidrule(r){1-2} \cmidrule(r){3-5} \cmidrule(l){6-8} \cmidrule(l){9-11}
		
		Attention~\cite{li2018attention} & X-152 & 46.5 & \textbf{55.9} & 32.5 & 81.0 & \textbf{83.7} & 77.0 & 56.1 & \textbf{66.3} & 40.7 \\
		
		UPSNet~\cite{xiong2019upsnet} & R-101 & 46.6 & 53.2 & 36.7 & 80.5 & 81.5 & 78.9 & 56.9 & 64.6 & 45.3 \\
		
		Ours (w/ mask) & R-101 & \textbf{47.2} & 53.5 & \textbf{37.7} & \textbf{81.1} & 82.3 & \textbf{79.2} & \textbf{57.2} & 64.3 & \textbf{46.3} \\
		
		\bottomrule
	\end{tabularx}
	}
	\vspace{2pt}
	\caption{Performance on the COCO \textit{test-dev} set. Bb.: backbone, R: ResNet, X: ResNeXt.}
	\label{tab:coco_test_set_results}
\end{table}

%% file: figures/vis_affinity.tex
\begin{figure*}[]
	\setlength{\tabcolsep}{2pt}
	\centering
	\begin{tabularx}{\textwidth}{YYYY}
		\global \def \VisScale{1}
		
		\global \def \im{munster_000091_000019}
		\global \def \affloc{449,97}
		\includegraphics[width=\VisScale\linewidth]{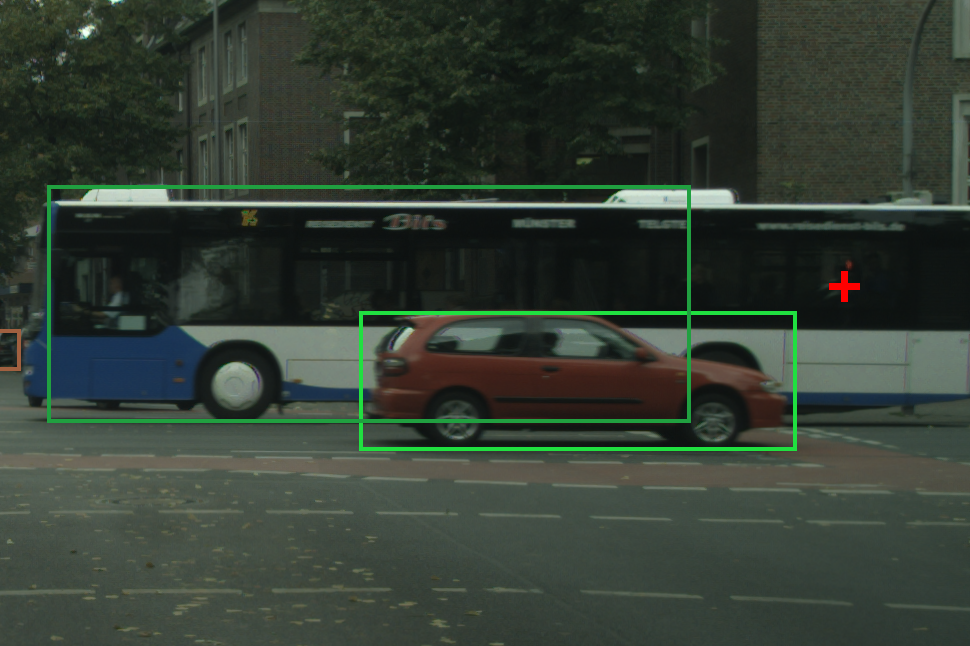}&
		\includegraphics[width=\VisScale\linewidth]{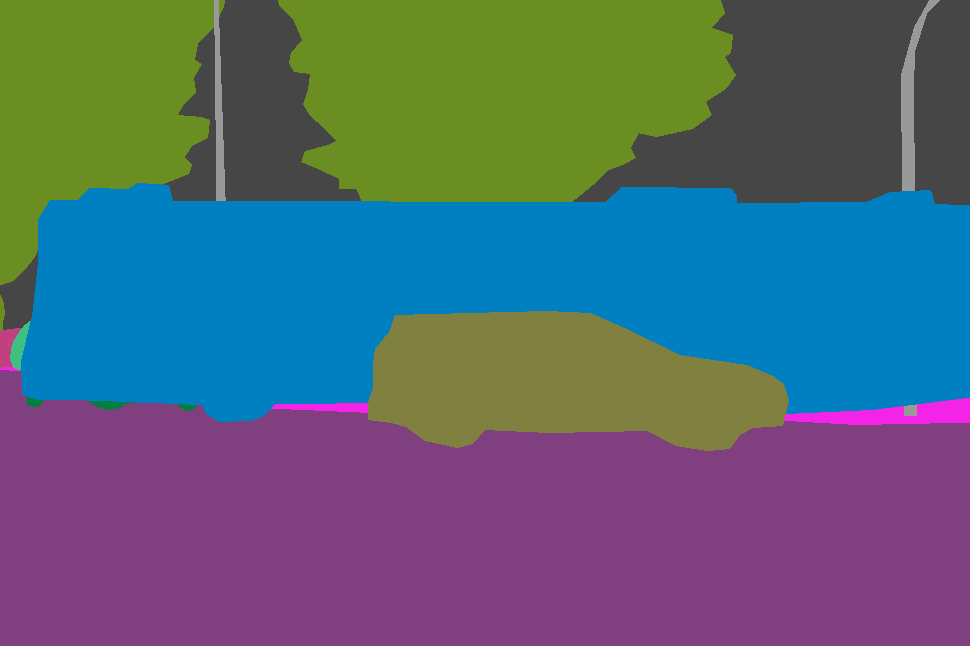}&
		\includegraphics[width=\VisScale\linewidth]{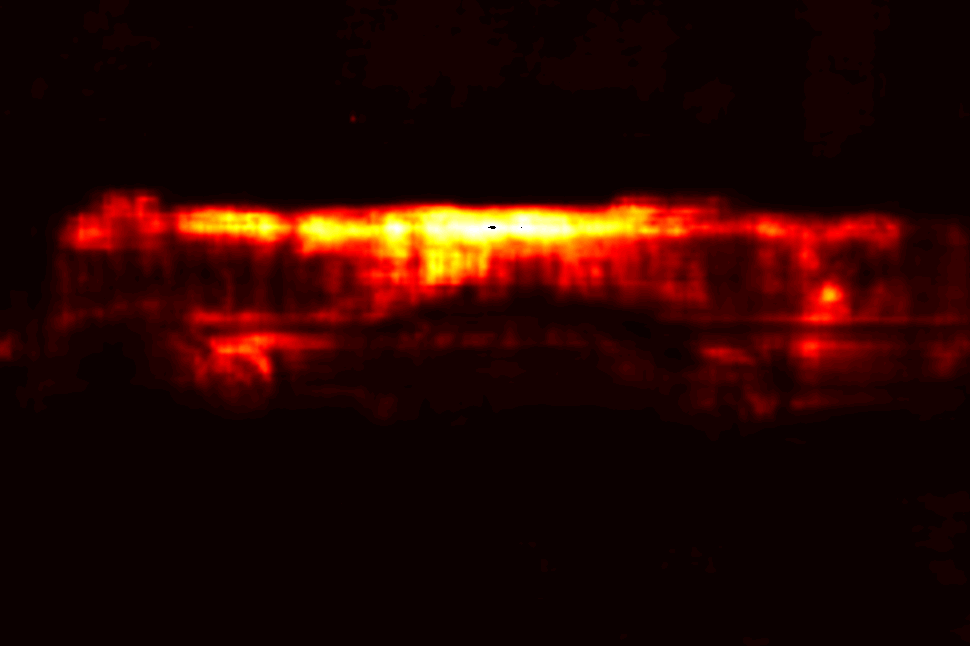}&
		\includegraphics[width=\VisScale\linewidth]{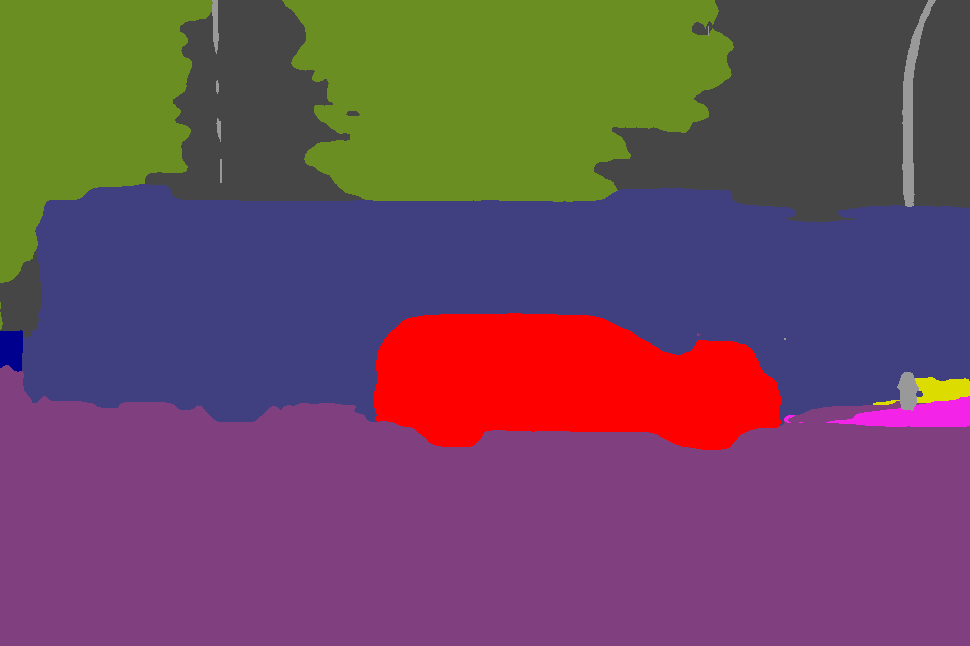}\\
		
		\global \def \im{munster_000056_000019}
		\global \def \affloc{94,79}
		\includegraphics[width=\VisScale\linewidth]{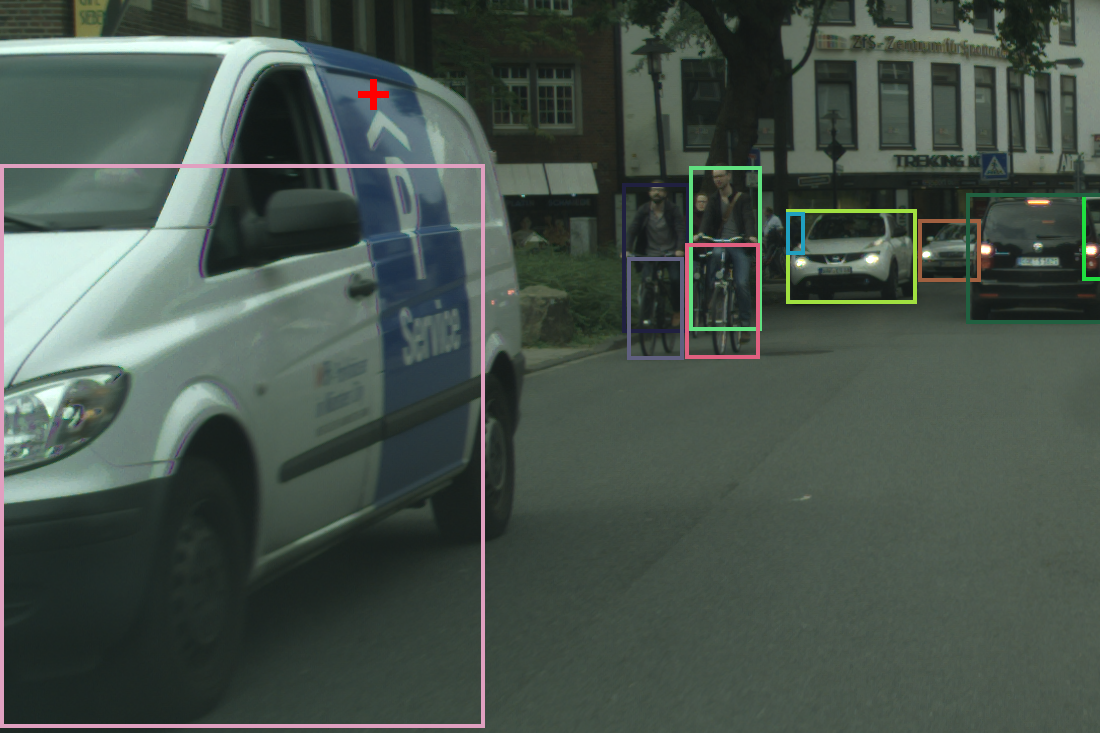}&
		\includegraphics[width=\VisScale\linewidth]{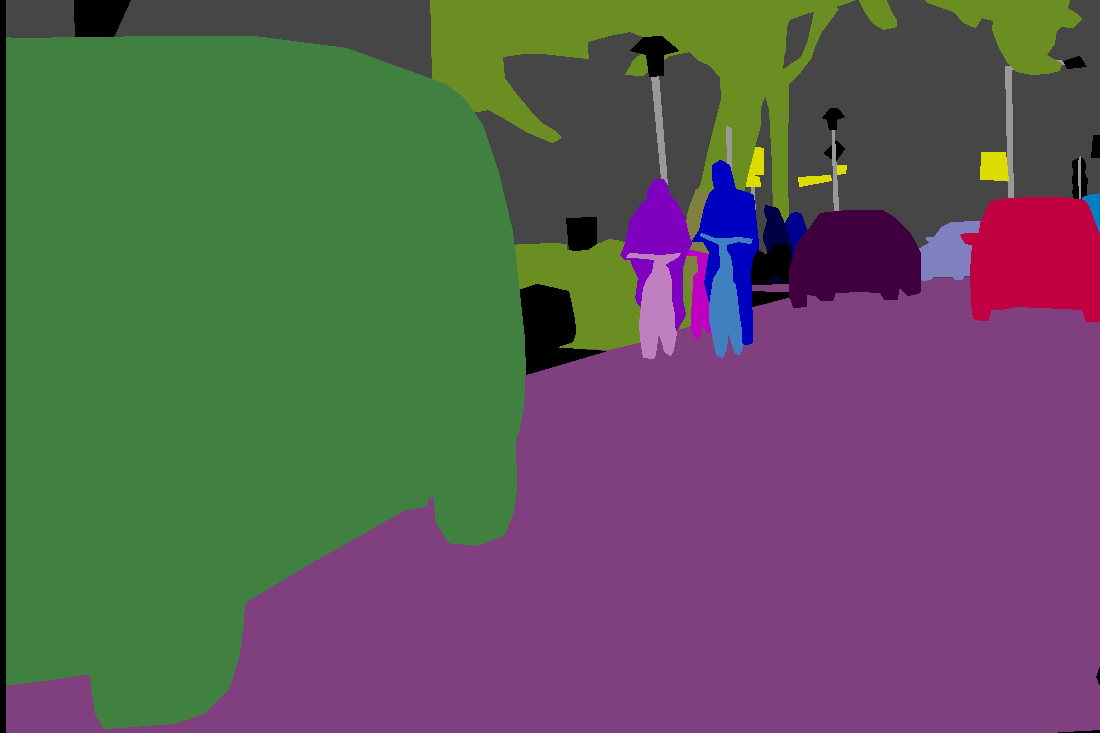}&
		\includegraphics[width=\VisScale\linewidth]{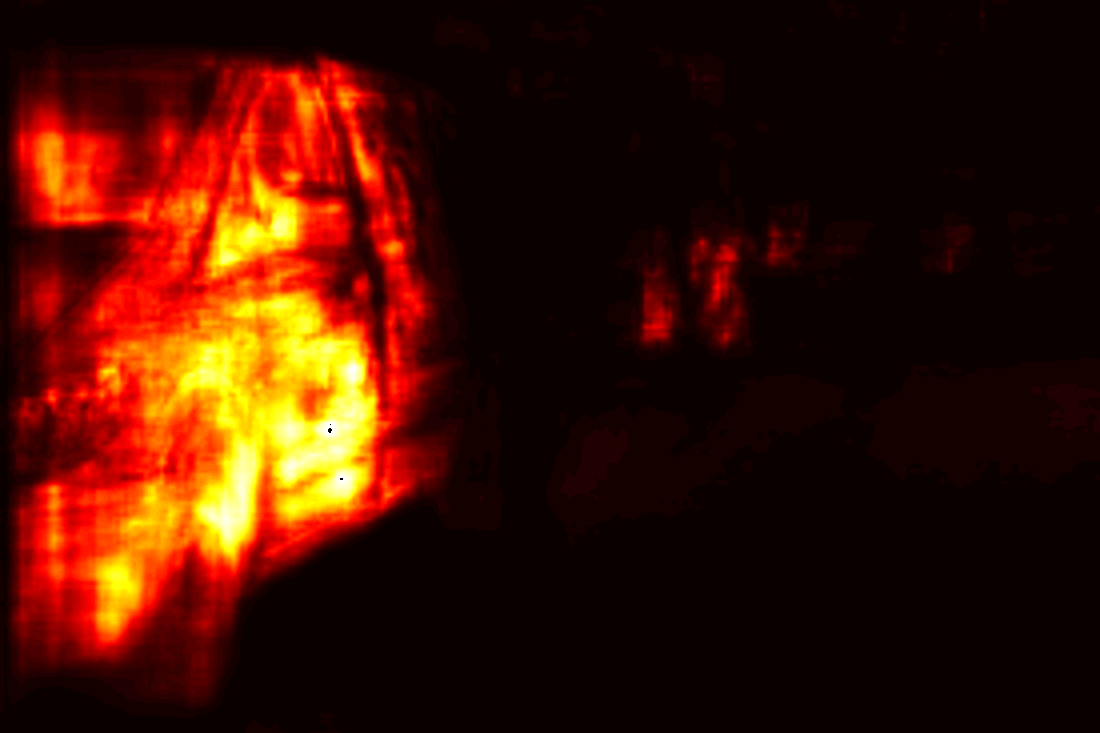}&
		\includegraphics[width=\VisScale\linewidth]{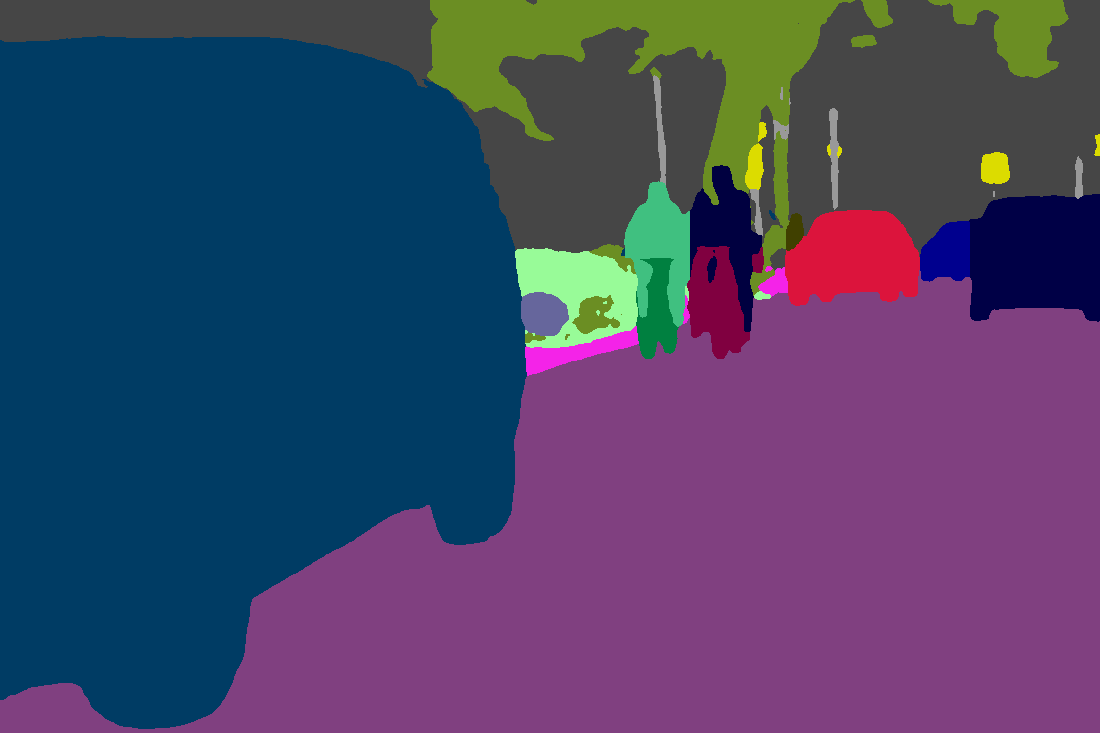}\\
	
		{\footnotesize{(a) Image with boxes}}&{\footnotesize{(b) Ground truth}}&{\footnotesize{(c) Instance affinities}}&{\footnotesize{(d) Panoptic prediction}}\\
	\end{tabularx}
	\vspace{2pt}
	\caption{Examples of predicted instance affinities.
	The instance affinities shown in (c) are for the cross-marked pixels in (a).
	Observe that the predicted bounding boxes (shown in (a)) for the bus in Row 1 and the frontal car in the Row 2 fail to enclose the full object. Rule-based fusion in~\cite{kirillov2018panoptic,kirillov2019panoptic} cannot recover from such localisation errors as their segments are constrained to pixels inside bounding boxes. In contrast, our model is able to still segment full objects by predicting strong affinities between the marked locations with rest of the instance.}
	\label{fig:vis_affinity}
\end{figure*}

%% file: figures/cityscapes_results.tex
\begin{figure*}[h!]
	\centering
	\begin{tabularx}{\textwidth}{YYYY}
		\global \def \VisScale{1.06}
		
		\includegraphics[width=\VisScale\linewidth]{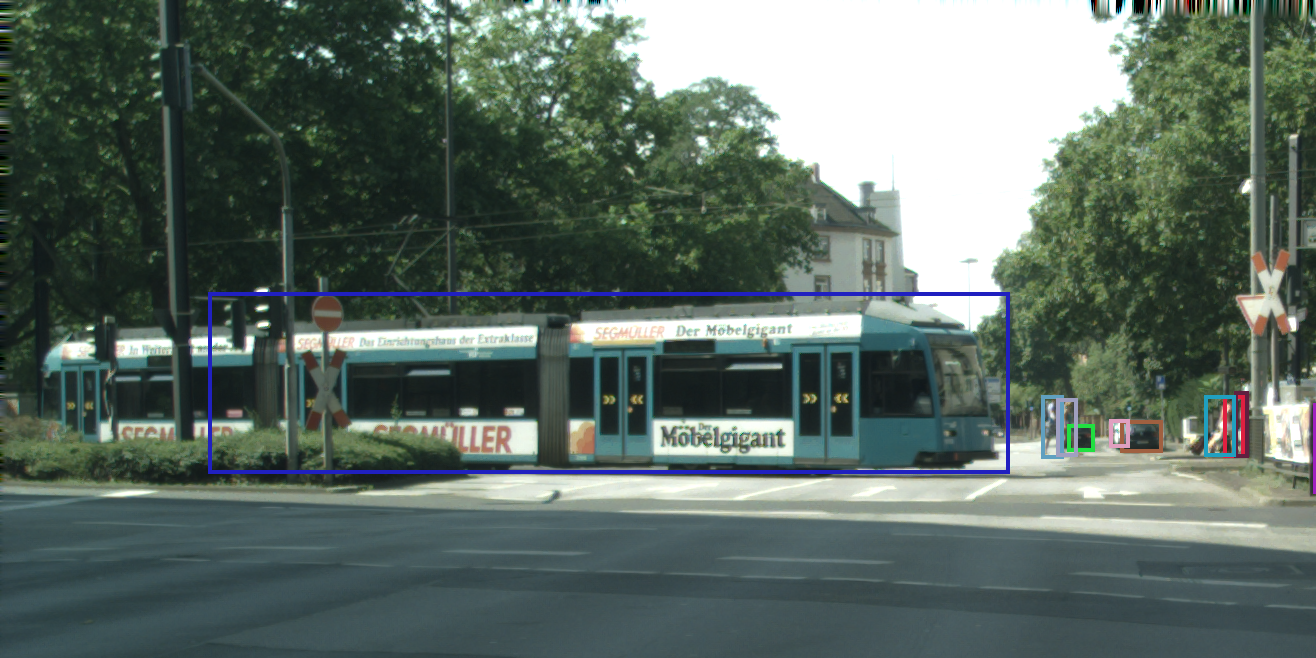}&
		\includegraphics[width=\VisScale\linewidth]{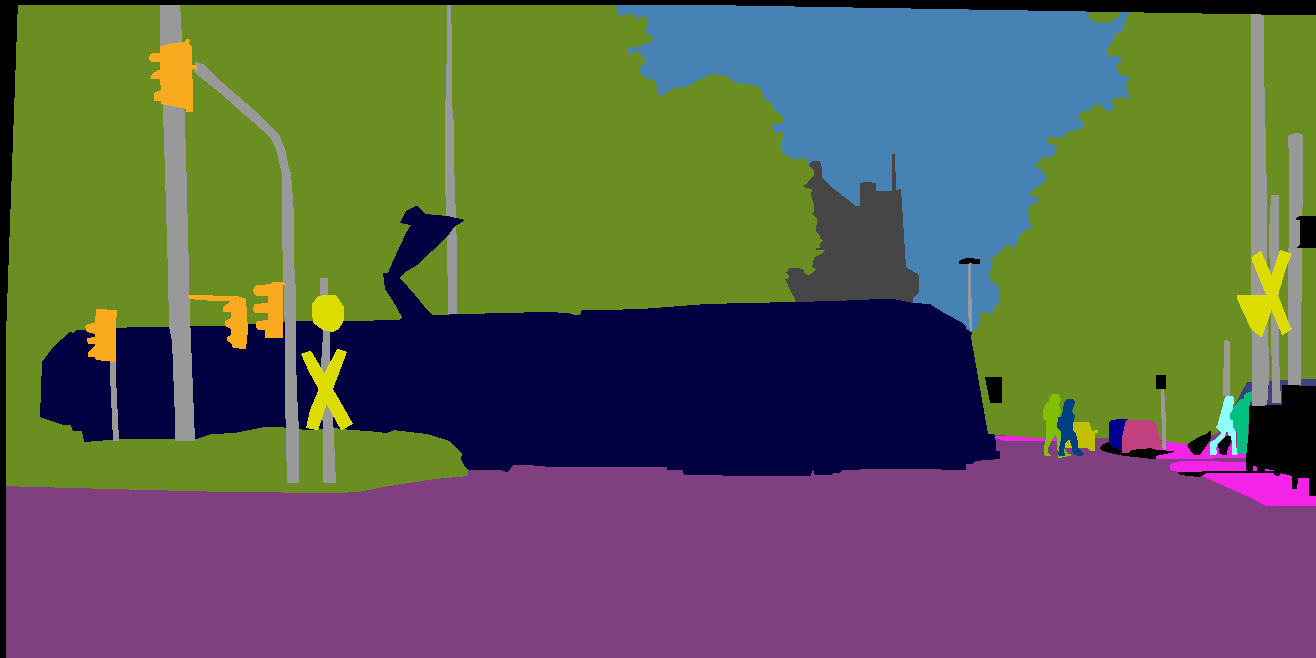}&
		\includegraphics[width=\VisScale\linewidth]{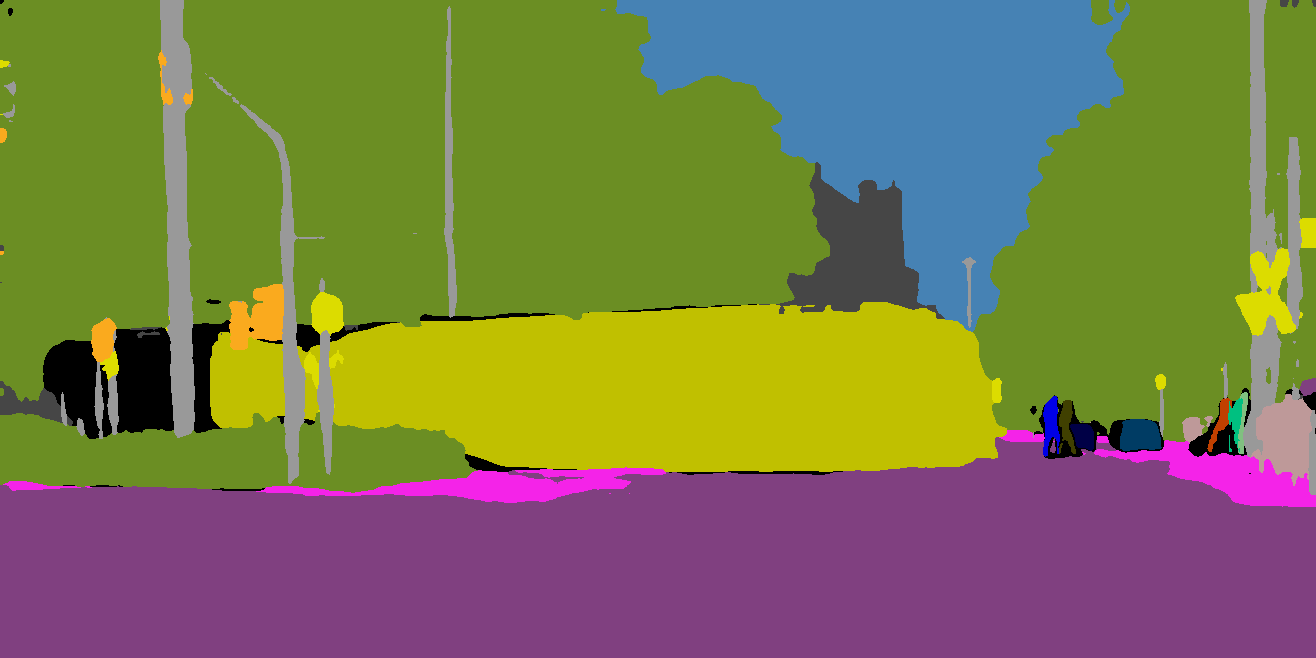}&
		\includegraphics[width=\VisScale\linewidth]{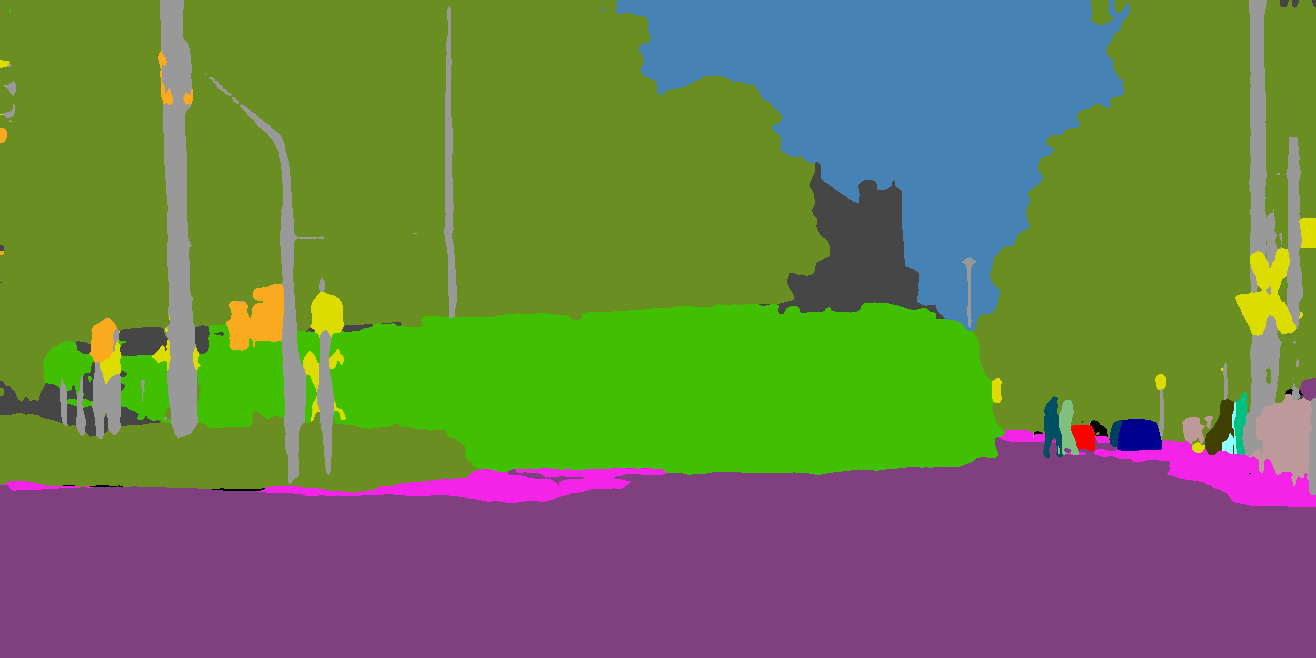}\\
		
		\includegraphics[width=\VisScale\linewidth]{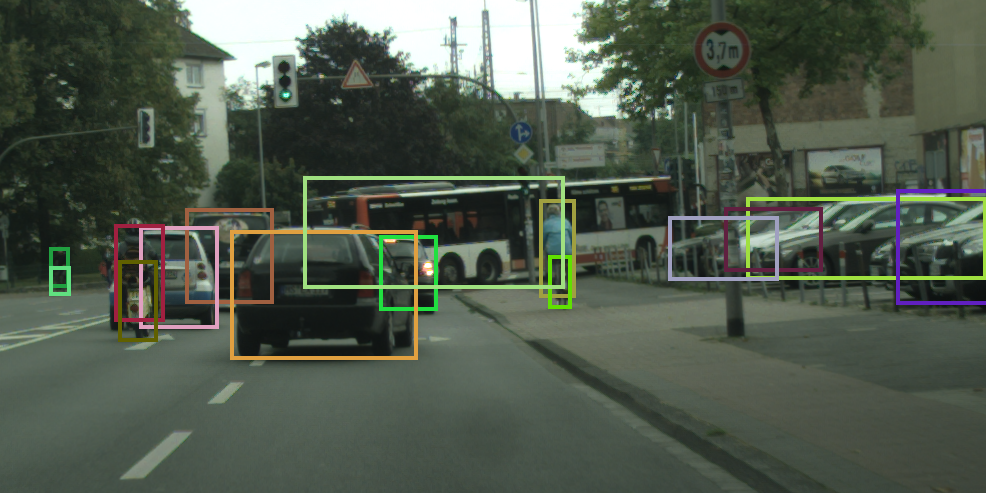}&
		\includegraphics[width=\VisScale\linewidth]{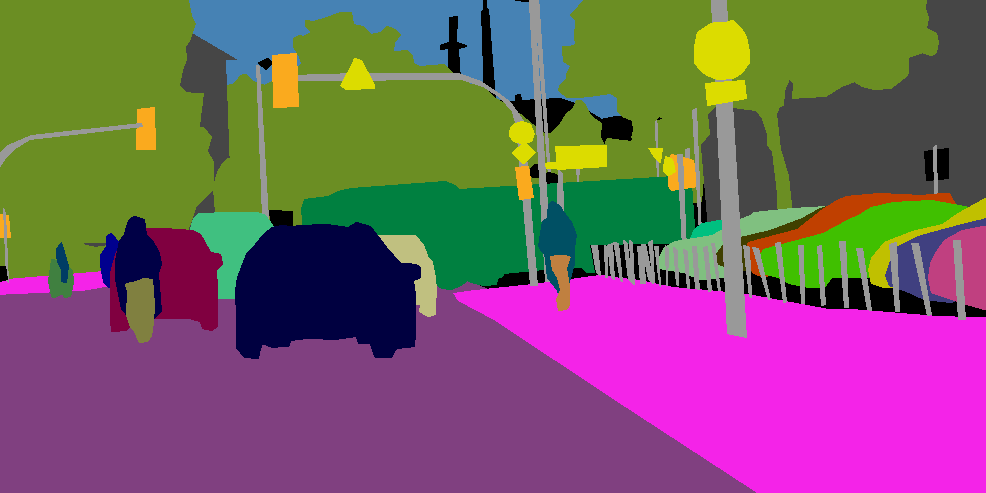}&
		\includegraphics[width=\VisScale\linewidth]{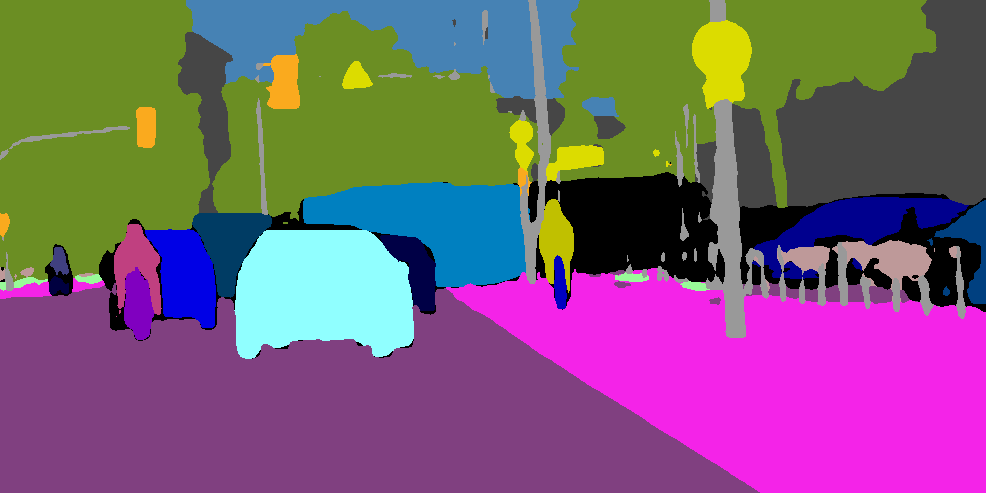}&
		\includegraphics[width=\VisScale\linewidth]{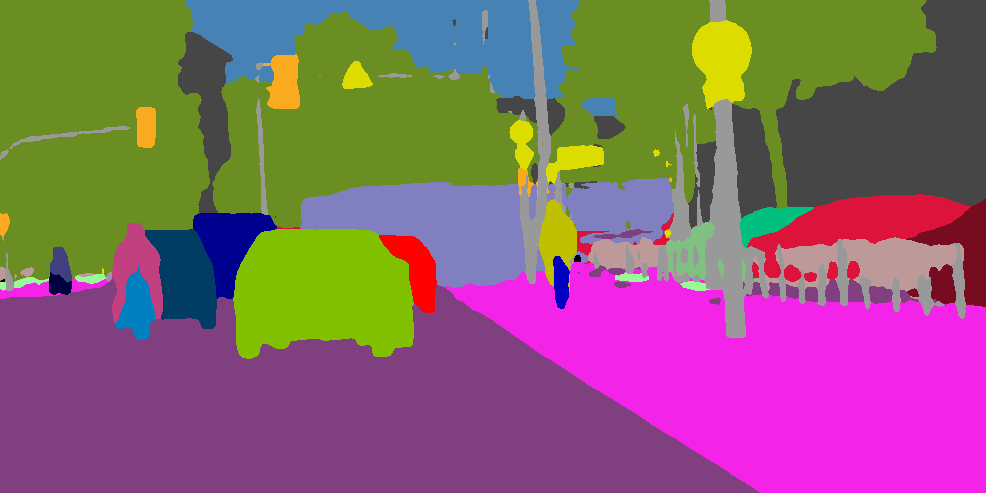}\\
		
		{\footnotesize{(a) Image with boxes}}&{\footnotesize{(b) Ground truth}}&{\footnotesize{(c) Heuristic merger~\cite{kirillov2019panoptic}}}&{\footnotesize{(d) Our prediction}}\\
	\end{tabularx}
	\vspace{2pt}
	\caption{Qualitative results. The input images are shown with the predicted bounding boxes overlaid above. In column (c), swathes of ``void'' region are clearly visible for pixels where assignment cannot be made by heuristics.
	In contrast, our panoptic segmentation results are robust to incoherence in segmentation and localisation cues, and can explain more pixels in an image.
	}
	\label{fig:qual}
\end{figure*}

%% file: contents/conclusion.tex
\section{Conclusion}
We have presented an end-to-end panoptic segmentation approach that exploits a novel pairwise instance affinity operation.
It is lightweight, learnt from data, and capable of modelling a dynamic number of instances.
By integrating information across the image in a differentiable manner, the instance affinity operation with the panoptic matching loss enables end-to-end training and heuristics-free inference, leading to improved qualities for panoptic segmentation.
Furthermore, our method bestows additional flexibility upon network design, allowing our model to perform well even if it only uses bounding boxes as localisation cues.

%% file: contents/acknowledgement.tex
\paragraph{Acknowledgements}
This work was supported by Huawei Technologies Co., Ltd., the ERC grant ERC-2012-AdG 321162-HELIOS, EPSRC grant Seebibyte EP/M013774/1 and EPSRC/MURI grant EP/N019474/1. We would also like to thank the Royal Academy of Engineering and FiveAI.

%% file: supplemental/supp_wrapper.tex
\appendix
\section*{Appendices}
\renewcommand{\thesubsection}{\Alph{subsection}}
\renewcommand{\thetable}{\Alph{table}}
\renewcommand{\thefigure}{\Alph{figure}}

\input{supplemental/architecture.tex}

\input{supplemental/implementation.tex}

\input{supplemental/evaluation.tex}

\input{supplemental/quantitative_results.tex}

\input{supplemental/qualitative_results.tex}

%% file: supplemental/architecture.tex
\subsection{Architecture and design}
\subsubsection{Semantic segmentation submodule}
\input{figures/model/semantic_seg_submod.tex}
Our semantic segmentation submodule is modified from~\cite{xiong2019upsnet}, by performing Group Normalisation~\cite{wu2018group} after each $3\times3$ convolution. We illustrate the pipeline in Fig.~\ref{fig:semantic_head}. Note that the architecture of the \textit{feature decoder} inside this submodule is also adopted by our dense instance affinity head to extract affinity features $\bm{Q}$. This submodule is supervised by a cross-entropy loss, unless otherwise stated.

\subsubsection{Object detection submodule}
In our experiments, we use the standard box head from Faster-RCNN~\cite{ren2015faster} and optionally the mask head from Mask-RCNN~\cite{he2017mask} for this submodule, following~\cite{xiong2019upsnet,kirillov2019panoptic}. For the mask head, we use the Lovasz Hinge loss to replace the binary cross entropy loss. Thanks to the modular design of our network, it is easy to substitute it with any other detector architecture.

\subsubsection{Dynamic potential head}
\input{tables/dyn_head_ablation.tex}
\input{tables/conf_mat.tex}
We refer to the design variant B and C presented in Sec.~\FAKEREF{3.4.1} (Fig.~\FAKEREF{4}). 
At first glance, variant B, which multiplies semantic segmentation probabilities $V_i(c_k)$ with mask scores $M_i(k)$, appears to be a more appropriate method than variant C which sums probabilities instead. The output of variant B is high only when both inputs are unanimously high. This can filter out spurious misclassifications from either input, and improve robustness towards false positive predictions. Indeed, on Cityscapes, we observe that variant B achieves a $1.1$ PQ lead over the variant C counterpart (first row of Table~\ref{tab:dyn_head_ablation}).

However, on COCO, we notice a high tendency for the semantic segmentation submodule to mistake ``things'' for ``stuff'' (Table.~\ref{tab:conf_mat}2). The multiplicative action of variant B can systematically and substantially weaken the panoptic logits for ``thing'' classes, relative to the unattenuated panoptic logits of ``stuff'' classes. This can be undesirable for models whose semantic segmentation submodule is already prone to misclassifying ``things'' as ``stuff''.
On the other hand, the opposite is true for variant C, as summation strengthens panoptic logits of ``things'' in comparison to unmodified ``stuff'' scores. 
This led us to use variant C for COCO, and we observe a 0.7 PQ improvement in comparison to B (second row of Table~\ref{tab:dyn_head_ablation}).

\subsubsection{Training with predicted detections}
\input{tables/ablation_train_with_gt_boxes.tex}
In contrast with the practice in~\cite{xiong2019upsnet}, we argue that, during training, the dynamic potential head should use predicted detections instead of ground truth ones to construct its output $\bm{\Psi}$. This allows the network to learn from realistic examples, and build up its robustness towards imperfections in detection localisation and scoring. To test our hypothesis, we carried out an ablation study on Cityscapes using our mask-free model. When training with ground truth boxes, a uniform score of $1.0$ is used for their confidence scores. Results are shown in Table~\ref{tab:train_with_predicted_boxes}. As expected, training with predicted detections yields performance improvements across all panoptic metrics, including a $0.4$ increase in PQ. A large boost in observed for $AP_{box}$ ($+1.3$), because training with predicted boxes allows gradients from the panoptic segmentation submodule to flow to the object detection submodule, giving it more fine-grained supervision. IoU has not changed, as this ablation setting does not affect the semantic segmentation module.

%% file: figures/model/semantic_seg_submod.tex
\begin{figure}
	\includegraphics[width=\linewidth]{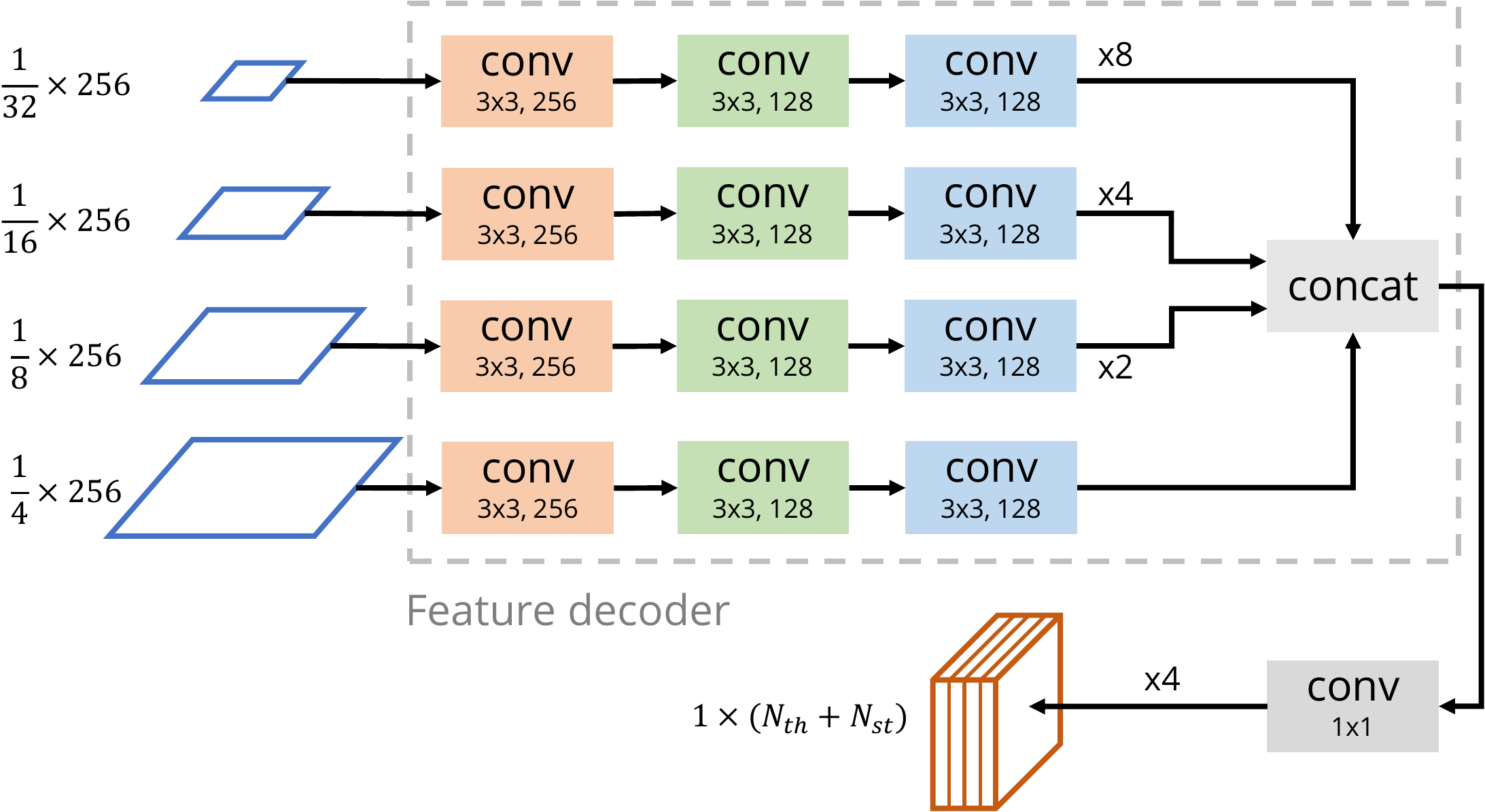}
	\vspace{2pt}
	\caption{Semantic segmentation submodule. Each $3\times3$ convolution block consists of a deformable convolution (with the indicated number of output channels), a Group Normalisation operation, and a ReLU activation. Weights are \textbf{shared} across $3\times3$ convolution blocks with the same colour code.}
	\label{fig:semantic_head}
\end{figure}

%% file: tables/dyn_head_ablation.tex
\begin{table}[t]
	\centering
	\begin{tabularx}{\linewidth}{lYYYYYY}
		\toprule
		& \multicolumn{3}{c}{Variant B} & \multicolumn{3}{c}{Variant C} \\
		Dataset & PQ & SQ & RQ & PQ & SQ & RQ \\
		\cmidrule(r){1-1} \cmidrule(r){2-4} \cmidrule(r){5-7}
		Cityscapes & \textbf{61.4} & \textbf{81.8} & \textbf{74.7} & 60.3 & 80.8 & 73.5 \\
		COCO & 42.7 & 79.4 & 52.2 & \textbf{43.4} & \textbf{79.6} & \textbf{53.0} \\
		\bottomrule
	\end{tabularx}
	\vspace{2pt}
	\caption{Ablation study on two design variants for the dynamic potential head. On Cityscapes, variant B outperforms variant C, whereas on COCO, variant C achieves higher accuracies.}
	\label{tab:dyn_head_ablation}
\end{table}

%% file: tables/conf_mat.tex
\begin{table}[t]
	\centering
	\begin{tabularx}{0.48\linewidth}{llYY}
		\toprule
		& & \multicolumn{2}{c}{Classified as} \\
		& & th. & st. \\
		\cmidrule(r){1-4}
		\multirow{2}{*}{GT} & th. & 95.1 & 4.9 \\
		& st. & 0.0 & 100.0 \\
		\bottomrule
		\\[-1em]
		\multicolumn{4}{c}{(1) Cityscapes}
	\end{tabularx}
	\begin{tabularx}{0.48\linewidth}{llYY}
		\toprule
		& & \multicolumn{2}{c}{Classified as} \\
		& & th. & st. \\
		\cmidrule(r){1-4}
		\multirow{2}{*}{GT} & th. & 90.1 & 9.9 \\
		& st. & 4.8 & 95.2 \\
		\bottomrule
		\\[-1em]
		\multicolumn{4}{c}{(2) COCO}
	\end{tabularx}
	\vspace{2pt}
	\caption{Confusion matrices between ``thing'' and ``stuff'' for semantic segmentation submodule outputs on Cityscapes and COCO validation sets. All numbers are percentages, normalised row-wise.}
	\label{tab:conf_mat}
\end{table}

%% file: tables/ablation_train_with_gt_boxes.tex
\begin{table}
	\begin{tabularx}{\linewidth}{lYYYYY}
		\toprule
		Dets. for training & PQ & SQ & RQ & IoU & $AP_{box}$ \\
		\cmidrule{1-6}
		Ground truths & 58.6 & 80.0 & 72.0 & 77.8 & 36.8 \\
		Predictions & \textbf{59.0} & \textbf{80.1} & \textbf{72.4} & 77.8 & \textbf{38.1} \\
		\bottomrule
	\end{tabularx}
	\vspace{2pt}
	\caption{Comparison between two different training strategies. The top row uses ground truth detections to train the panoptic segmentation submodule, whereas the bottom row uses the ones predicted by the network on-the-fly. Results are reported on the Cityscapes validation set.}
	\label{tab:train_with_predicted_boxes}
\end{table}

%% file: supplemental/implementation.tex
\subsection{Implementation details}

\paragraph{Cityscapes training.}
We run our experiments on four V100-32GB GPUs. This allows us to load each GPU with eight image crops and obtain an effective batch size of 32.
The large number of crops per GPU enables us to use a Lovasz Softmax loss~\cite{berman2018lovasz} instead of a cross entropy loss for supervising semantic segmentation, which we found to be effective.
Following~\cite{kirillov2019panoptic}, we use a base learning rate of $0.01$, a weight decay of $0.0001$, and train for a total of $65$k iterations.
The learning rate is reduced by $10$ folds after the first $40$k iterations, and once more after another $15$k iterations.
Additionally, we adopt a ``warm-up'' period at the start of training -- linearly increasing the learning rate from a third of the base rate to the full rate in 500 iterations, which helps stabilise the training.

We augment input images on-the-fly during training to reduce the network's tendency to overfit. Our augmentation pipeline resizes the input image by a random factor between $0.5$ and $2$, takes a random $512 \times 1024$ crop, and applies a horizontal flip with $50$\% chance. On top of these techqniues, we also apply image relighting, randomly adjusting the brightness, contrast, hue, and saturation of the image by a small amount, as used in \cite{kirillov2019panoptic}.

\paragraph{COCO training.} On COCO, as the dataset is larger than Cityscapes, less overfitting is observed. Therefore, in terms of data augmentation techniques, we only apply resizing where the shorter size is resized to $800$ and the longer size is kept under $1333$, and random horizontal flipping with $0.5$ probability.

\paragraph{Miscellaneous.} We use ImageNet pretrained ResNet-50 to initialise all experiments. The batch normalisation statistics are kept unchanged, though further performance gains are likely if they are finetuned on the target dataset. When a normalisation step is used in either the semantic or panoptic submodules, we use the Group Normalisation operation~\cite{wu2018group}, as it is less sensitive to batch sizes.

\paragraph{Inference.}
We conduct single-scale inference for all experiments, letting the network process and make predictions on full-resolution images in a single forward pass. Note that only detection predictions whose confidence scores are more than a threshold are fed into the dynamic potential head during inference, to minimise unnecessary computation. This cut-off is $0.5$ and $0.75$ for Cityscapes and COCO respectively.

%% file: supplemental/evaluation.tex
\subsection{Evaluation of ``stuff''}

The PQ metrics effectively treats ``stuff'' classes as image-wide instances -- making all ``stuff'' segments undergo the same matching procedure with ground truth segments as ``thing'' segments. While this approach has its merits including a unified evaluation logic and a simplified PQ implementation, it should be noted that matching ``stuff'' predictions to ground truth is not strictly necessary, since at most one ``stuff'' instance for each ``stuff'' class is present per image.

Furthermore, this approach towards ``stuff'' is neither robust nor fair as a measure for ``stuff'' segmentation quality, and arguably encourages post-processing of panoptic predictions.
Under the PQ formulation, misclassifying even a single pixel into a ``stuff' class absent in the ground truth will increment false positive detections by one, and such mistakes -- exacerbated by the relatively small number of ground truth ``stuff'' segments in a dataset  -- attract a large penalty on the ``stuff'' RQ, even though the practical impact on perceptual quality is minimal.
This also contrasts in spirit with the mean IoU metric widely adopted to measure semantic segmentation quality, as the mean IoU accumulates intersection and union counts over the whole dataset and is minimally affected by individual pixels.

On the other hand, CNN-based semantic segmentation models are typically prone to produce spurious misclassifications, as they usually do not explicitly enforce smoothness.
As a result, recent panoptic segmentation works~\cite{kirillov2018panoptic,li2018attention,li2018learning,kirillov2019panoptic,xiong2019upsnet,yang2019deeperlab} collectively resort to setting small ``stuff'' segments to ``void'' in the final panoptic segmentation.
Therefore, to foster meaningful comparison with other state-of-the-art panoptic segmentation approaches,
unless specified otherwise,
we also carry out this strategy as part of evaluation.

\paragraph{Effects of trimming small stuff segments on evaluation metrics. }
\input{tables/pq_vs_iou_stuff.tex}
On Cityscapes validation set, we test our full model, our re-implemented Panoptic FPN~\cite{kirillov2019panoptic}, and the released UPSNet model~\cite{xiong2019upsnet} with and without trimming off small ``stuff'' regions, to quantitatively assess the impact of this step on state-of-the-art models. The findings are reported in Table~\ref{tab:pq_vs_iou_stuff}.

The results show that PQ and RQ are very sensitive to such operations, as removing small stuff segments consistently results in an increase of approximately 2 points for ``stuff'' PQ, and 2.5 points for ``stuff'' RQ. This can be largely attributed to the reduced number of false positive stuff segments. On the other hand, the ``stuff'' IoU metric is insensitive to such modifications, as in all three cases, it suffers a slight decrease of 0.1 or 0.2 points. This prompts us to believe that ``stuff'' IoU is a better metric for capturing ``stuff'' segmentation quality than the ``thing''-centric PQ family.

%% file: tables/pq_vs_iou_stuff.tex
\begin{table}[t]
	\centering
	\caption{Comparison of various evaluation metrics for ``stuff'', before and after small stuff areas are set to ``void'' on Cityscapes validation set. Note that the IoU$^{st}$ here is computed from the final panoptic segmentation, by combining instances of the same semantic class. This is different from the IoU metrics reported in Table~\FAKEREF{1} and~\FAKEREF{3}, which measure the quality of the semantic segmentation input to the heuristic merger / our panoptic segmentation submodule.}
	\label{tab:pq_vs_iou_stuff}
	\vspace{5pt}
	\begin{tabularx}{\linewidth}{lYYYYY}
		\toprule
		Model & Trim stuff & PQ$^{st}$ & SQ$^{st}$ & RQ$^{st}$ & IoU$^{st}$ \\
		
		\cmidrule(r){1-6}
		
		Pan.~FPN~\cite{kirillov2019panoptic}* & \xmark &  59.9 & 79.3 & 72.9 & 74.7 \\
		Pan.~FPN~\cite{kirillov2019panoptic}* & \cmark & 62.0 & 79.6 & 75.5 & 74.5 \\
		& & +2.1 & +0.3 & +2.6 & -0.2 \\
		
		\cmidrule(r){1-6}
		
		UPSNet~\cite{xiong2019upsnet}$\dagger$ & \xmark & 60.5 & 79.8 & 73.6 & 75.8 \\
		UPSNet~\cite{xiong2019upsnet}$\dagger$ & \cmark & 62.8 & 80.0 & 76.3 & 75.7 \\
		& & +2.3 & +0.2 & +2.7 & -0.1 \\
		
		\cmidrule(r){1-6}
		
		Ours & \xmark & 64.2 & 81.4 & 77.1 & 78.3 \\
		Ours & \cmark & 66.3 & 81.8 & 79.4 & 78.2 \\
		& & +2.1 & +0.4 & +2.3 & -0.1 \\
		
		\bottomrule
	\end{tabularx}
	\vspace{1pt}
	\footnotesize{* Results obtained from our re-implementation of Panoptic FPN. \\$\dagger$ Results obtained by running the public inference script of~\cite{xiong2019upsnet}.}
\end{table}

%% file: supplemental/quantitative_results.tex
\subsection{Detailed validation set results}
\input{tables/full_val_results.tex}
We report the detailed results of our models on the Cityscapes and COCO validation sets in Table~\ref{tab:full_val_results}. In addition to the metrics reported in the main paper, this table also includes breakdowns of SQ and RQ by ``stuff'' and ``thing''.

%% file: tables/full_val_results.tex
\begin{table*}[t]
	\centering
	\scalebox{0.91}{
		\begin{tabularx}{1.10\linewidth}{llYYYYYYYYYYYYYYYY}
			\toprule
			& & \multicolumn{3}{c}{PQ} & \multicolumn{3}{c}{SQ} & \multicolumn{3}{c}{RQ} & \multicolumn{3}{c}{IoU} & $AP$ & $AP$\\
			Dataset & Method & all & th. & st. & all & th. & st. & all & th. & st. & all & th. & st. & mask & box \\
			\cmidrule(r){1-2} \cmidrule(r){3-5} \cmidrule(r){6-8} \cmidrule(l){9-11} \cmidrule(l){12-14} \cmidrule(l){15-16}
			
			Cityscapes & Ours (w/o mask) & 59.0 & 50.2 & 65.3 & 80.1 & 78.4 & 81.2 & 72.4 & 63.9 & 78.6 & 77.8 & 78.7 & 77.2 & -- & 38.1\\
			
			Cityscapes & Ours (w/ mask) & 61.4 & 54.7 & 66.3 & 81.1 & 80.0 & 81.8 & 74.7 & 68.2 & 79.4 & 79.5 & 81.0 & 78.4 & 33.7 & 38.8\\
			
			\cmidrule(r){1-2} \cmidrule(r){3-5} \cmidrule(r){6-8} \cmidrule(l){9-11} \cmidrule(l){12-14} \cmidrule(l){15-16}
			
			COCO & Ours (w/ mask) & 43.4 & 48.6 & 35.5 & 79.6 & 80.0 & 78.9 & 53.0 & 59.2 & 43.8 & 53.7 & 60.4 & 43.6 & 36.4 & 40.5 \\
			
			\bottomrule
		\end{tabularx}
	}
	\vspace{1pt}
	\caption{Full panoptic segmentation results on Cityscapes validation set and COCO validation set.
	All models are ResNet-50 based, and tested with a \textit{single-scale} inference scheme, without test-time augmentation.
	}
	\label{tab:full_val_results}
\end{table*}

%% file: supplemental/qualitative_results.tex
\subsection{Visualisation of learnt instance affinities}
\input{figures/supp_vis_affinity.tex}
Additional visualisations of some predicted instance affinities are provided in Fig.~\ref{fig:vis_affinity_supp}. Note that these instance affinities are extracted from our mask-free model. Interestingly, the model has learnt to resolve cars regions covered by multiple car bounding boxes -- a problem difficult for methods only using boxes as localisation cues -- by creating strong instance affinities to the bottoms and tyres of cars. The model has found that these regions of cars are normally not covered by multiple bounding boxes, and therefore it is most helpful for instance discrimination by associating uncertain pixels with these regions. 

\subsection{Qualitative results}
\input{figures/supp_cityscapes_results.tex}
\input{figures/supp_coco_results.tex}
We show more qualitative results in Fig.~\ref{fig:supp_cityscapes} and~\ref{fig:supp_coco}, and comparisons to previous state-of-the-art methods~\cite{kirillov2019panoptic,xiong2019upsnet}.

%% file: figures/supp_vis_affinity.tex
\begin{figure*}
	\setlength{\tabcolsep}{2pt}
	\centering
	\begin{tabularx}{\textwidth}{YYYY}
		\global \def \VisScale{1}
		
		\global \def \im{frankfurt_000001_001464}
		\global \def \affloc{473,81}
		\includegraphics[width=\VisScale\linewidth]{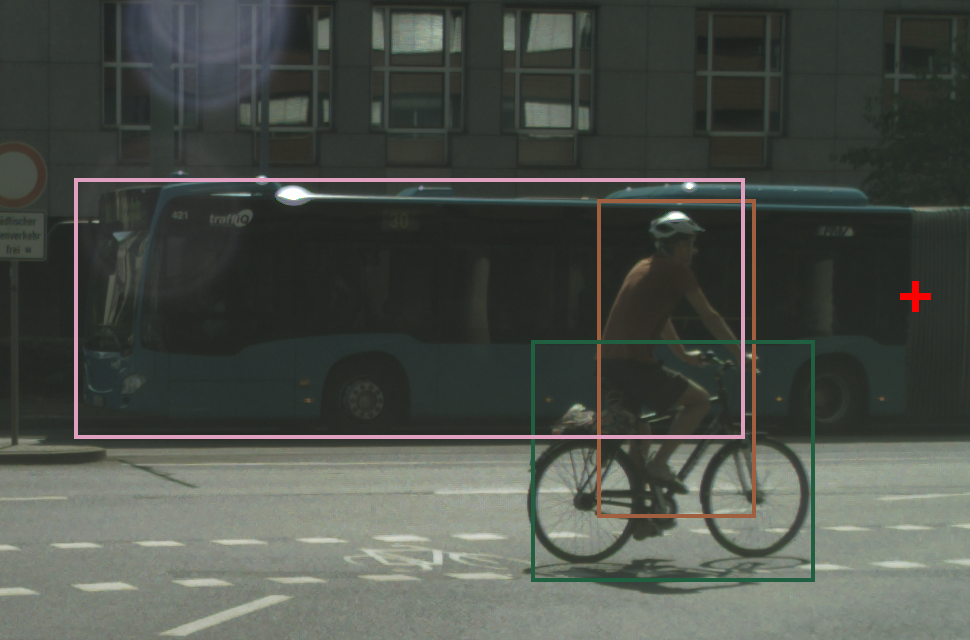}&
		\includegraphics[width=\VisScale\linewidth]{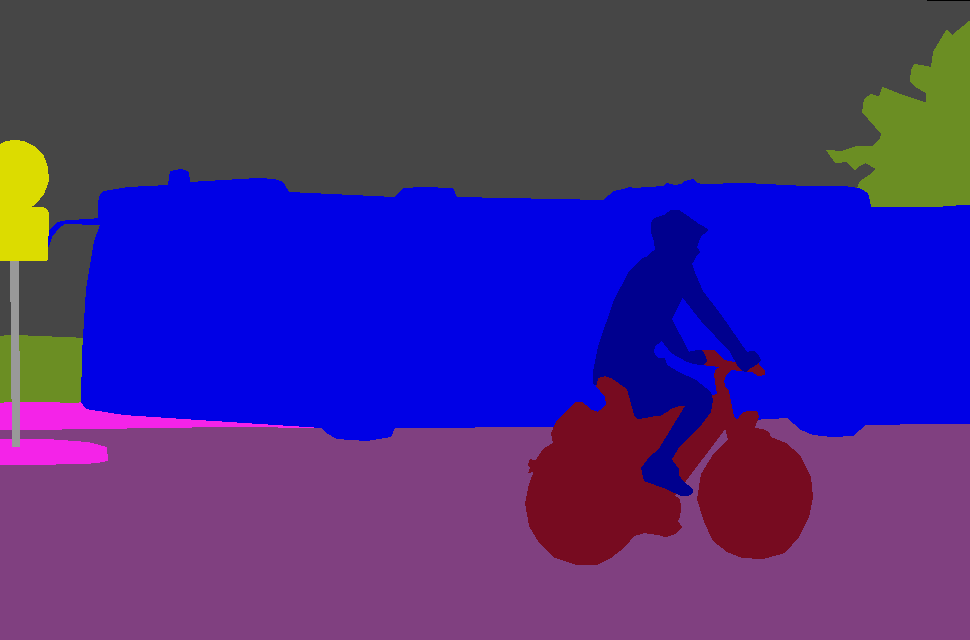}&
		\includegraphics[width=\VisScale\linewidth]{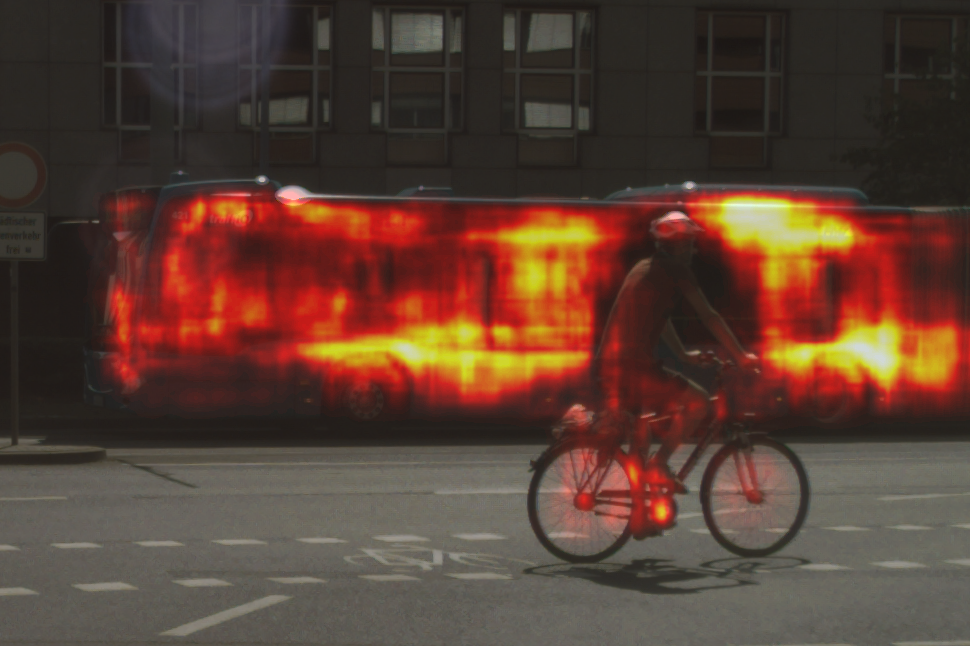}&
		\includegraphics[width=\VisScale\linewidth]{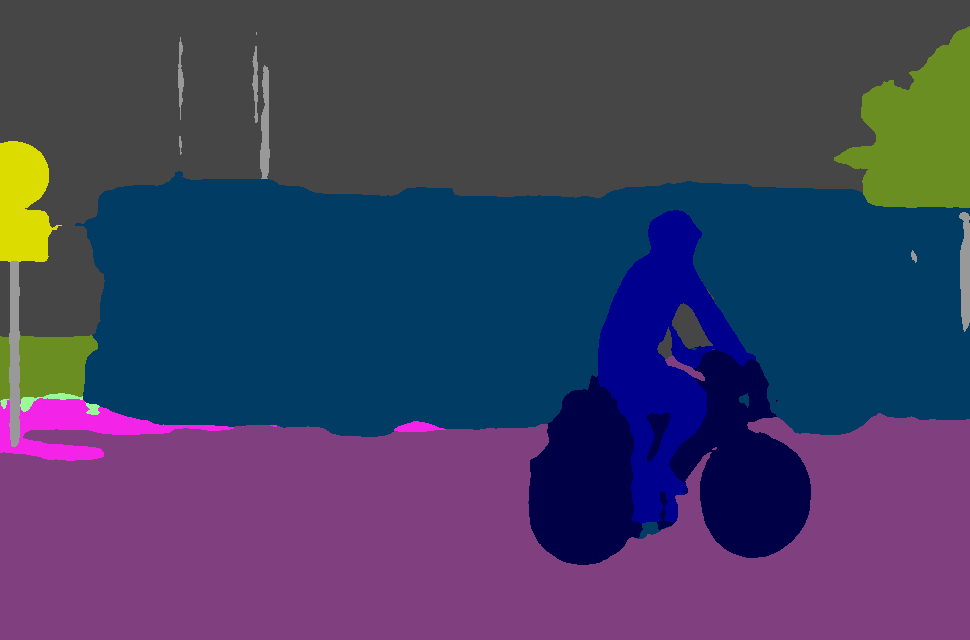}\\
		
		\global \def \im{frankfurt_000000_005898}
		\includegraphics[width=\VisScale\linewidth]{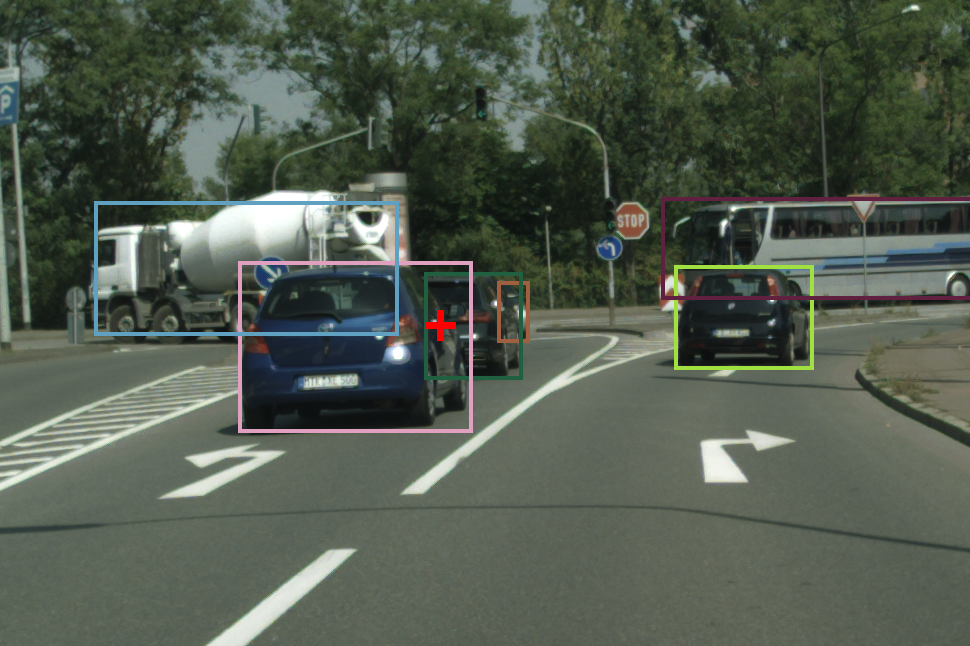}&
		\includegraphics[width=\VisScale\linewidth]{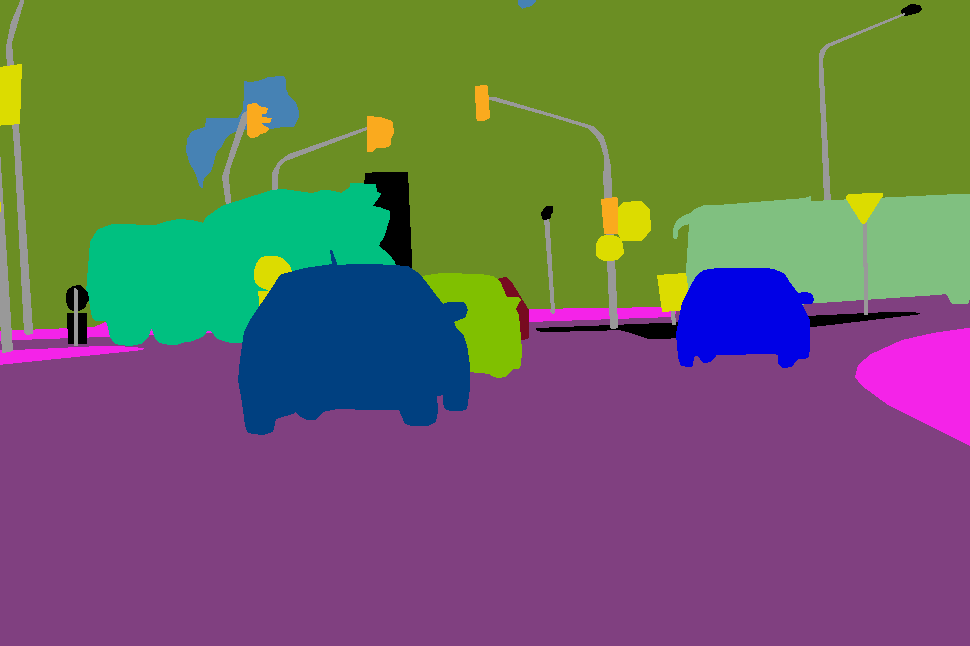}&
		\includegraphics[width=\VisScale\linewidth]{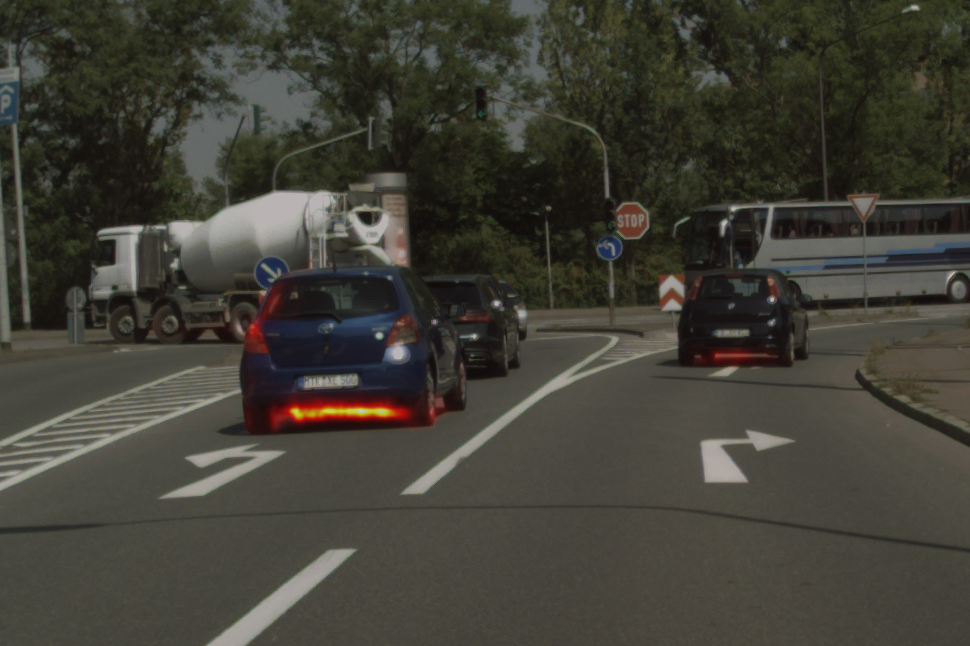}&
		\includegraphics[width=\VisScale\linewidth]{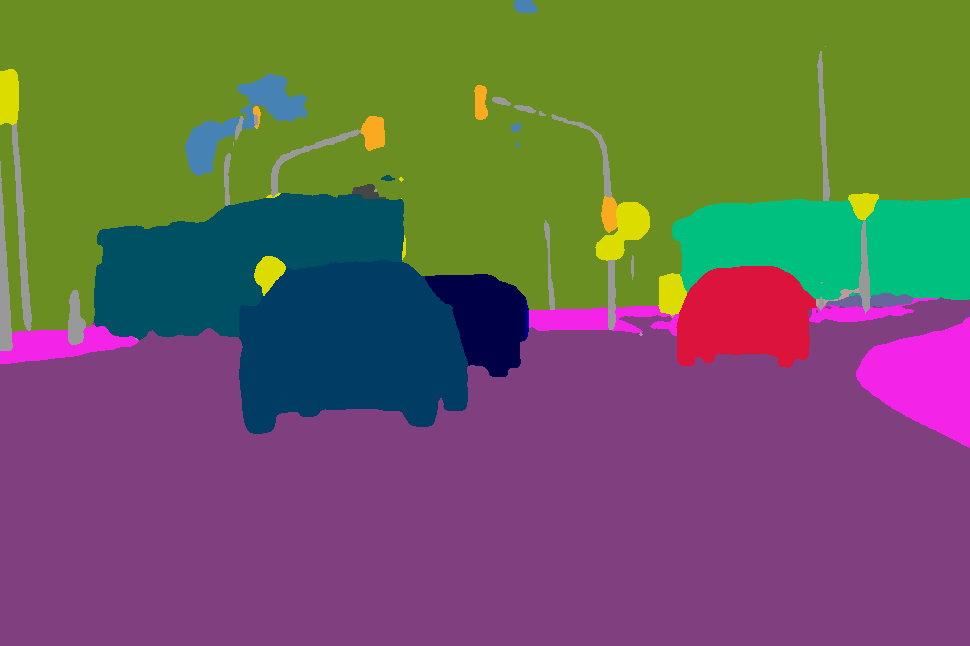}\\
		
		\global \def \im{frankfurt_000001_015328}
		\includegraphics[width=\VisScale\linewidth]{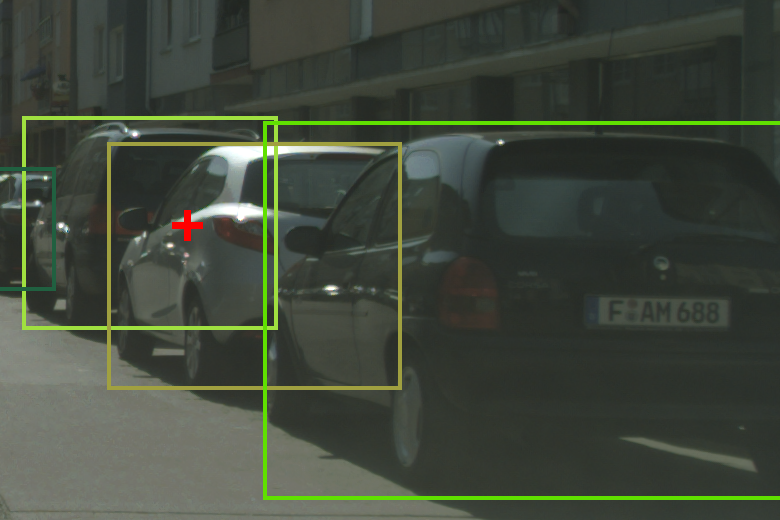}&
		\includegraphics[width=\VisScale\linewidth]{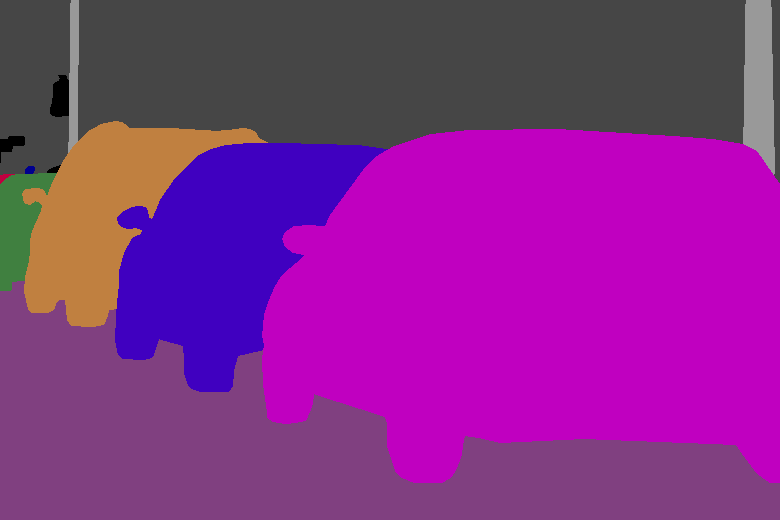}&
		\includegraphics[width=\VisScale\linewidth]{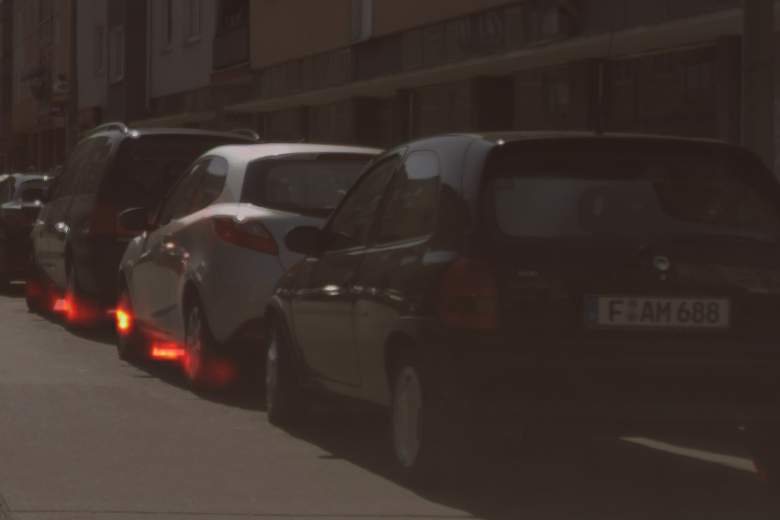}&
		\includegraphics[width=\VisScale\linewidth]{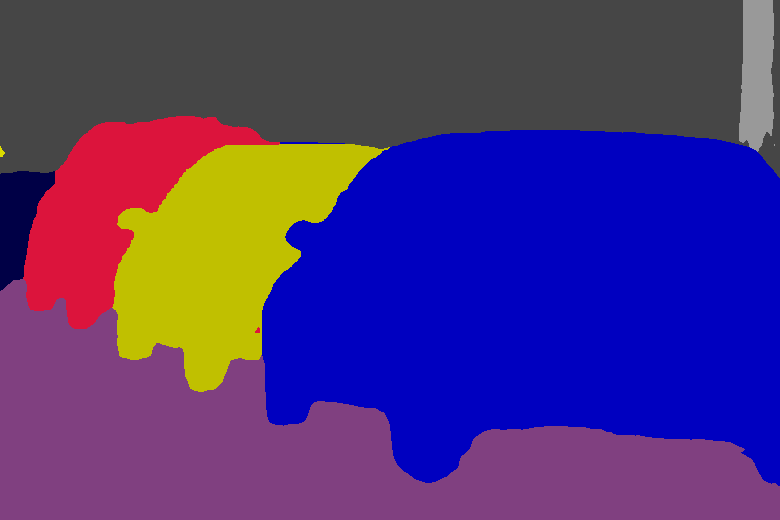}\\
		
		\global \def \im{lindau_000042_000019}
		\includegraphics[width=\VisScale\linewidth]{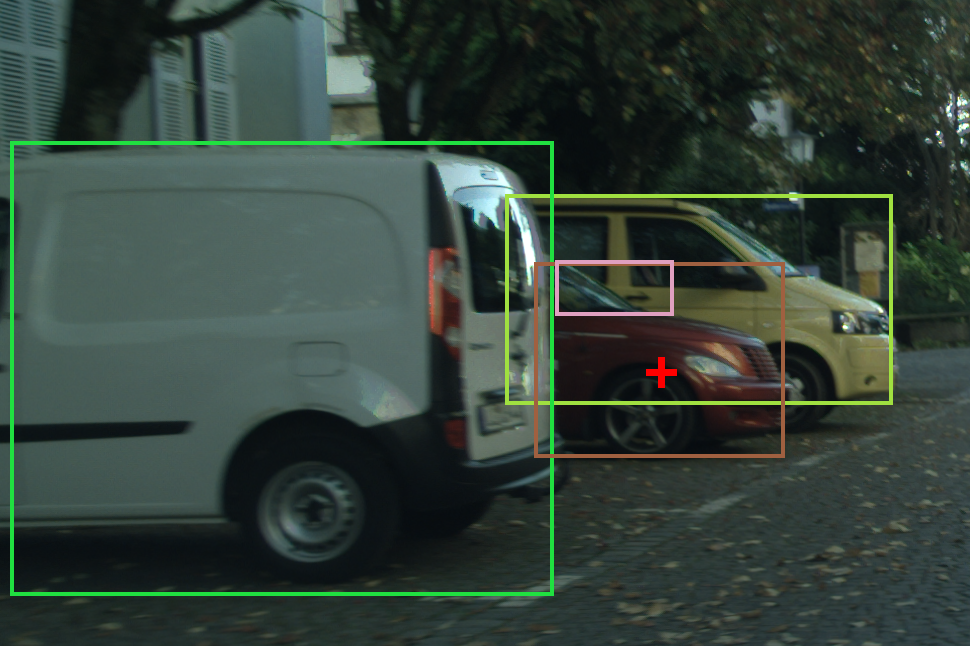}&
		\includegraphics[width=\VisScale\linewidth]{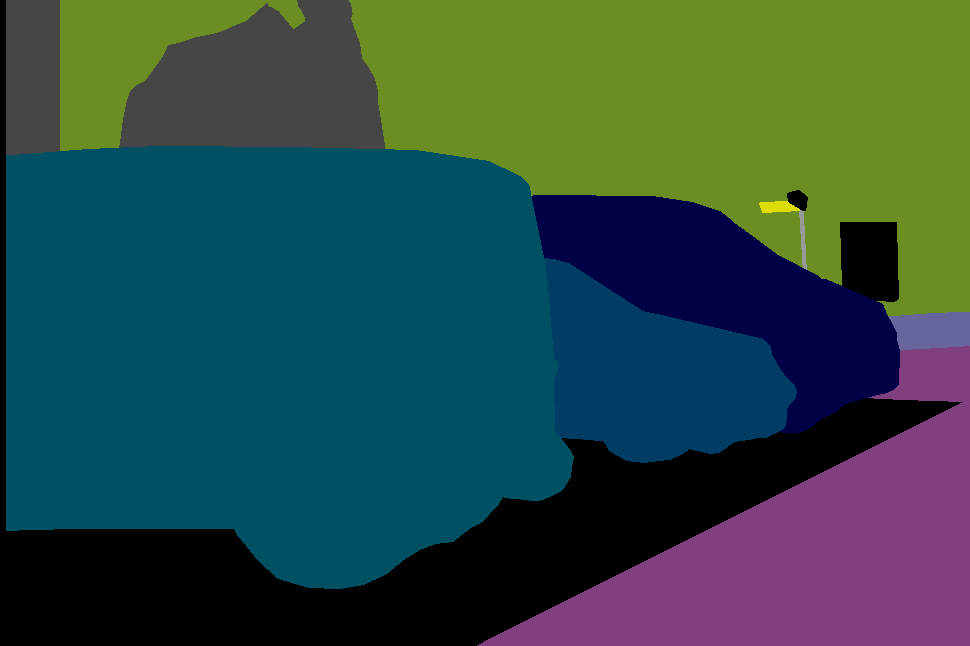}&
		\includegraphics[width=\VisScale\linewidth]{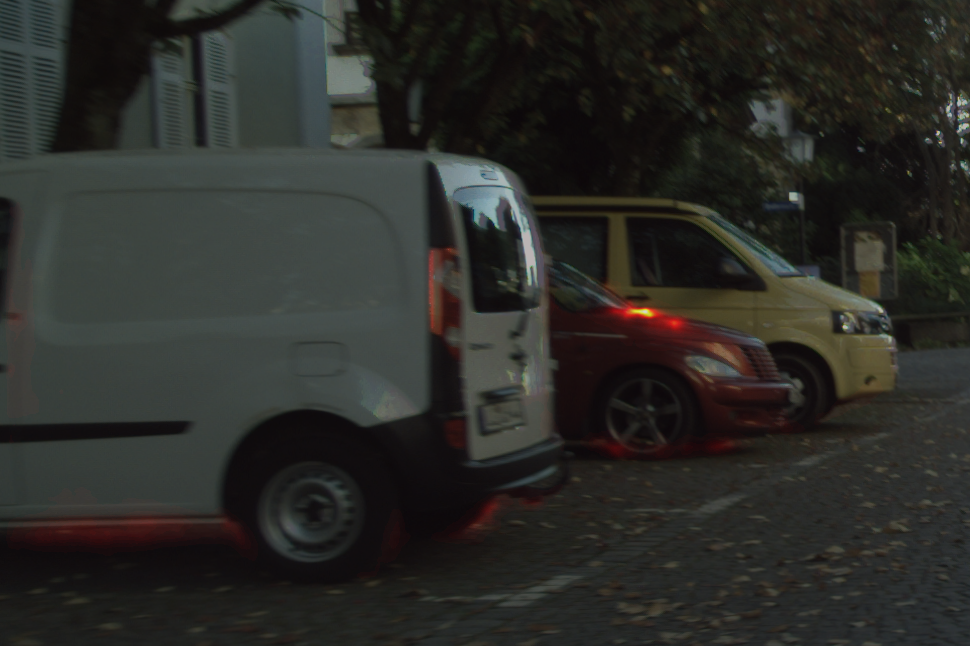}&
		\includegraphics[width=\VisScale\linewidth]{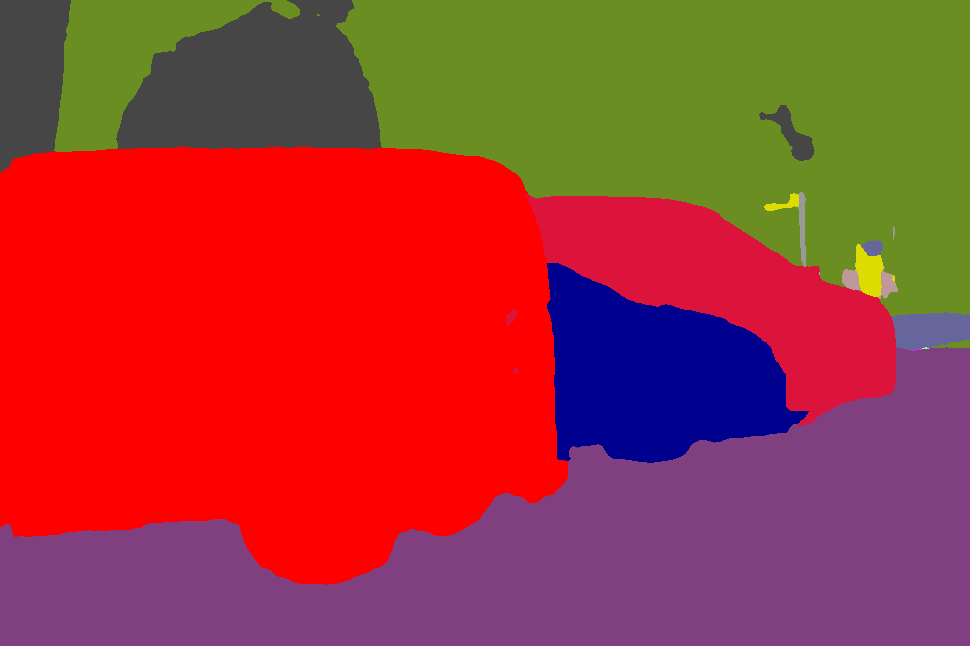}\\
		
		\global \def \im{munster_000103_000019}
		\includegraphics[width=\VisScale\linewidth]{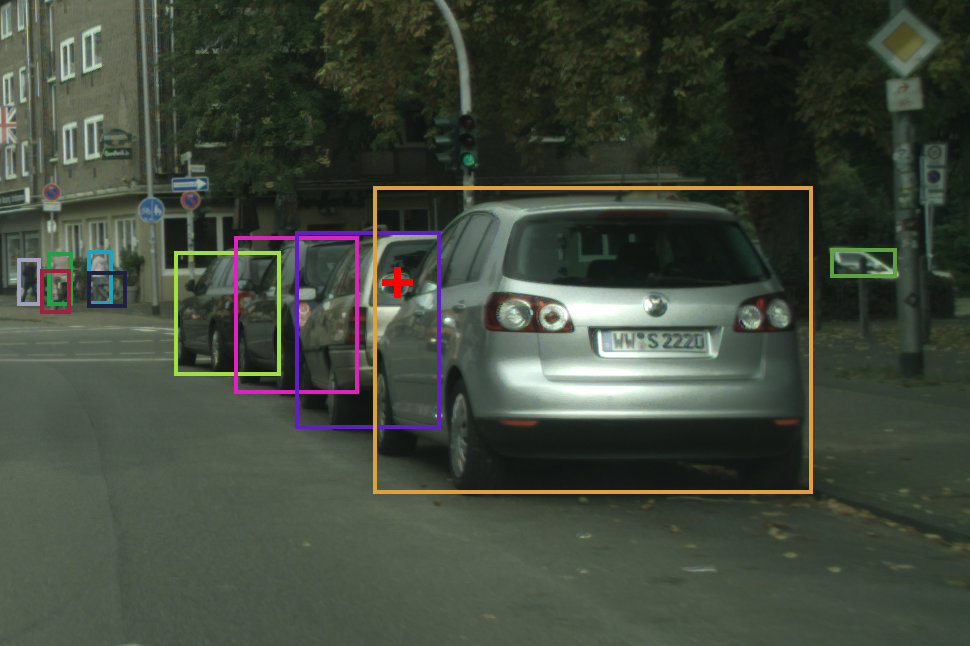}&
		\includegraphics[width=\VisScale\linewidth]{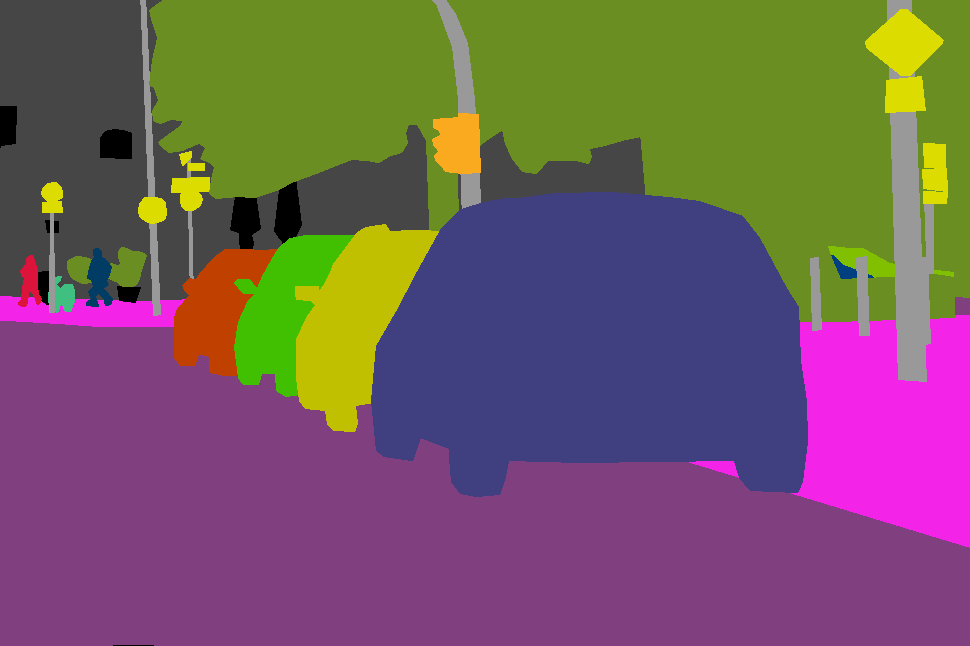}&
		\includegraphics[width=\VisScale\linewidth]{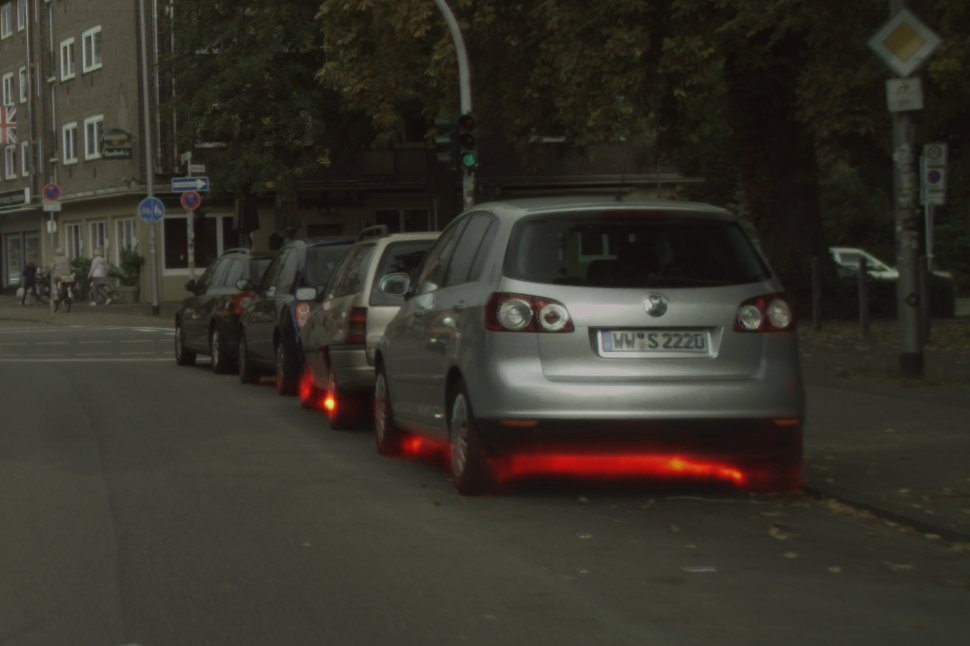}&
		\includegraphics[width=\VisScale\linewidth]{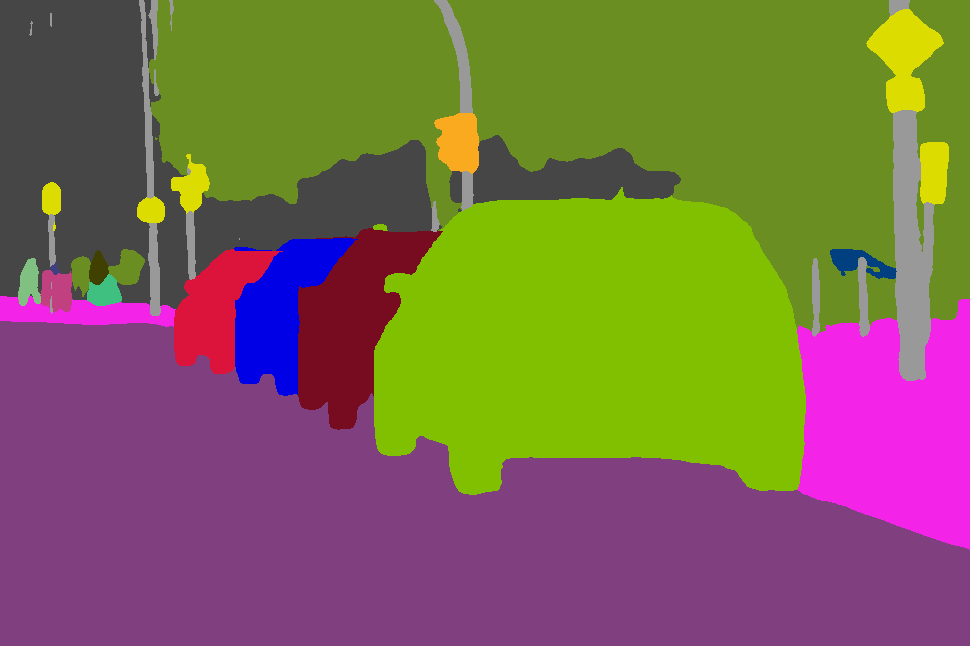}\\
	
		{\footnotesize{(a) Image with boxes}}&{\footnotesize{(b) Ground truth}}&{\footnotesize{(c) Instance affinities}}&{\footnotesize{(d) Panoptic prediction}}\\
	\end{tabularx}
	\vspace{2pt}
	\caption{Additional examples of instance affinities. In (c), we show the instance affinities -- overlaid on input images to aid visualisation -- of the cross-marked pixels in (a). These affinities and predictions are predicted by our mask-free models which use only bounding boxes. They can be seen to help segment full objects when bounding box localisation is poor (Row 1), and attribute pixels within multiple bounding boxes to the correct instances (Row 2 to 5). For Row 4, our proposed method is able to overcome a false positive detection, as the dynamic potential is robust towards false detections. For Row 5, the cross-marked pixel is on the wing mirror of the closest silver car, and our fine-grained instance affinity is able to attribute the mirror to the correct car, while the ground truth has failed to correctly label as such.
	}
	\label{fig:vis_affinity_supp}
\end{figure*}

%% file: figures/supp_cityscapes_results.tex
\begin{figure*}
	\setlength{\tabcolsep}{2pt}
	\centering
	\begin{tabularx}{\textwidth}{YYYY}
		\global \def \VisScale{1}

		\global \def \im{frankfurt_000001_047178}
		\includegraphics[width=\VisScale\linewidth]{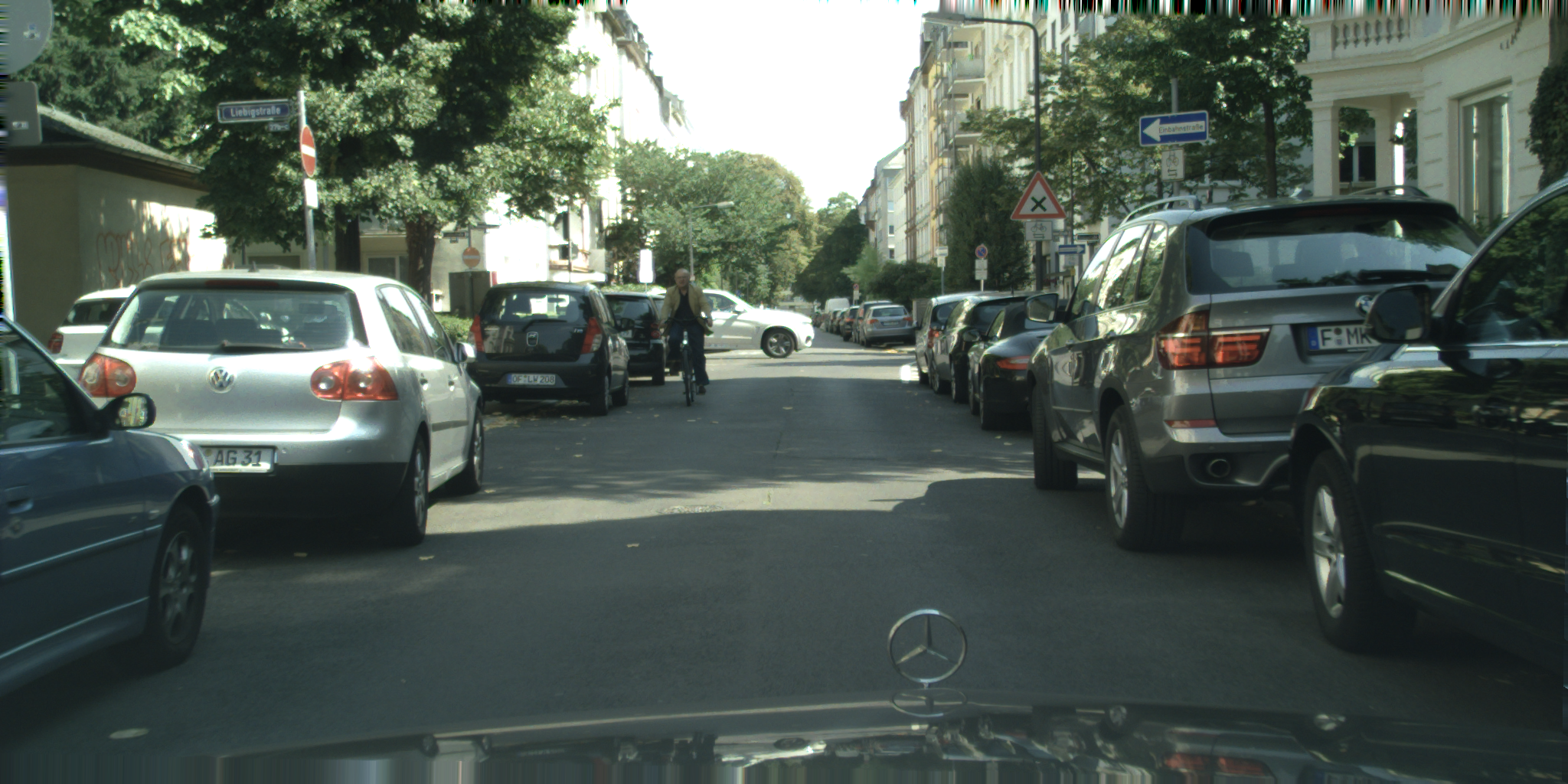}&
		\includegraphics[width=\VisScale\linewidth]{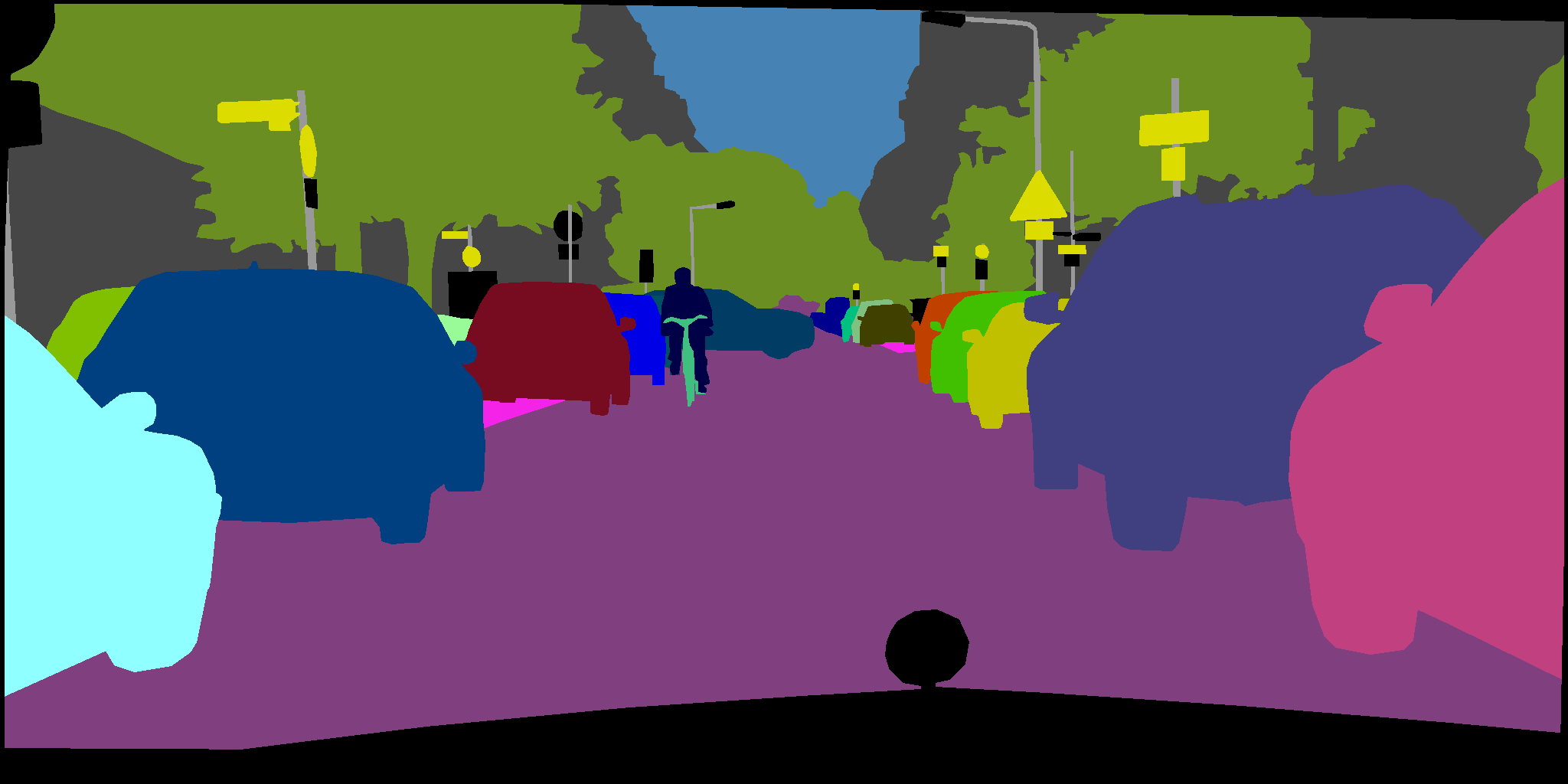}&
		\includegraphics[width=\VisScale\linewidth]{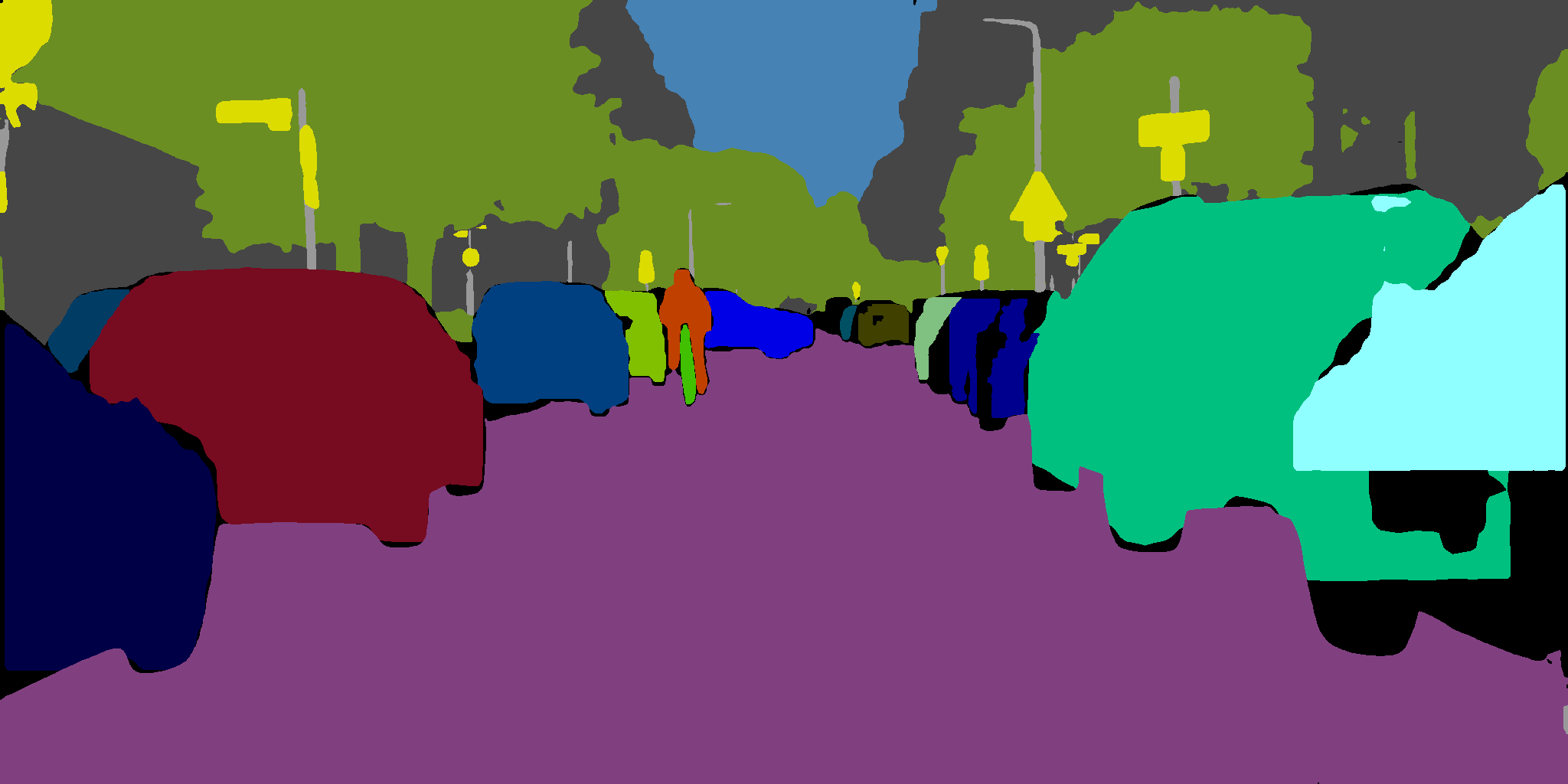}&
		\includegraphics[width=\VisScale\linewidth]{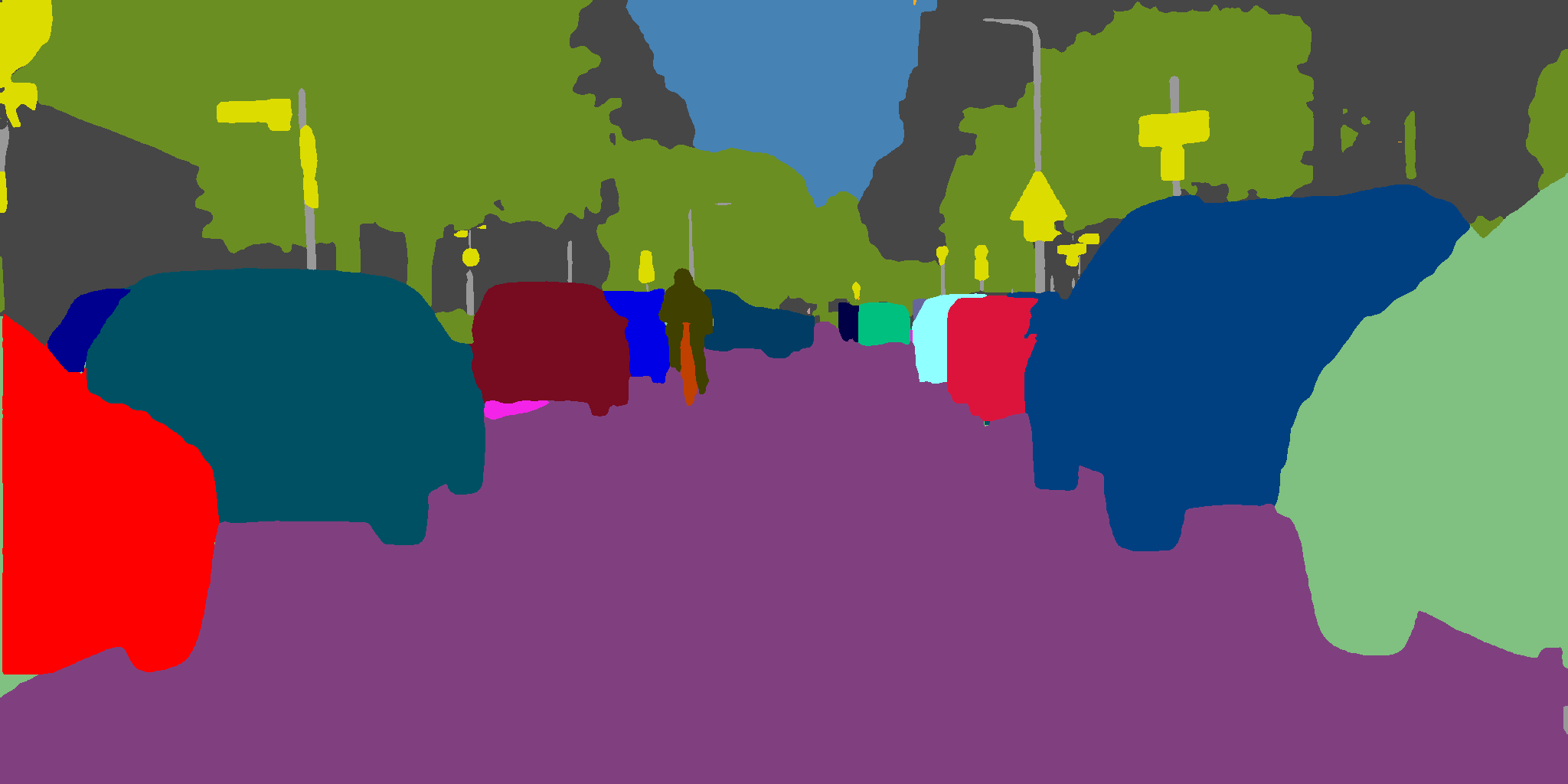}\\
		
		\global \def \im{frankfurt_000001_047552}
		\includegraphics[width=\VisScale\linewidth]{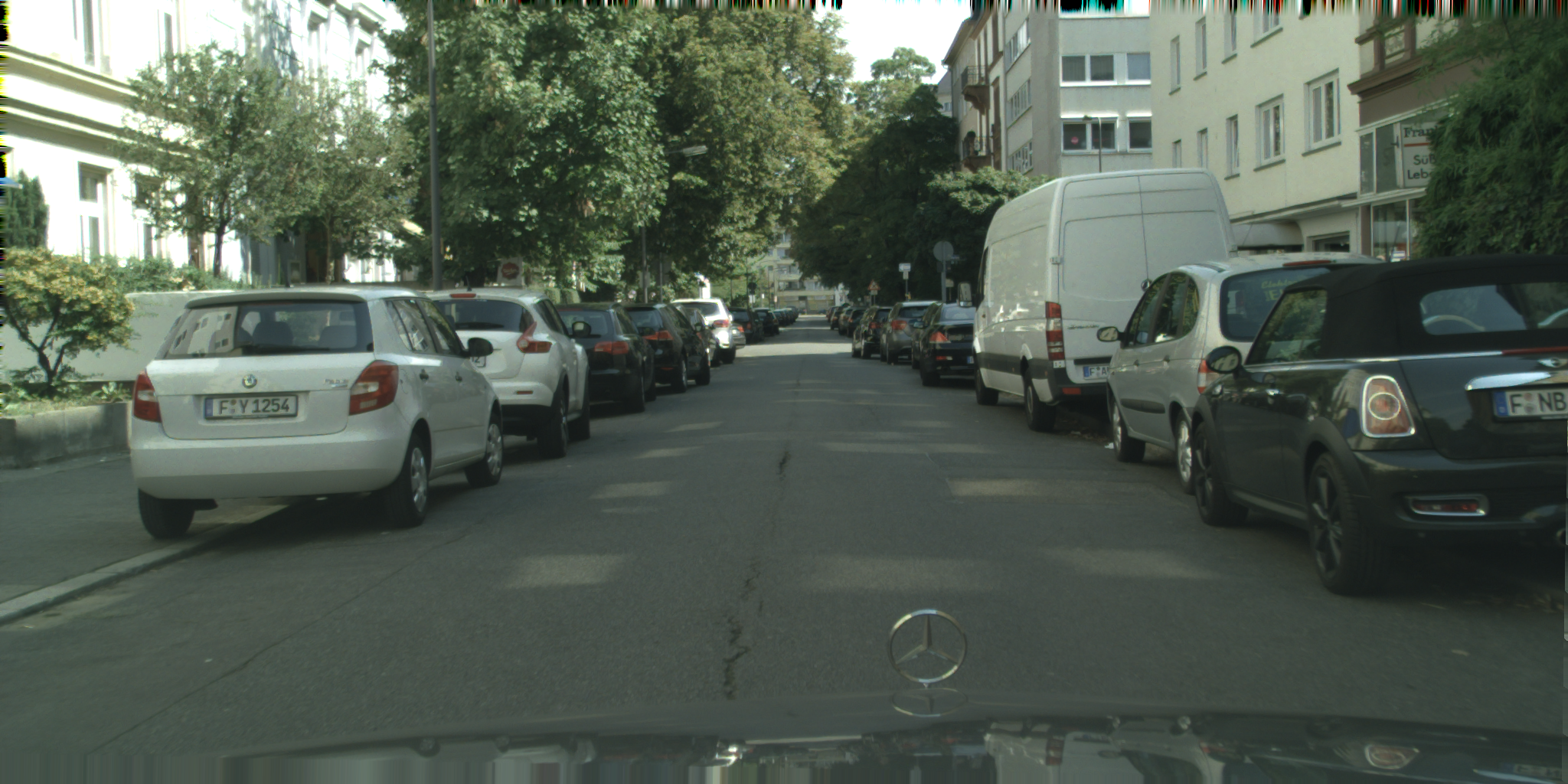}&
		\includegraphics[width=\VisScale\linewidth]{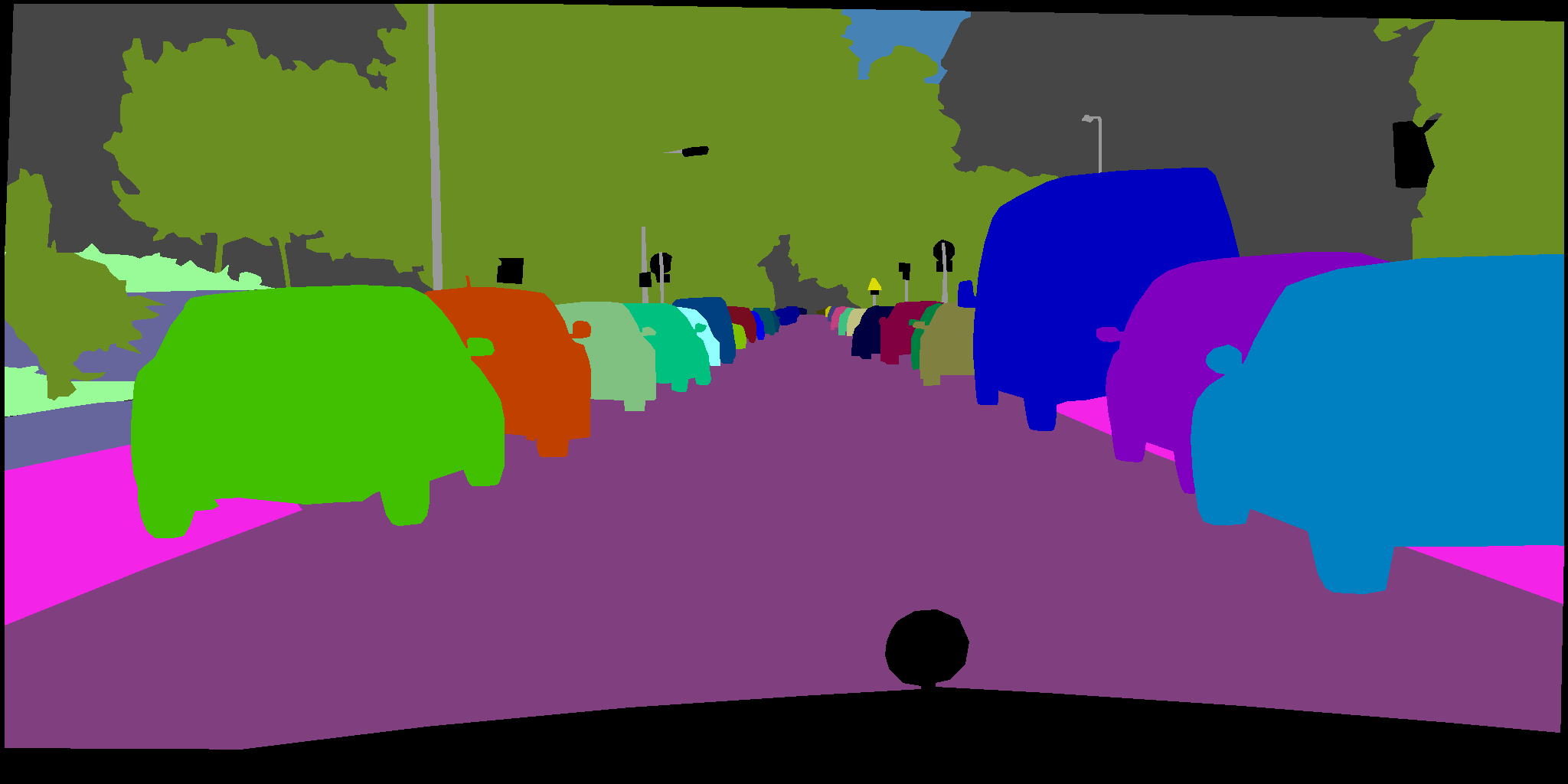}&
		\includegraphics[width=\VisScale\linewidth]{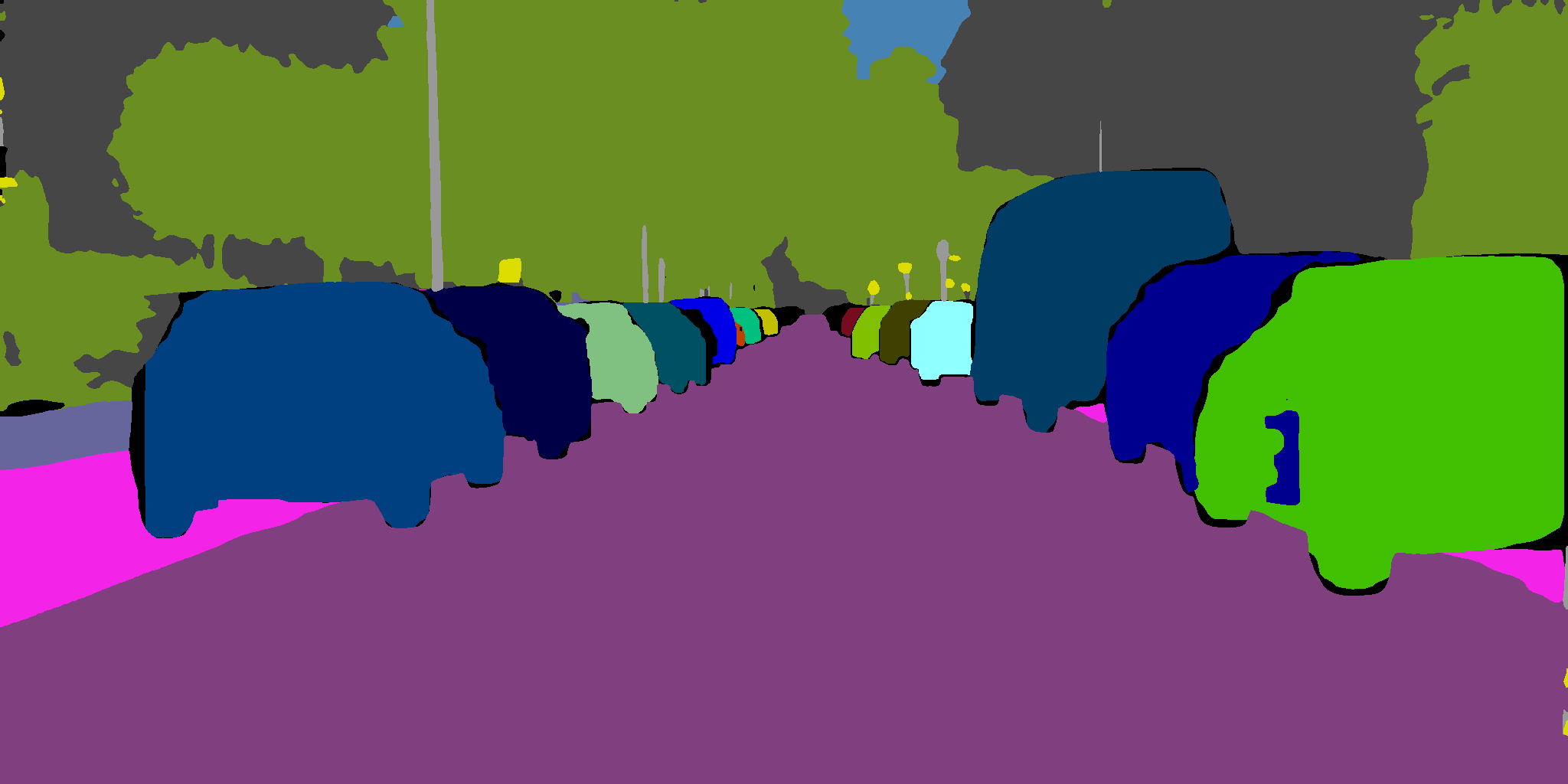}&
		\includegraphics[width=\VisScale\linewidth]{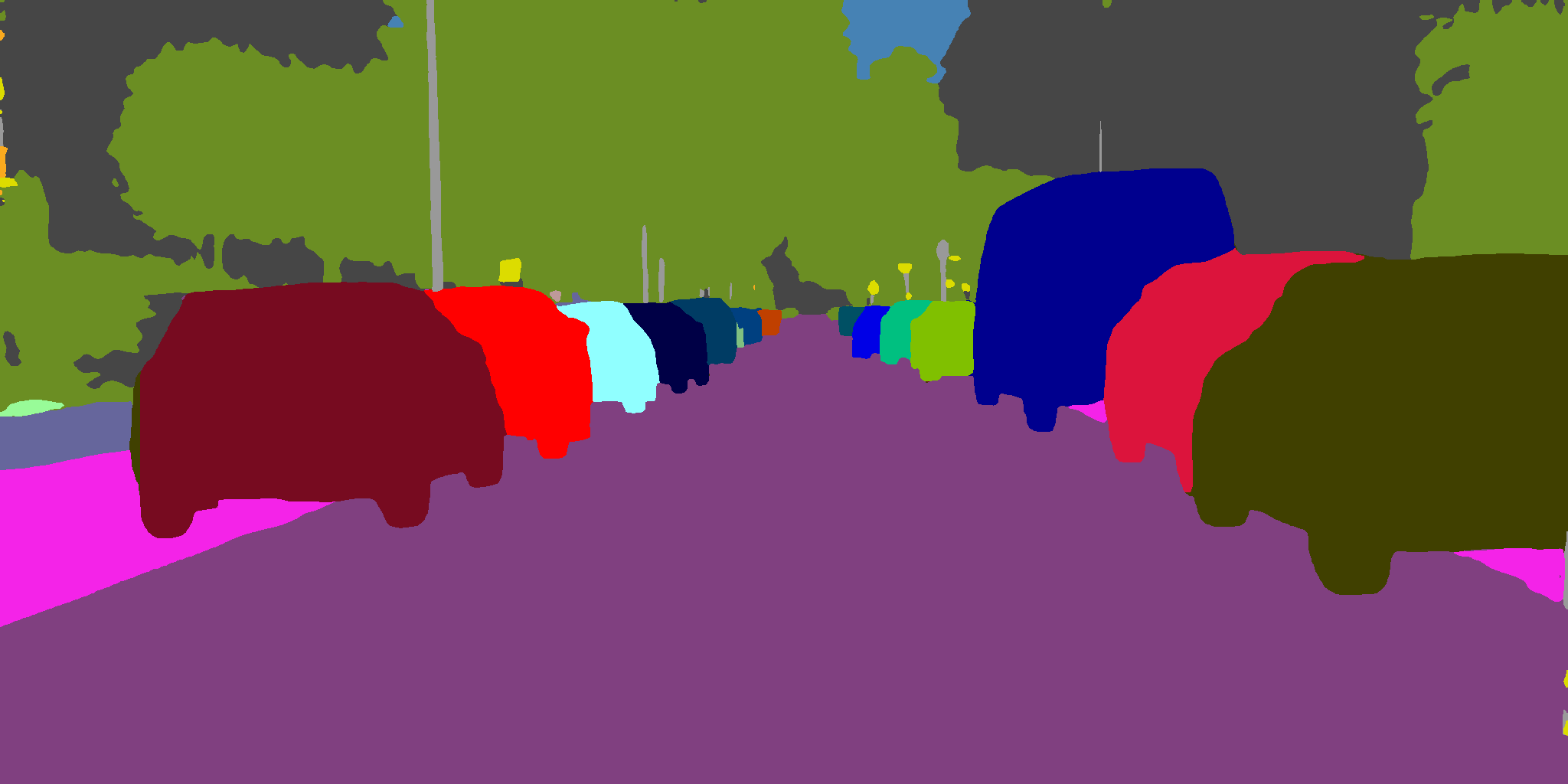}\\
		
		\global \def \im{frankfurt_000001_073243}
		\includegraphics[width=\VisScale\linewidth]{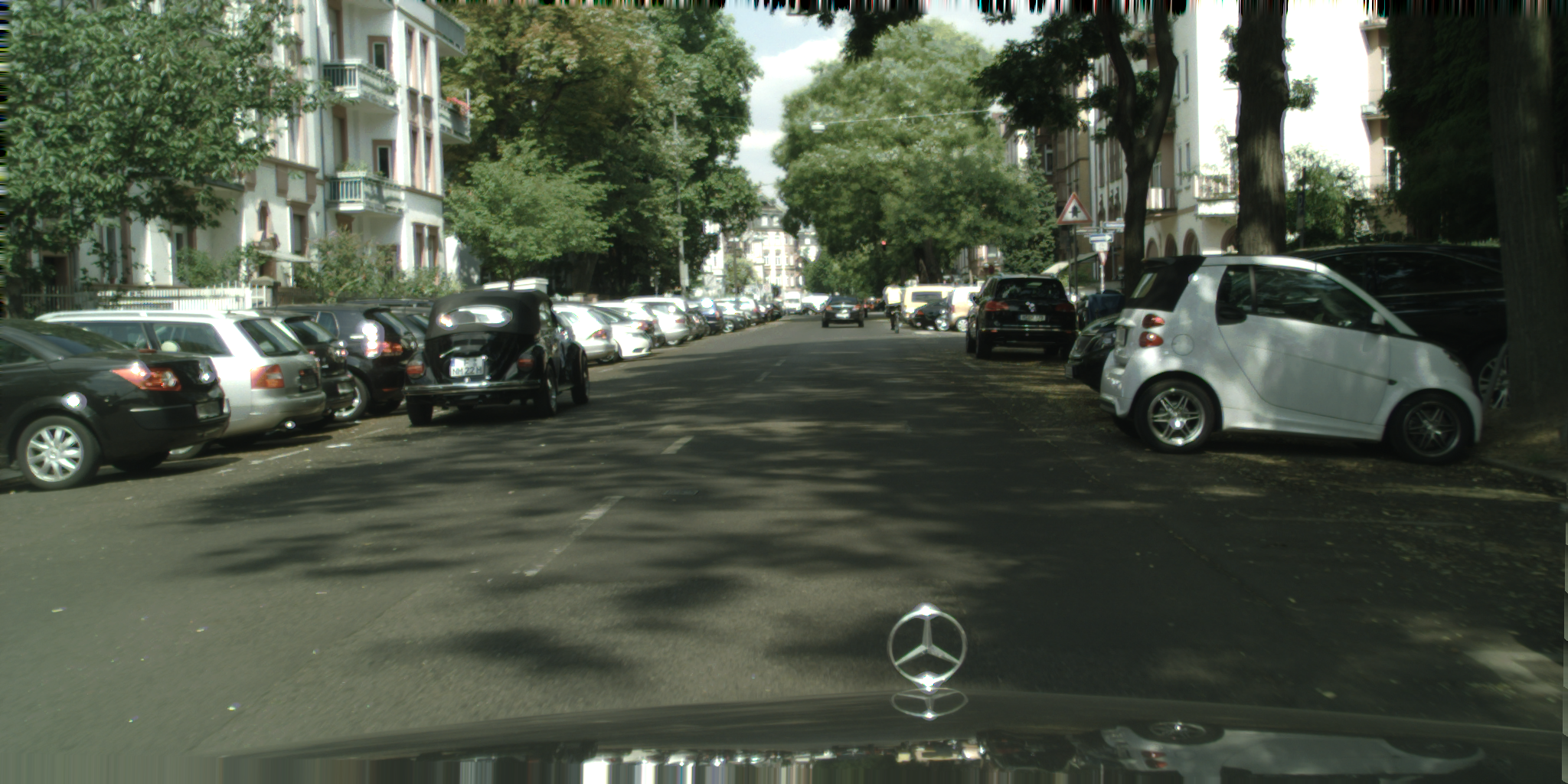}&
		\includegraphics[width=\VisScale\linewidth]{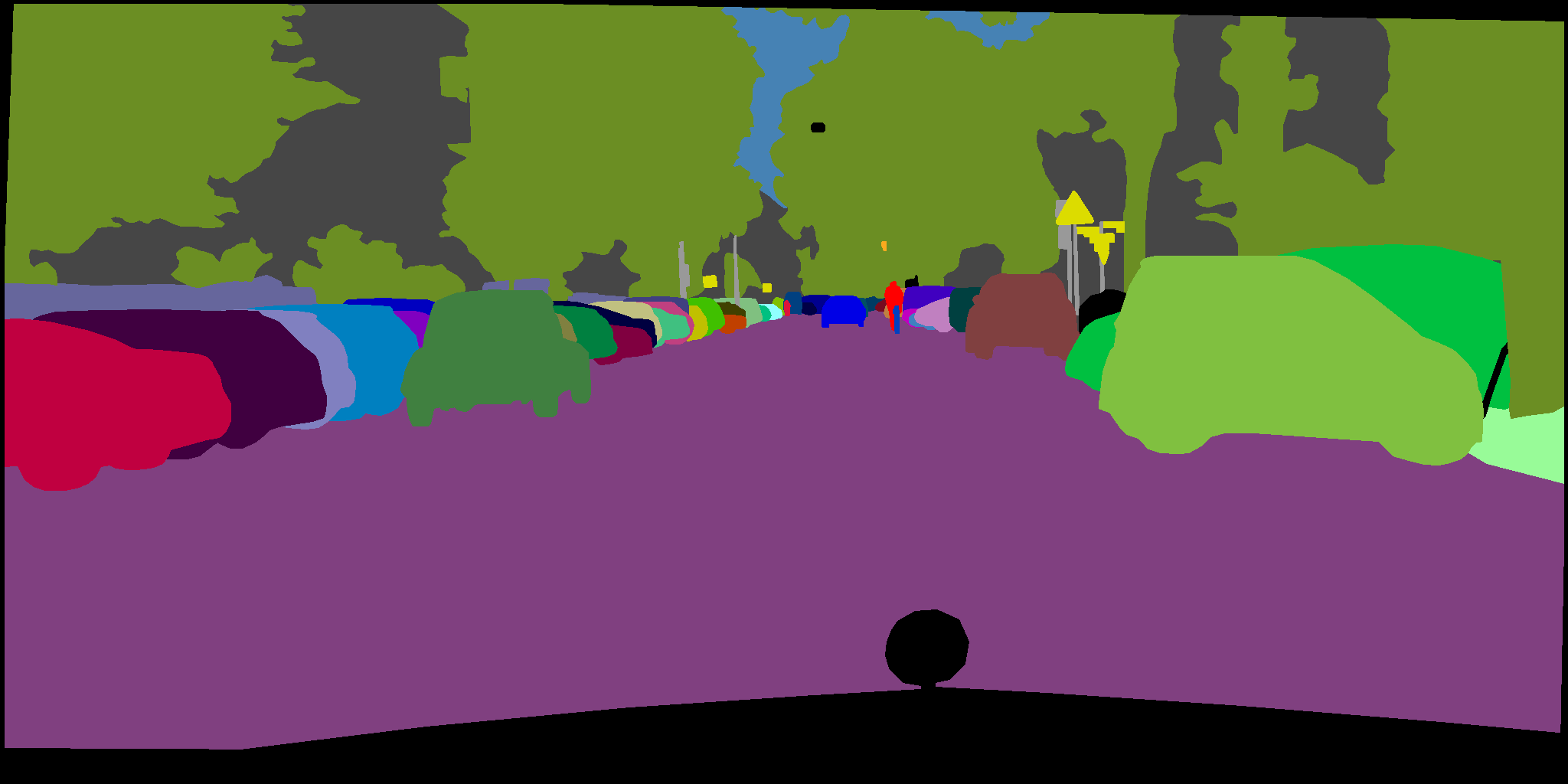}&
		\includegraphics[width=\VisScale\linewidth]{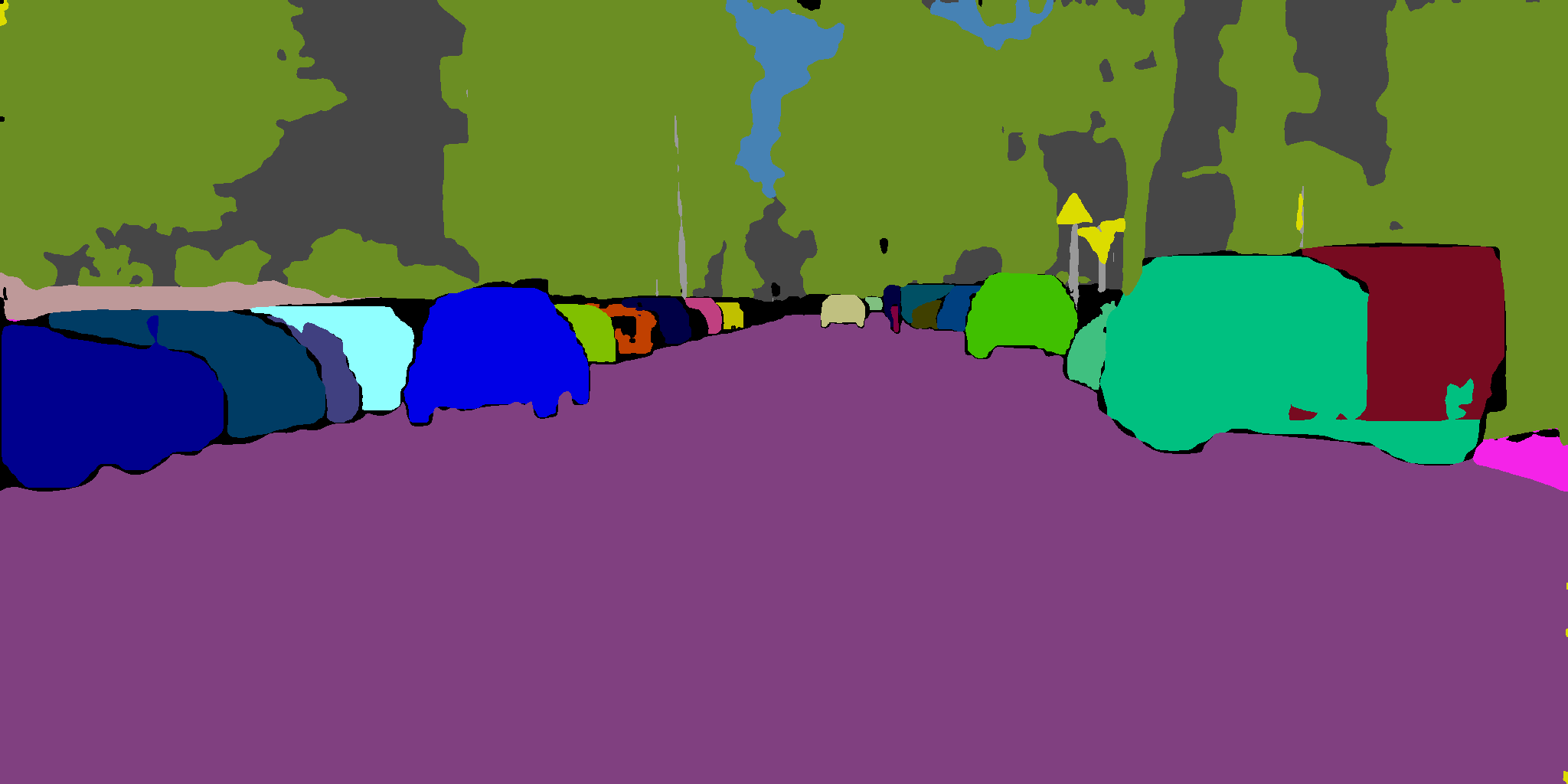}&
		\includegraphics[width=\VisScale\linewidth]{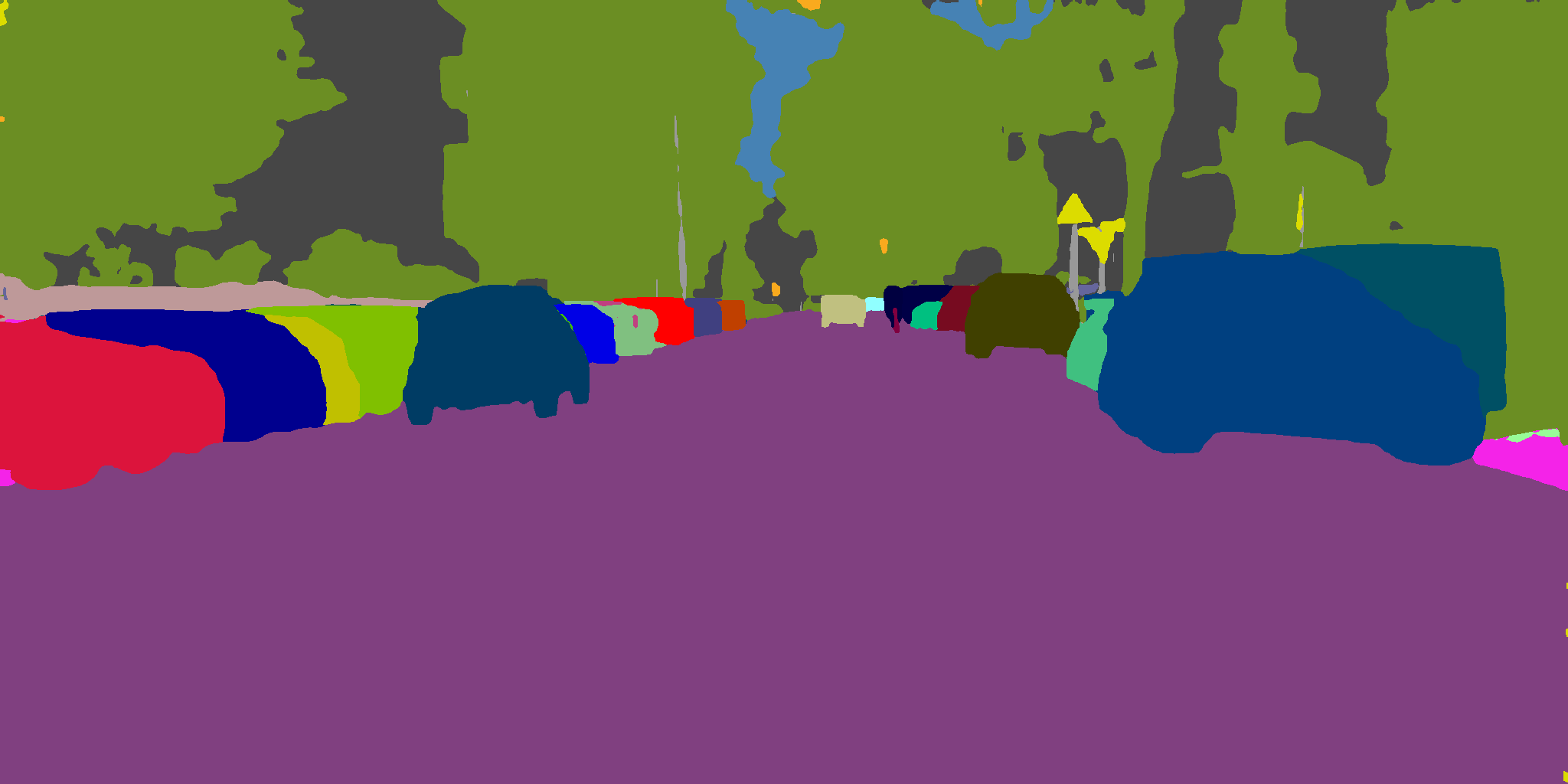}\\
		
		\global \def \im{munster_000019_000019}
		\includegraphics[width=\VisScale\linewidth]{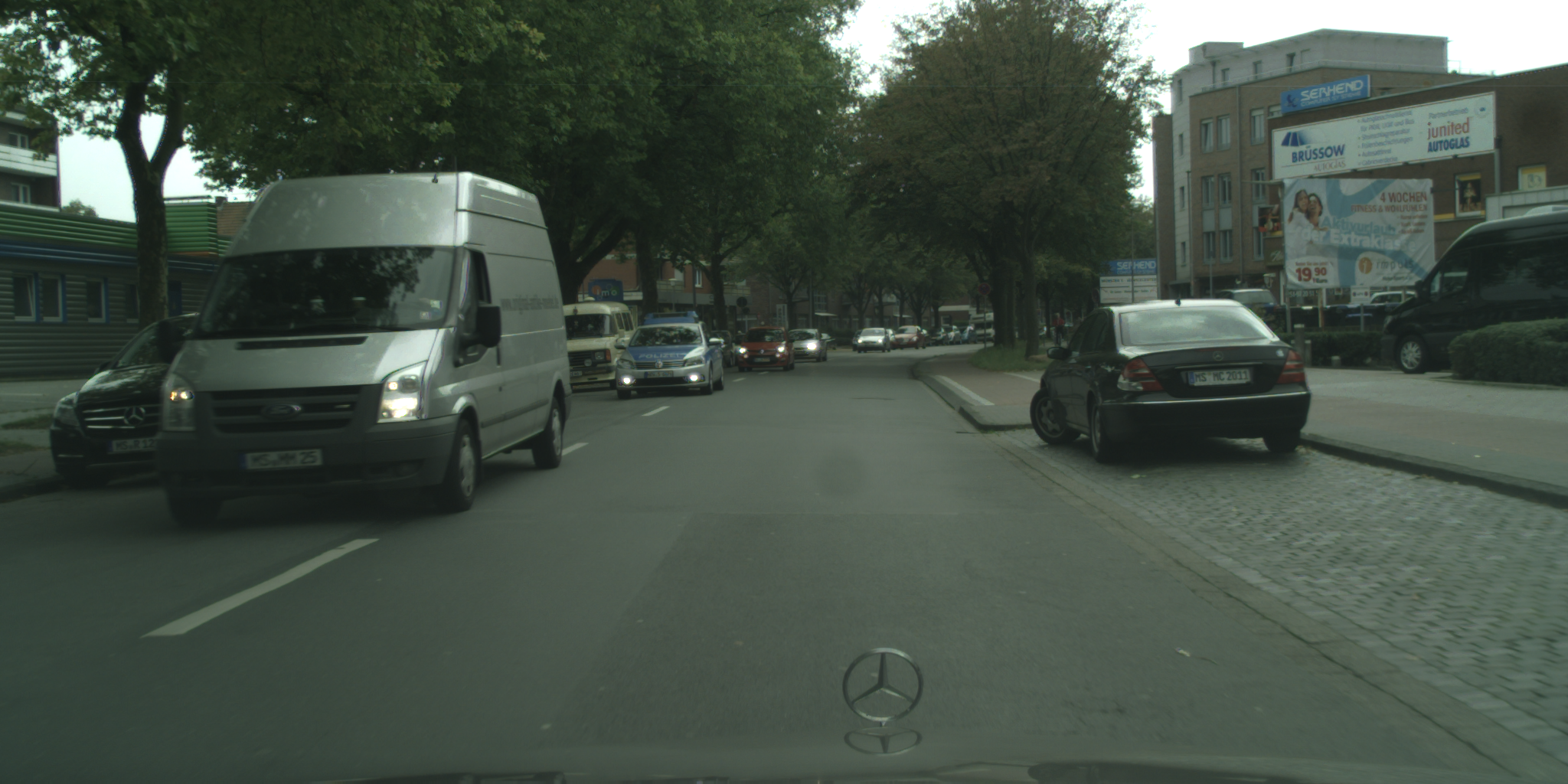}&
		\includegraphics[width=\VisScale\linewidth]{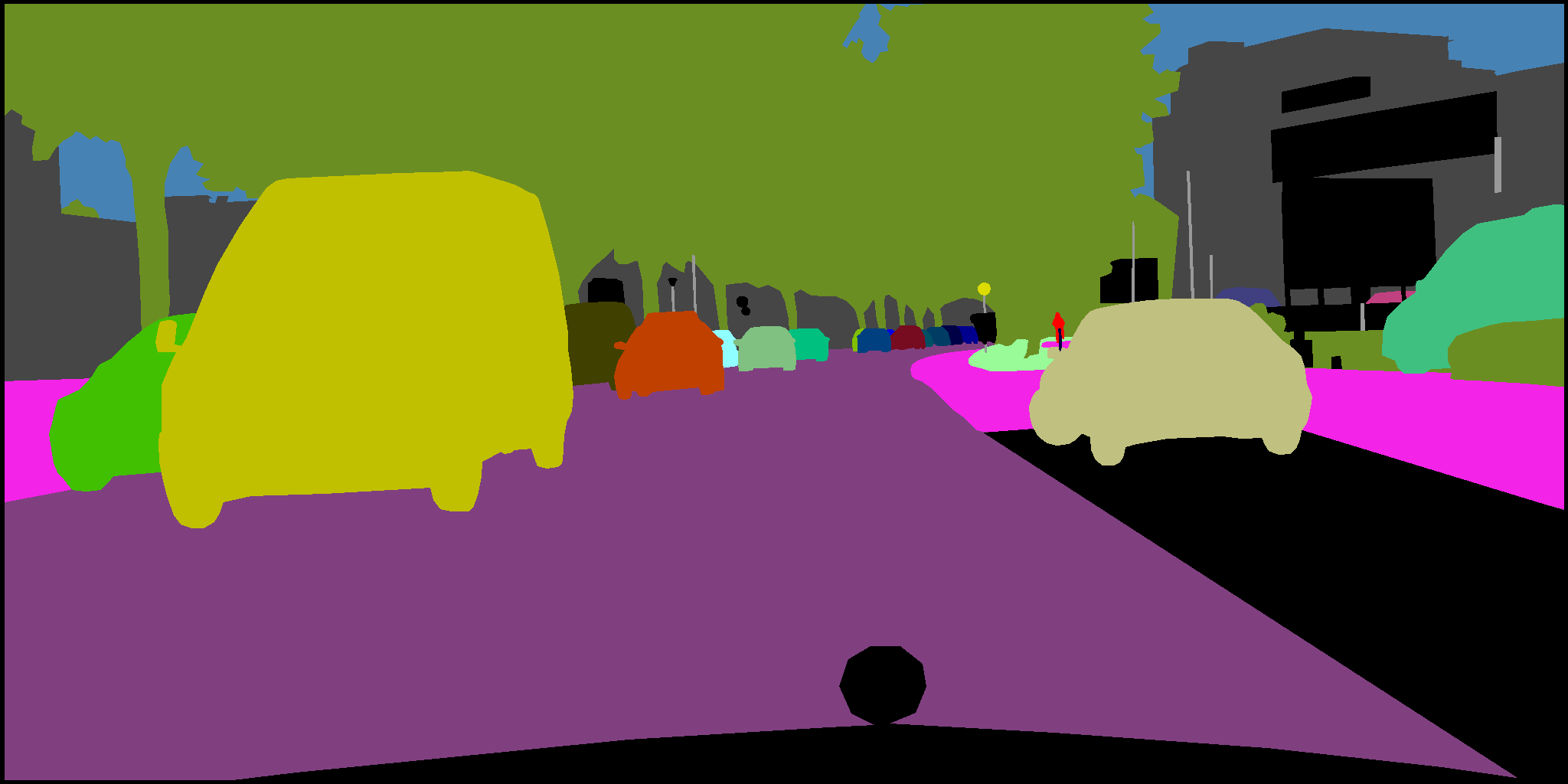}&
		\includegraphics[width=\VisScale\linewidth]{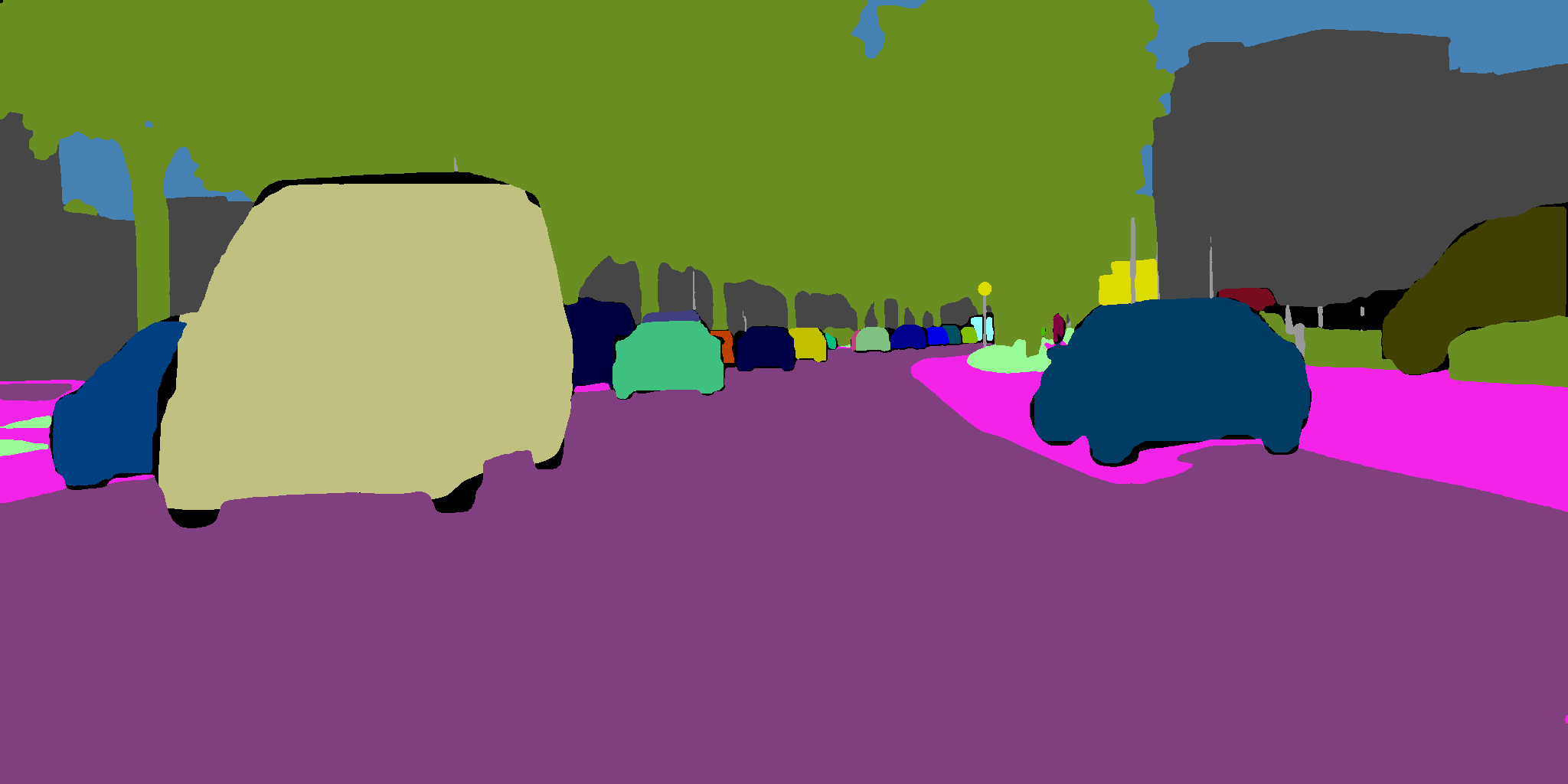}&
		\includegraphics[width=\VisScale\linewidth]{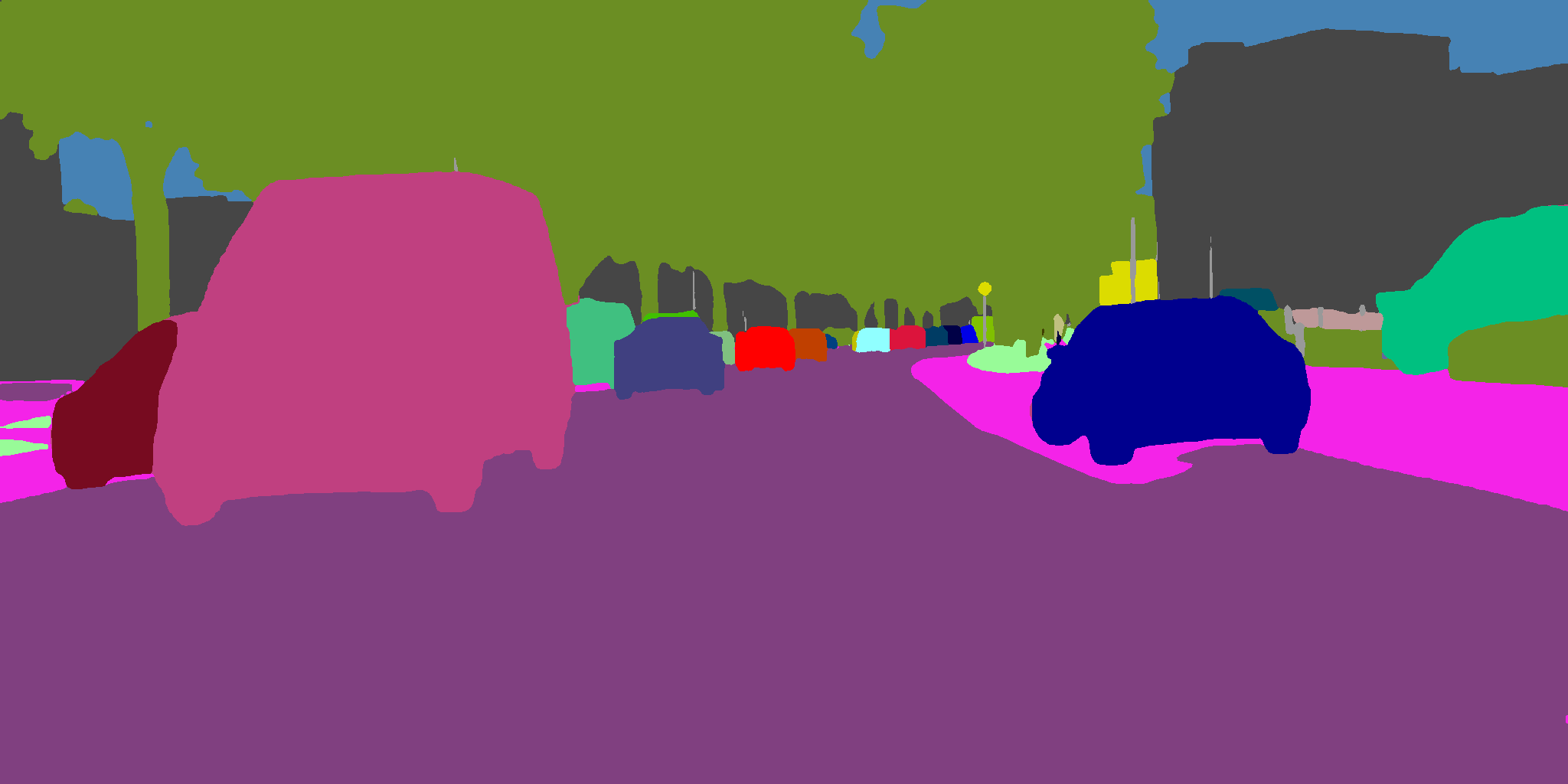}\\
		
		\global \def \im{munster_000128_000019}
		\includegraphics[width=\VisScale\linewidth]{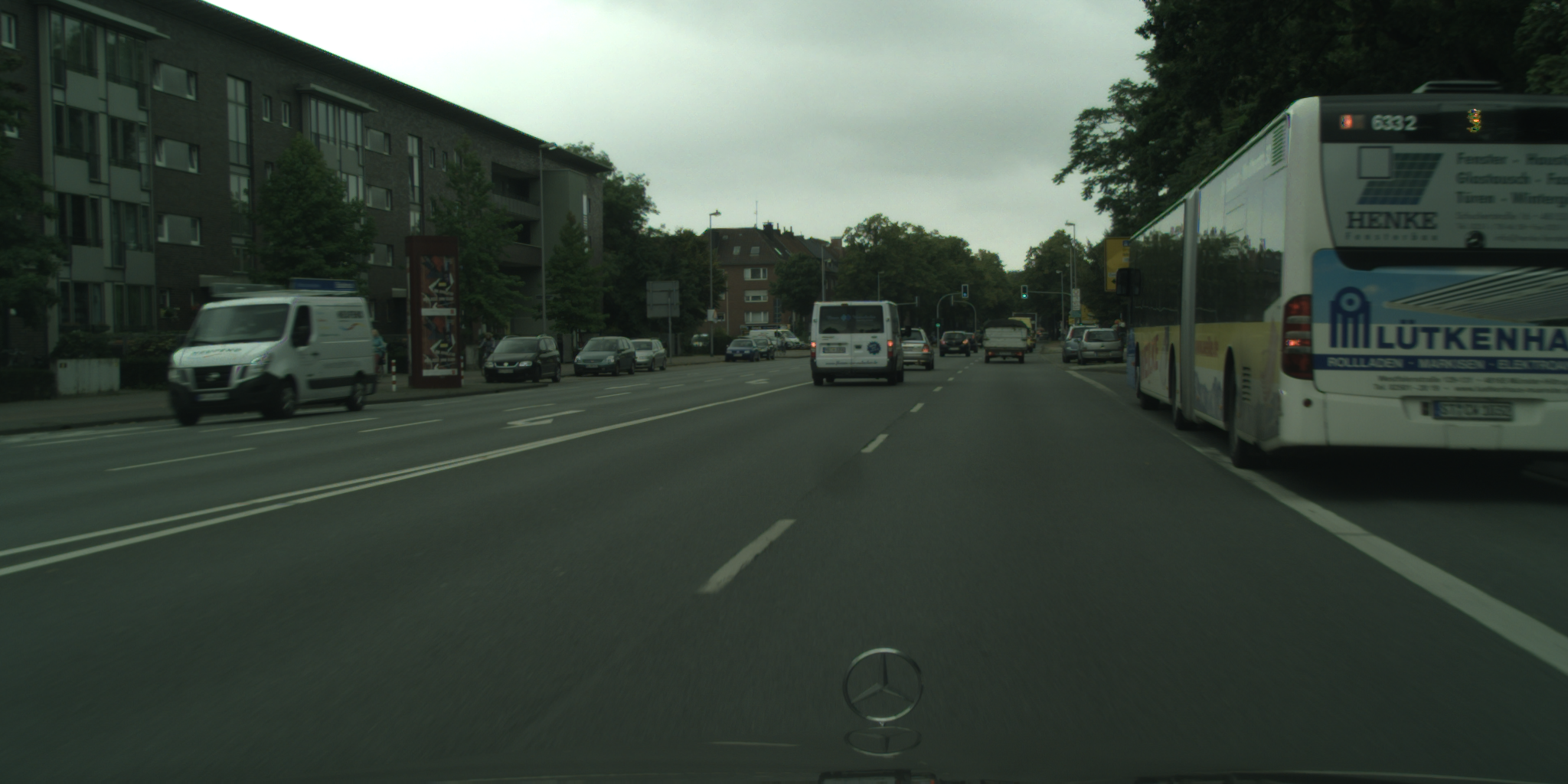}&
		\includegraphics[width=\VisScale\linewidth]{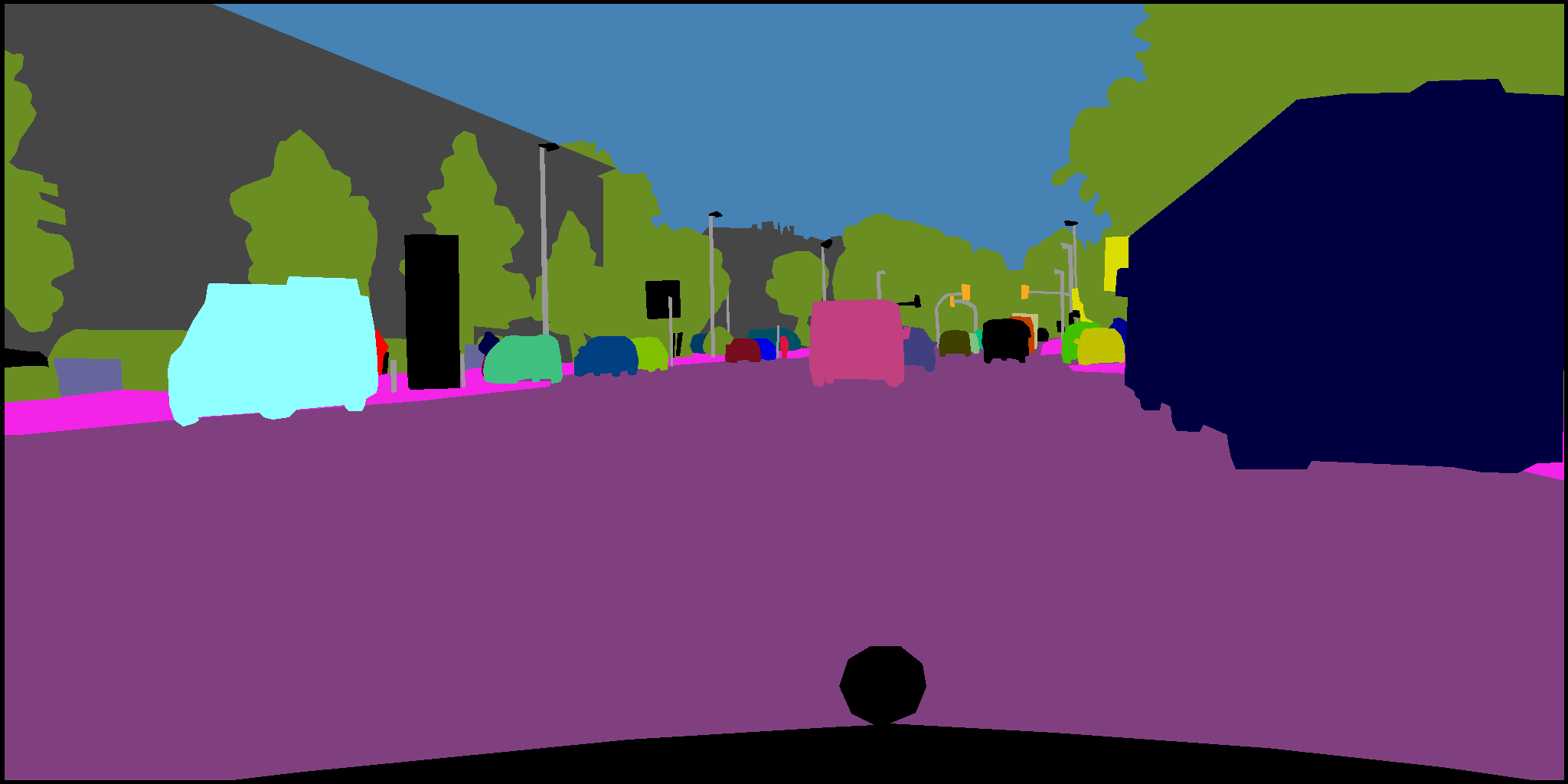}&
		\includegraphics[width=\VisScale\linewidth]{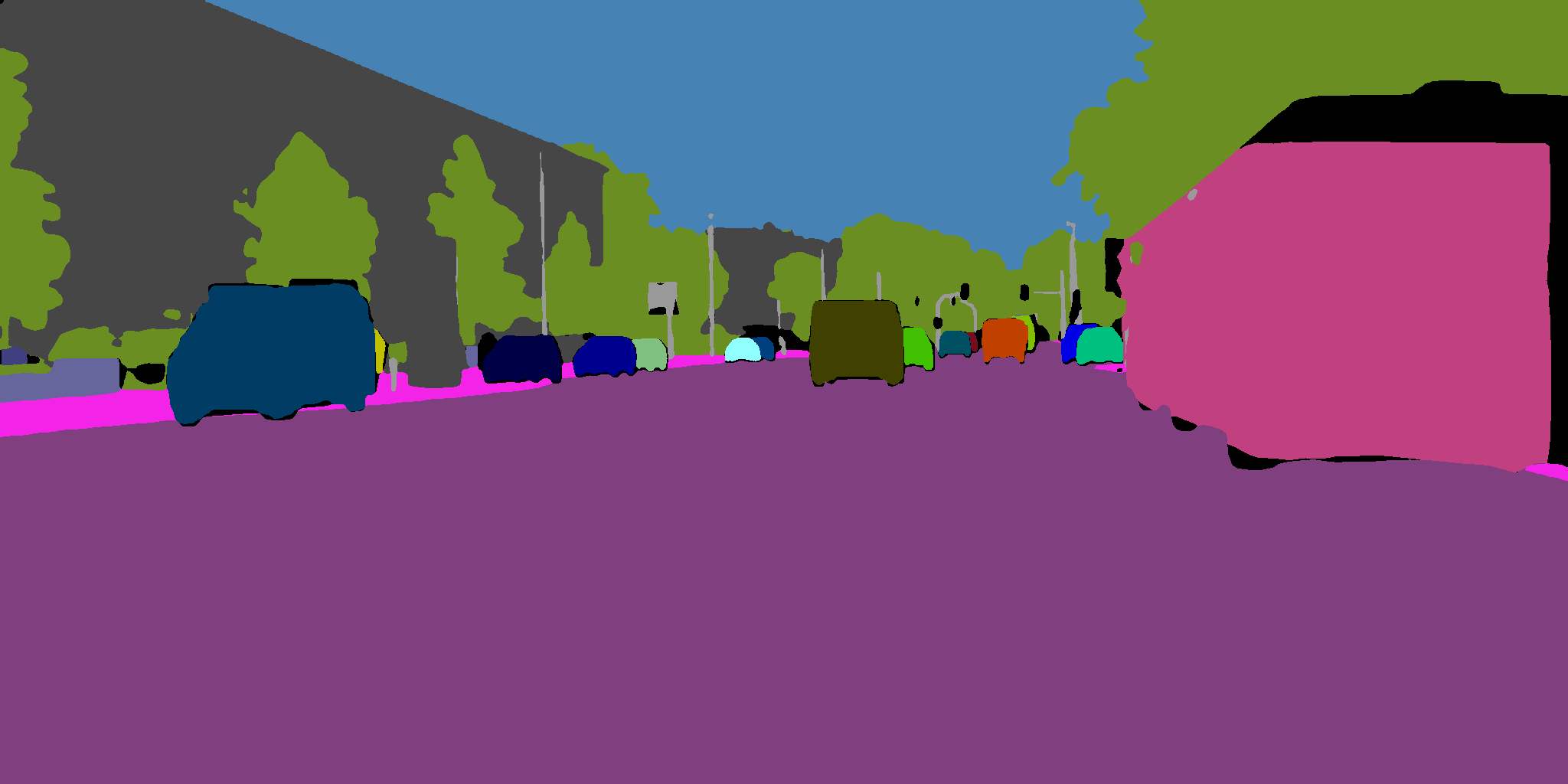}&
		\includegraphics[width=\VisScale\linewidth]{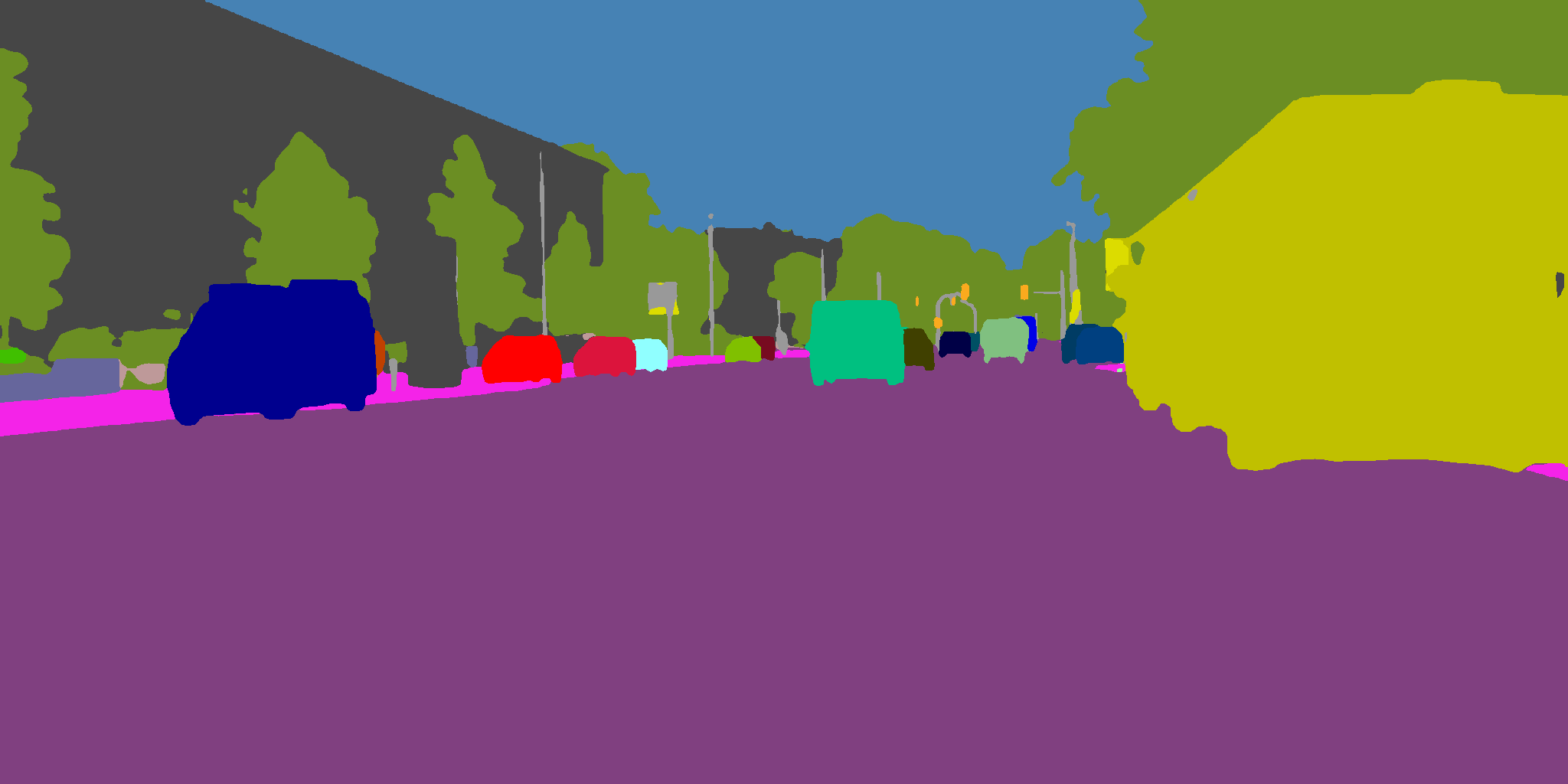}\\
		
		{\footnotesize{(a) Image}}&{\footnotesize{(b) Ground truth}}&{\footnotesize{(c) Heuristic Fusion~\cite{kirillov2018panoptic}}}&{\footnotesize{(d) Ours}}\\
	\end{tabularx}
	\vspace{2pt}
	\caption{Qualitative results on Cityscapes. Column (c) and (d) are produced by the same model under different inference strategies -- either by heuristic merger~\cite{kirillov2018panoptic} or with our proposed panoptic segmentation submodule. Row 1 to 3 shows that our model are able to revise erroneous cues and resolve conflicts between overlapping object masks. Row 4 and 5 demonstrate our network's ability to segment outside boxes, when boxes do not cover the full extent of an object.
	}
	\label{fig:supp_cityscapes}
\end{figure*}

%% file: figures/supp_coco_results.tex
\begin{figure*}
	\setlength{\tabcolsep}{2pt}
	\centering
	\begin{tabularx}{\textwidth}{YYYY}
		\global \def \VisScale{1}

		\global \def \im{000000076731}
		\includegraphics[width=\VisScale\linewidth]{figures/coco_results/supp/image/COCO_val2014_\im}&
		\includegraphics[width=\VisScale\linewidth]{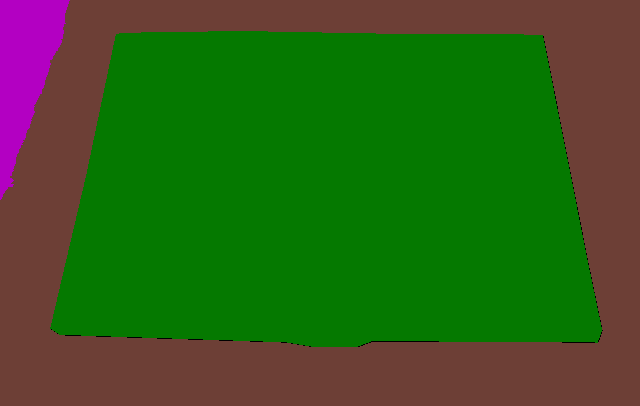}&
		\includegraphics[width=\VisScale\linewidth]{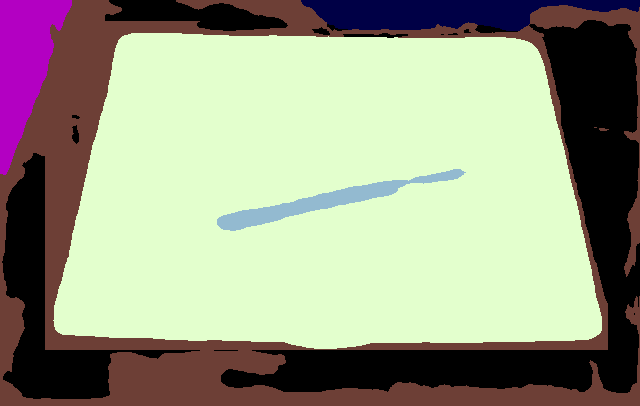}&
		\includegraphics[width=\VisScale\linewidth]{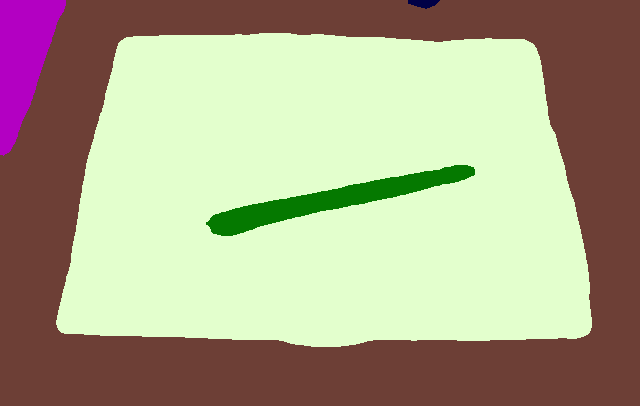}\\
		
		\global \def \im{000000046031}
		\includegraphics[width=\VisScale\linewidth]{figures/coco_results/supp/image/COCO_val2014_\im}&
		\includegraphics[width=\VisScale\linewidth]{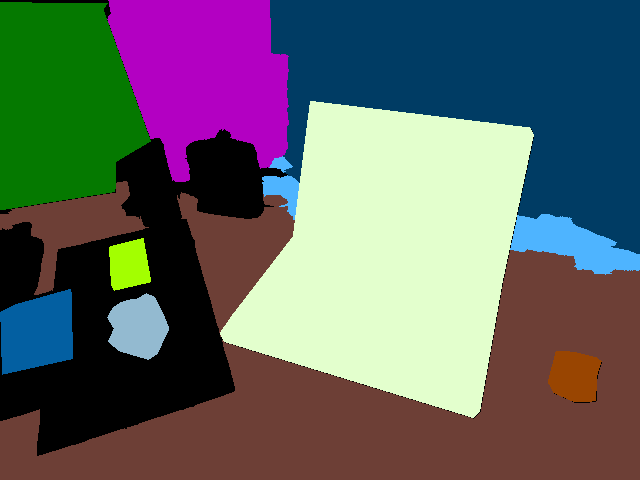}&
		\includegraphics[width=\VisScale\linewidth]{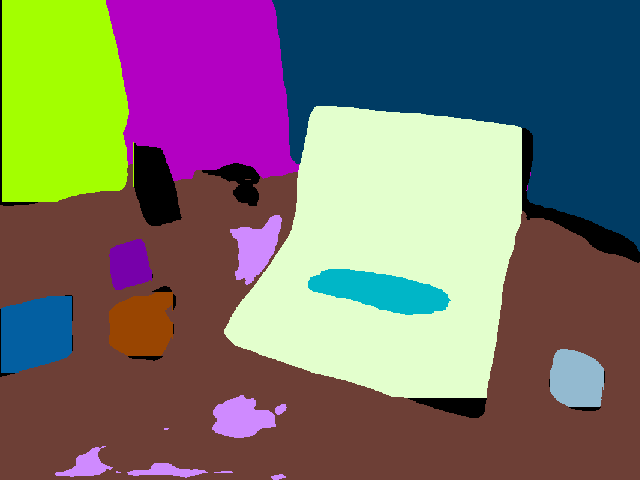}&
		\includegraphics[width=\VisScale\linewidth]{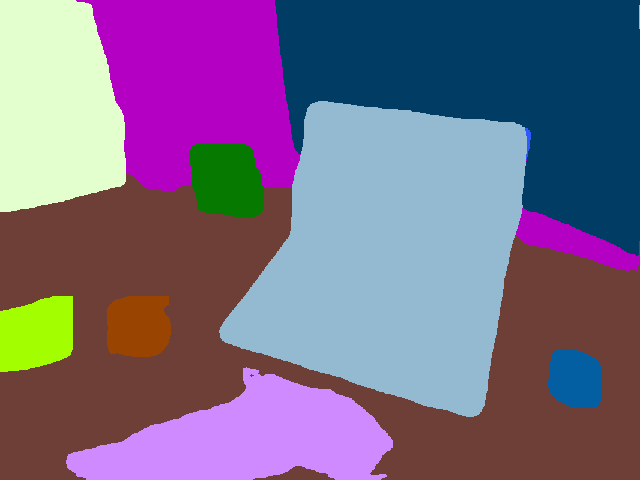}\\
		
		\global \def \im{000000050811}
		\includegraphics[width=\VisScale\linewidth]{figures/coco_results/supp/image/COCO_val2014_\im}&
		\includegraphics[width=\VisScale\linewidth]{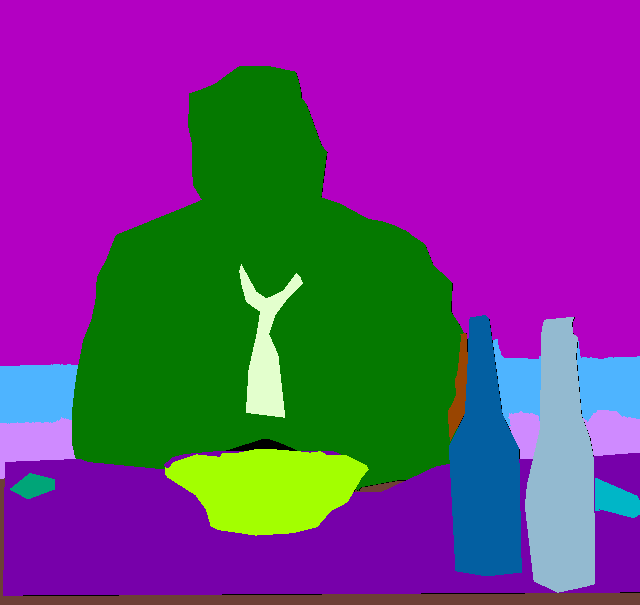}&
		\includegraphics[width=\VisScale\linewidth]{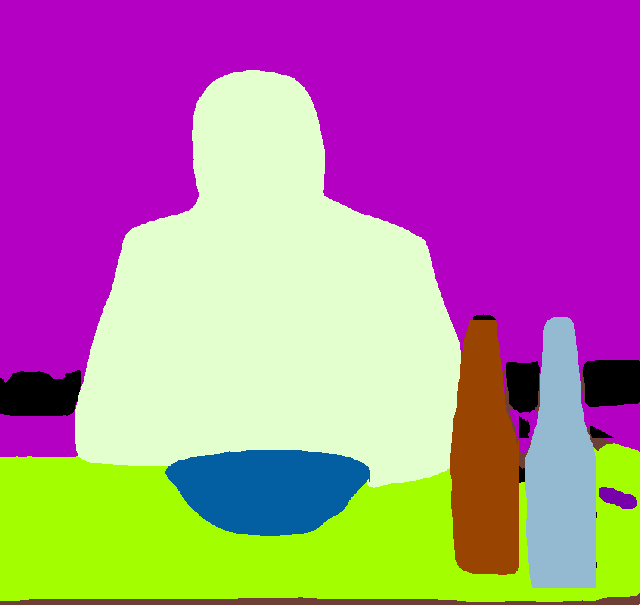}&
		\includegraphics[width=\VisScale\linewidth]{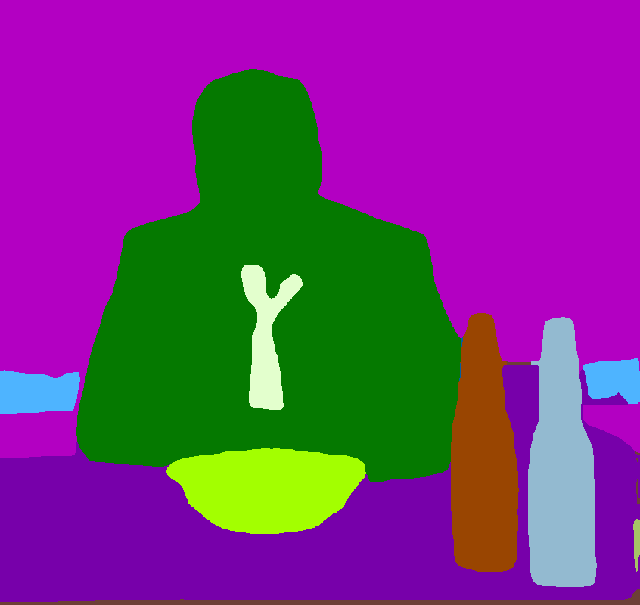}\\
		
		\global \def \im{000000019742}
		\includegraphics[width=\VisScale\linewidth]{figures/coco_results/supp/image/COCO_val2014_\im}&
		\includegraphics[width=\VisScale\linewidth]{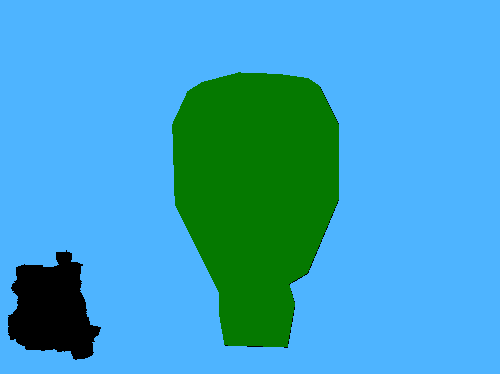}&
		\includegraphics[width=\VisScale\linewidth]{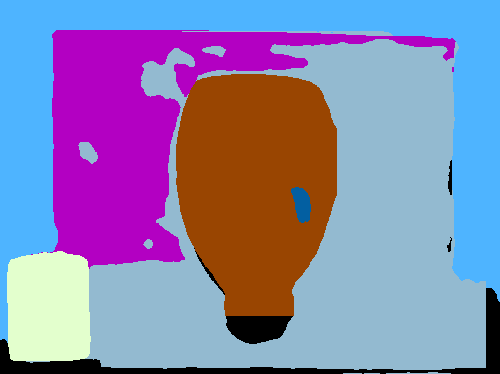}&
		\includegraphics[width=\VisScale\linewidth]{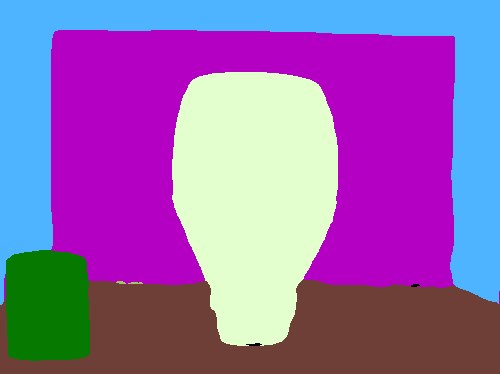}\\
		
		\global \def \im{000000075456}
		\includegraphics[width=\VisScale\linewidth]{figures/coco_results/supp/image/COCO_val2014_\im}&
		\includegraphics[width=\VisScale\linewidth]{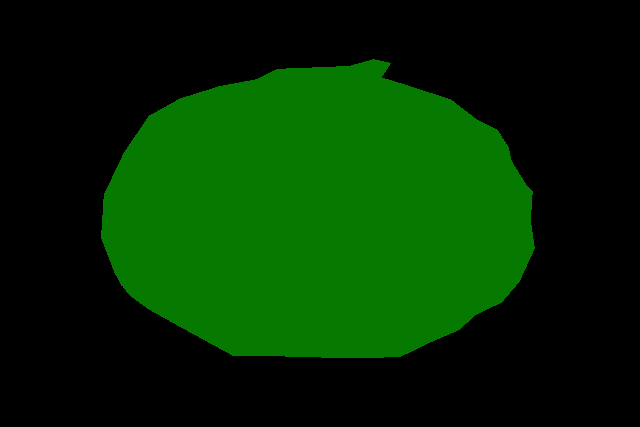}&
		\includegraphics[width=\VisScale\linewidth]{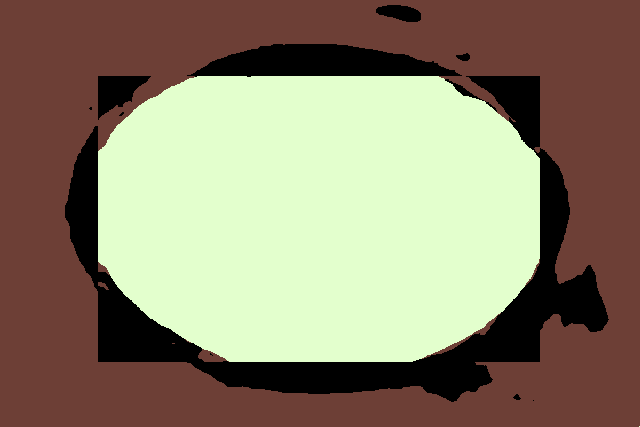}&
		\includegraphics[width=\VisScale\linewidth]{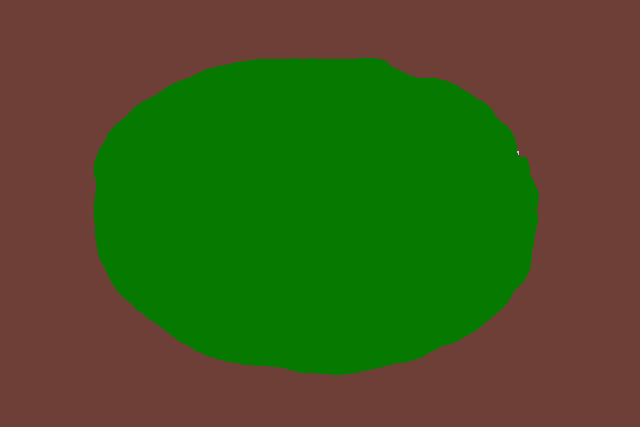}\\

		{\footnotesize{(a) Image}}&{\footnotesize{(b) Ground truth}}&{\footnotesize{(c) UPSNet~\cite{xiong2019upsnet}}}&{\footnotesize{(d) Ours}}\\
	\end{tabularx}
	\vspace{2pt}
	\caption{Qualitative results on COCO. Column (c) is produced by running the publicly available inference script of~\cite{xiong2019upsnet}. With our parametrised panoptic segmentation submodule, we are able to produce more coherent, accurate, and visually appealing predictions than the parameter-free approach of~\cite{xiong2019upsnet}.
	}
	\label{fig:supp_coco}
\end{figure*}